\documentclass{article}
\usepackage{enumitem}

\PassOptionsToPackage{numbers, compress}{natbib}

\usepackage[preprint]{neurips_2026}

\usepackage[utf8]{inputenc} 
\usepackage[T1]{fontenc}    
\usepackage{hyperref}       
\usepackage{url}            
\usepackage{booktabs}       
\usepackage{amsfonts}       
\usepackage{nicefrac}       
\usepackage{microtype}      
\usepackage{amsmath}
\usepackage{longtable}
\usepackage{multirow}
\usepackage{rotating}
\usepackage{graphicx}
\usepackage{subcaption}
\usepackage{listings}
\usepackage{adjustbox}
\usepackage{amsmath}
\usepackage{comment}
\usepackage[table]{xcolor}
\usepackage{placeins}
\usepackage{pgf} 
\usepackage{amssymb}    
\usepackage{pifont}     
\usepackage{array}      
\usepackage[scaled=0.85]{beramono} 
\usepackage{contour}
\usepackage{xurl}
\usepackage{tabularx} 
\usepackage{float}
\contourlength{0.3pt}

\definecolor{pastelgreen}{rgb}{0.71,0.94,0.71}
\definecolor{pastelred}{rgb}{0.96,0.71,0.71}

\usepackage{color-edits}
\addauthor[Emmy]{el}{purple}
\addauthor[Varun]{vg}{blue}
\addauthor[Steven]{sf}{orange}
\addauthor[Karan]{ks}{magenta}
\addauthor[Michael]{my}{red}
\addauthor[Derek]{dt}{violet}
\addauthor[Sachin]{sk}{cyan}

\newenvironment{tight_enumerate}[1][]{
\begin{enumerate}[leftmargin=*, #1]
  \setlength{\itemsep}{0pt}
  \setlength{\parskip}{0pt}
}{\end{enumerate}}

\newenvironment{tight_itemize}{
\begin{itemize}
  \setlength{\itemsep}{0pt}
  \setlength{\parskip}{0pt}
}{\end{itemize}}

\usepackage{xspace}
\newcommand{\hw}{\textsc{HalluWorld}}
\newcommand{\benchmarkname}{\hw\xspace}
\newcommand{\gridname}{\hw\textsc{-Grid}\xspace}
\newcommand{\terminalname}{\hw\textsc{-Terminal}\xspace}
\newcommand{\chessname}{\hw\textsc{-Chess}\xspace}

\title{HalluWorld: A Controlled Benchmark for Hallucination via Reference World Models}

\author{%
  Emmy Liu\thanks{All authors are affiliated with DegenAI Labs. Corresponding authors: Emmy Liu (\texttt{mengyan3@cs.cmu.edu}), Varun Gangal (\texttt{vgtomahawk@gmail.com}), and Steven Y. Feng (\texttt{syfeng@stanford.edu}).} \\
  Carnegie Mellon University \\
  \And
  Varun Gangal \\
  Patronus AI \\
  \And
  Michael Yu \\
  Independent Researcher \\
  \And
  Zhuofu Tao \\
  Independent Researcher \\
  \And
  Karan Singh \\
  Stanford University \\
  \And
  Sachin Kumar \\
  The Ohio State University \\
  \And
  Steven Y. Feng \\
  Stanford University \\
}

\begin{document}

\maketitle
\begingroup
\renewcommand\thefootnote{}
\footnotetext{Code and data: \texttt{\href{https://github.com/DegenAI-Labs/HalluWorld}{https://github.com/DegenAI-Labs/HalluWorld}}}
\addtocounter{footnote}{-1}
\endgroup


\begin{abstract}
    Hallucination remains a central failure mode of large language models, but existing benchmarks operationalize it inconsistently across tasks such as summarization, question answering, retrieval-augmented generation, and agentic interaction. This fragmentation makes it unclear whether a mitigation that works in one setting actually reduces hallucinations across contexts. 
    Current hallucination benchmarks either require human annotation and fixed references that may eventually be memorized, or rely on naturalistic observations often recorded in settings that are difficult to reproduce or test systematically. To enable further research on the root causes of hallucination, we introduce \textsc{HalluWorld}, an extensible benchmark framework grounded in an explicit reference-world formulation: a model hallucinates when it produces an observable claim that is false with respect to this reference world. Building on this view, we construct a family of synthetic and semi-synthetic benchmark environments in which the reference world is fully specified, the model's observable view is controlled, and hallucination labels can be generated automatically by construction. \benchmarkname spans multiple settings that are classically representative for AI, i.e., gridworlds, chess, and realistic terminal tasks. This enables controlled variation of key factors such as world complexity, observability, temporal change, and source-conflict policy, allowing us to disentangle hallucinations into more fine-grained error categories. We evaluate frontier and open-weight language models across these settings and find consistent patterns across domains: perceptual hallucination on directly observed information is near-solved for frontier models, while multi-step state tracking and causal forward simulation are still difficult for frontier models, and are not generally solved by extended thinking. In the terminal setting specifically, models also struggle with when to abstain from answering. The uneven profile of failures across probe types and domains suggest that different hallucinations arise from qualitatively distinct failure modes rather than reflecting a single underlying capability. Our results suggest that controlled reference worlds offer a scalable and reproducible path toward measuring and reducing hallucinations in modern language models.
\end{abstract}

\section{Introduction}

Despite years of effort on mitigation, hallucinations remain an unresolved problem in frontier language models (LM). As LM-based agents are increasingly used in high-stakes scenarios, it is imperative to understand exactly what conditions lead LMs to hallucinate and why. Despite being grouped under the same term, misreading a fact, misremembering a prior decision, incorrectly reasoning about the deployment context, or trusting misinformation are different failure modes which require different forms of mitigation. Although existing benchmarks have made progress in taxonomizing hallucinations \citep{ravichander-etal-2025-halogen, Zhang2025MIRAGE-Bench:Them, Bang2025HalluLens:Benchmark}, a critical limitation is that the environment, what the LM can observe, and how it must resolve conflicting information are fixed by design and cannot be varied independently, making it difficult to isolate failure modes. 

To address this, we ground our approach in a recent unified formulation of hallucination proposed by \citet{liu2026unifieddefinitionhallucinationits}%
, which decomposes hallucinations into a \textbf{reference world} encoding the set of ground truth states and transitions, a \textbf{view function} controlling what the model can observe, and a \textbf{conflict policy} deciding how conflicts between observable sources are settled. In this setting, a \textbf{hallucination} occurs when the model's output contradicts the ground-truth reference world. Making these components explicit allows us to stress-test models in a systematic way by independently varying one factor while keeping others fixed. To make this concrete, consider a terminal agent tasked with locating a configuration file: the same agent may answer correctly when given full directory output, hallucinate a file path when given partial output, and hallucinate a different path when a stale log contradicts what it can currently observe. By isolating which conditions trigger hallucination, we can proactively support the model with targeted interventions.

\begin{figure*}[!ht]
    \centering
    \includegraphics[width=0.95\textwidth]{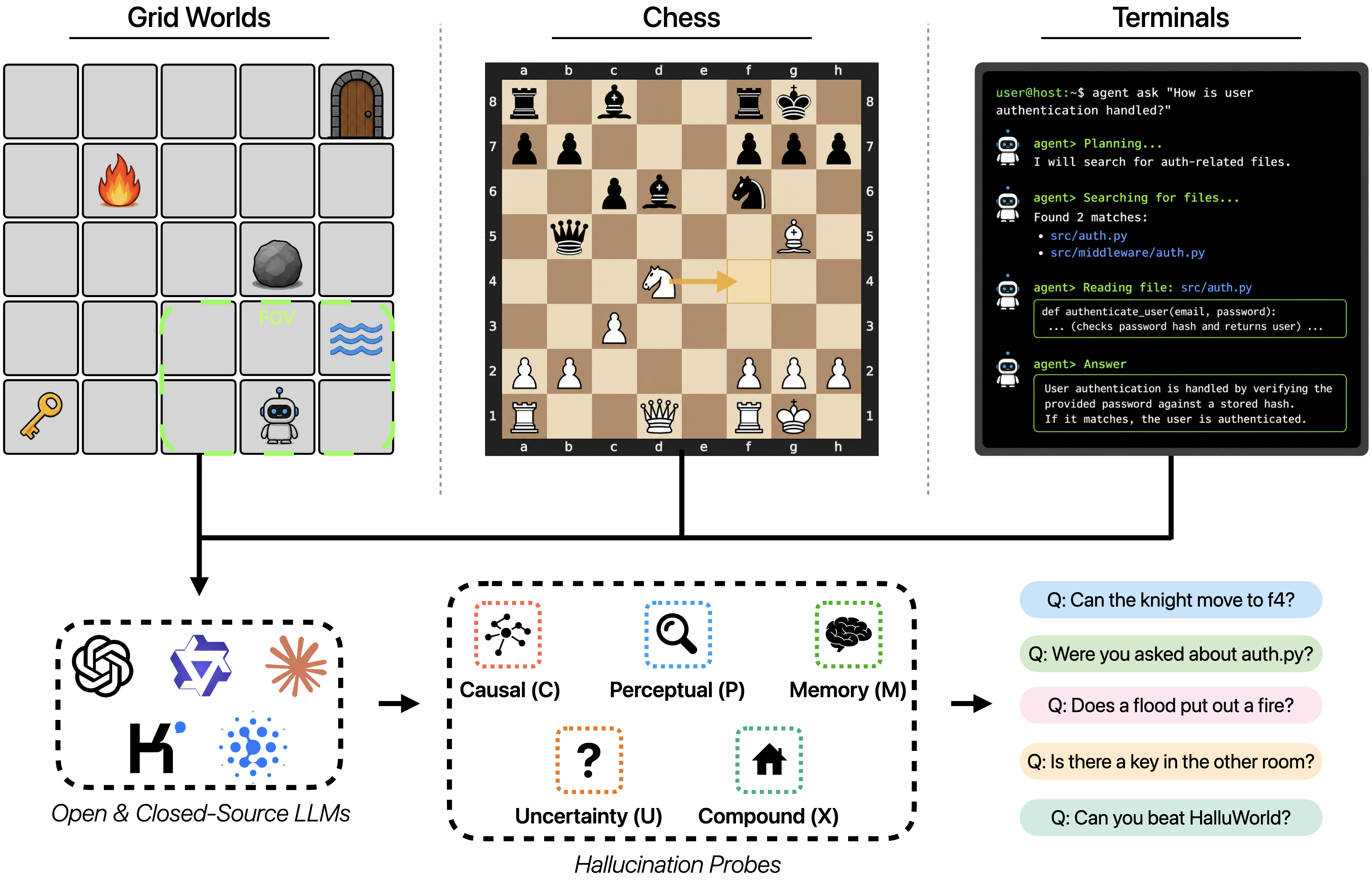}
    \caption{The \textbf{\textsc{HalluWorld} benchmark} spans three domains (gridworlds, chess, and terminals) and tests models using five probe categories targeting distinct cognitive skills: \textbf{Causal} (C) tests understanding of cause-effect relationships, \textbf{Perceptual} (P) tests spatial reasoning and object tracking, \textbf{Memory} (M) tests retention of past observations, \textbf{Uncertainty} (U) tests reasoning under partial observability, and \textbf{Compound/X} (X) tests multi-step reasoning across connected environments. Hallucination is measured by placed probes that query models about environment observations they have seen. 
    \benchmarkname qualitative examples from each of the three domains can be found in \S\ref{appendix:qualitative_analysis_illustration}.} 
    \label{fig:halluworld_overview}
\end{figure*}

We construct a family of benchmarks, \benchmarkname, with three domains spanning a customizable gridworld, chess environment, and agentic terminal environment. In these worlds, the ground truth labels for hallucinations are defined by construction, with no human labeling required. Crucially, \benchmarkname is designed for \textit{extensibility} -- new questions probing hallucinations can be easily added for each environment. 
Our contributions can be summarized as:
\vspace{-2mm}
\begin{tight_enumerate}
    \item We introduce \benchmarkname, a benchmark-framework that operationalizes hallucination as observable error relative to an explicit reference world and conflict policy.
    \item We instantiate our benchmark as a suite of controlled environments spanning gridworlds, chess, and terminal tasks. Our benchmark comprises 33 unique levels and 839 probe questions in the gridworld environment, 7 unique levels and 350 probes in the chess environment, and 110 unique terminal tasks and 529 probes in the terminal environment. Additionally, a level editor and trajectory recorder on the gridworld environment enable extension to more complex scenarios. 
    \item We benchmark a dozen frontier and open models on \benchmarkname and find that perceptual accuracy on direct observed information is near-solved on frontier models, while both memory hallucinations and causal forward simulation remain an issue even for frontier models. Furthermore, models consistently rely on environmental testimony over direct observation. Increasing thinking effort seldom uniformly helps, 
    suggesting that reasoning alone may not mitigate hallucinations.
\end{tight_enumerate}








\section{Related Works (Extended in \S\ref{app:related_works})}
Hallucination has traditionally been defined relative to a fixed source, e.g., unsupported content in summarization, and later broadened to factual errors in LLM outputs \citep{Lee2019HallucinationsTranslation,Maynez2020OnSummarization,Ji2024SurveyGeneration,Lin2022TruthfulQA:Falsehoods,Li2023HaluEval:Models}. Recent work highlights that benchmarks often conflate multiple failure modes and lack a consistent definition, motivating more structured formulations such as the ``world model'' view of hallucination \citep{Bang2025HalluLens:Benchmark,liu2026unifieddefinitionhallucinationits}.

Most existing benchmarks evaluate hallucination in \emph{static} settings: summarization and QA benchmarks define truth w.r.t. fixed documents or facts, while RAG benchmarks study conflicts between parametric and retrieved knowledge but treat provided context as the full reference \citep{Kryscinski2019EvaluatingSummarization,Min2023FActScore:Generation,Niu2024RAGTruth:Models,Friel2025RAGBench:Systems}. Recent work on RAG-considerate pretraining studies how models should allocate knowledge between parametric memory and retrieval during pretraining, highlighting that hallucination tendencies may depend on what information is made externally observable vs. learned in the parameters \cite{singh2026memorizeretrievescalinglaws}. Agent benchmarks such as \textsc{MIRAGE-Bench} and \textsc{AgentHallu} evaluate action-level hallucinations, but rely on snapshot audits or human annotation rather than fully specified, controllable worlds \citep{Zhang2025MIRAGE-Bench:Them,Liu2026AgentHallu:Agents}. 

\benchmarkname differs by evaluating hallucination in \emph{partially observed, evolving environments} with ground truth defined by simulator state. This enables automatic, reproducible labeling and controlled variation of observability, temporal dynamics, and conflicting evidence. By combining an explicit definition of hallucination with controllable environments and automatic labels, \benchmarkname provides a more precise and diagnostic framework for studying when and why hallucinations occur.

\section{The \benchmarkname Benchmark Suite}
\label{sec:halluworld_benchmark_suite}

\benchmarkname spans three domains: gridworlds, chess, and terminal tasks. Each instantiates a different set of challenges for models. Gridworlds are fully synthetic and customizable game-like environments, commonly used as testbeds for RL and planning \citep{sutton1998reinforcement,tovey2000gridworlds}. Hence, they provide a high degree of control over world state and specific challenges instantiated, while having minimal overlap with pretraining data other than game-specific priors. 
On the other hand, the terminal environment provides the most realistic scenarios for a deployed agent as the trajectories we use to generate the benchmark are based on real software engineering tasks. However, the complexity and noisiness of terminal contexts makes it somewhat harder to isolate specific failure modes. Chess occupies a middle ground: it is a real and well-specified domain with representation in pretraining, but it is easy to generate new board states and check correctness.

\textbf{Instantiating the reference world model framework.}
Our benchmark suite directly instantiates the formal hallucination definition proposed in recent work~\cite{liu2026unifieddefinitionhallucinationits}, which defines hallucination as observable world-model error w.r.t. a reference world $\mathcal{W} = (\mathcal{S}, \mathcal{H}, \mathcal{R})$, view function $V$, and truth function $T_{\mathcal{W},\mathcal{P}}$.
Table~\ref{tab:world-model-instantiation} shows how each environment family specifies these components.
In all three domains, $\mathcal{W}$ is \emph{well specified and programmatically known}, enabling automatic ground-truth generation without human/model annotation. Conflict policies $\mathcal{P}$ play a 
role when adversarial information arises (e.g., misleading signposts in gridworlds, transposed FEN strings in chess).

\begin{table}[t]
\centering
\small
\caption{Instantiation of the reference world model framework across the three \benchmarkname families. Each domain provides a fully specified $\mathcal{W}$, controllable $V$, and computable $T_{\mathcal{W},\mathcal{P}}$.}
\label{tab:world-model-instantiation}
\resizebox{\textwidth}{!}{
\begin{tabular}{lp{3.5cm}p{4cm}p{3.5cm}}
\toprule
\textbf{Family} & \textbf{$\mathcal{W} = (\mathcal{S}, \mathcal{H}, \mathcal{R})$} & \textbf{$V$ (View Function)} & \textbf{$T_{\mathcal{W},\mathcal{P}}$ Computation} \\
\midrule
\gridname &
  $\mathcal{S}$: Grid state (agent position, objects, door states);
  $\mathcal{H}$: Action history;
  $\mathcal{R}$: MiniGrid physics (movement, fire, flood, pressure plates) &
  FOV $\times$ serializer (Symbolic, Grid, Memory); controls observability and format &
  Computed from env state: presence, count, location, causal outcomes \\
\midrule
\chessname &
  $\mathcal{S}+\mathcal{H}$: FEN+Move history (PGN);
  $\mathcal{R}$: Chess rules &
  FEN + side-to-move &
  Computed via chess engine: legality, captures, pins \\
\midrule
\terminalname &
  $\mathcal{S}$: Filesystem state + process state;
  $\mathcal{H}$: Command execution trace;
  $\mathcal{R}$: Bash semantics + file operations &
  Terminal output trace (stdout/stderr) provides partial filesystem view &
  Co-generated with probes from trace: file existence, command success, state changes \\
\bottomrule
\end{tabular}
}
\end{table}

\textbf{Shared probe categories.}
\label{sec:shared-probe-categories}
We use a set of five \textbf{probe categories} for each domain, each
targeting a distinct cognitive demand. 
\textbf{Perceptual (P)} probes test accurate read-out of values directly present in the current observation.
\textbf{Memory (M)} probes require tracking values or states across multiple prior observations.
\textbf{Causal (C)} probes require reasoning about cause-effect relationships or forward-simulating the outcome of actions under environment mechanics.
\textbf{Uncertainty (U)} probes require recognizing the limits of available evidence and abstaining from a definite answer when context is insufficient. Chess omits uncertainty probes by design (\S\ref{sec:chess_explanation}).
\textbf{Compound (X)} probes require integrating evidence across multiple context sections, room visits, or artifact types simultaneously.

\subsection{\gridname: Interactive Environments for Controlled Hallucination Probing}
\label{sec:halluworld-grid_explanation}
\gridname is a family of hand-crafted gridworld environments, built on MiniGrid~\citep{minigrid}. It contains \textbf{33 unique levels} and 839 probe questions (see \S\ref{appendix:gridworld-all-levels} for a full list of levels). Hallucination labels are generated automatically from environment state. We provide a level editor and trajectory recorder to allow crafting of more complex environment-probe tuples (see \S\ref{appendix:gridworld-tools}). \textbf{Serializers} control \emph{how} the environment state is presented to the model; in other words, a manipulation of $V$ 
independent of the content itself. We implement three formats (Table~\ref{tab:serializers}).
 
\begin{table}[h]
\centering
\small
\caption{Serializer formats used in \gridname. Each controls the presentation of world state to the model, independently of what the state contains.}
\label{tab:serializers}
\resizebox{\textwidth}{!}{\begin{tabular}{llp{6cm}}
\toprule
\textbf{Serializer} & \textbf{Format} & \textbf{Primary test} \\
\midrule
Symbolic & Structured text object list & Reading comprehension, counting, state tracking \\
Grid     & ASCII art of full grid      & Spatial parsing, layout reasoning \\
Memory   & Symbolic + Grid combined, prior observations included & Temporal tracking, change detection \\
\bottomrule
\end{tabular}}
\end{table}


\textbf{Evaluation Details:} Levels are organized into the five probe categories (\S\ref{sec:shared-probe-categories}), with specific instantiations: \textbf{P} (6 levels) tests static-scene perception under varying
object density, orientations, and change conditions.
\textbf{M} (6 levels) probes temporal integration via multi-room traversal and
object tracking under river physics.
\textbf{C} (9 levels) introduces interactive mechanics:  fire, flood, and pressure plates, with harder levels presenting adversarial notice-boards that contradict observations (see
\S\ref{appendix:gridworld-mechanics}).
\textbf{U} (5 levels) spans unobserved rooms, signposts with varying 
reliability, and scenes with unanswerable questions.
\textbf{X} (7 levels) combines the above into extended multi-room episodes
probed across the full observation history.\footnote{The adversarial notice-boards and signposts, along with the agent's observations, are an example of multiple info sources, with conflict resolution policy $\mathcal{P}$ being that the agent should rely on its own observations (source of truth).} Within each level, probes take one of six closed-form answer types: \textbf{presence} (yes/no object detection),
\textbf{count} (exact integer), \textbf{state} (attribute/door status),
\textbf{location} (coordinate), \textbf{causal} (forward simulation outcome), and
\textbf{uncertainty} (answer concretely or \textit{``can't determine''}).

 

\subsection{\chessname: Structured World-Model Probes in a Rule-Governed Domain} 
\label{sec:chess_explanation}


Chess provides a complementary environment to gridworlds: chess is a domain well represented in training data~\citep{lichess2022database}, and a rich set of cognitively distinct question types that map naturally onto our probe categories. Chess also probes an adversarial form of hallucination: positions are drawn from real Lichess puzzles, so the model may simultaneously hold strong distributional priors over ``typical'' game continuations \emph{and} be confronted with an observed position that contradicts them.

\textbf{Environment and observation format.}
Each episode draws a position from a curated pool of Lichess FEN \citep{edwards1994standard} strings
(varied rating and theme strata). Before the probe is presented, $k$ random legal continuations are executed from that position (default $k=14$), advancing the board into novel territory and suppressing opening-book exploitation. The serializer then renders a fixed observation format: i) an ASCII 8$\times$8 board grid with labeled files (a--h) and ranks (1--8), ii) the side to move, and iii) an FEN string (standardized, single-line text used to describe a specific board position; optionally misleading with two pieces transposed).

\textbf{Probes.}
We evaluate 7 probes (see \S\ref{appendix:chess_all_probes}) across 3 of the probe categories
(§\ref{sec:halluworld_benchmark_suite}). Uncertainty is omitted as the observed board state always fully specifies the current position. Perceptual probes require static read-out from the current board.
Causal probes require forward simulation: applying one or two hypothetical moves
and evaluating a property of the resulting position.
Memory probes remove the board, requiring the model to maintain accurate
game state across a long move sequence. 
\textsc{ChatContext} mode 
prepends 4-16 past-game moves with commentary before the real observation (chess analogue of gridworld X levels), testing whether models separate context signal from noise.

\subsection{\terminalname: Agentic Hallucinations in Terminal Tasks}
\label{sec:halluworld-terminal}

\terminalname\ grounds \benchmarkname\ in realistic agentic deployment: the
reference world $\mathcal{W}$ is the full state of a live Linux filesystem, and
the view $V$ is the raw terminal history visible to the agent at a given trajectory step, which is $\approx$30-60K chars long.

\textbf{Probe construction.}
We derive 529 probes from 132 \textsc{Terminal-Bench} \citep{terminal-bench} trajectories spanning 110
software-engineering tasks (e.g., kernel builds, ML framework configuration, etc.), executed by a \texttt{gpt-4o-mini} agent (to elicit diverse error modes instead of optimized solutions from stronger models) 
in a live Linux environment. Probes are LLM-generated with (\texttt{gpt-5.4}, \texttt{reasoning\_effort=high}) and
grounded in file-system diffs (e.g. SHA-256 hashes, mtimes, directory listings) at the
trigger step.
We retain probes with \texttt{answerability\_score\,=\,5} and
\texttt{difficulty\_score\,$\geq$\,4}, obtaining an answerable and hard final set of probes.\footnote{Both scores are
assigned by the probe-generating LLM during creation. Answerability (1--5) measures
whether the probe has unambiguous ground truth from supplied context; difficulty
(1--5) measures the extent of reasoning required to answer.}
All 5 probe categories 
are instantiated with approximately equal coverage ($\approx105$-$108$ probes each). Each probe targets 1/5 hallucination failure modes: \textit{stale memory}
(187 probes) i.e., relying on an outdated value rather than the most recent terminal
output; \textit{cross-category reasoning} (110) i.e., failing to bridge between context
sections such as a process listing and a directory snapshot; \textit{uncertainty
overclaim} (107) i.e., asserting a definite answer with insufficient context;
\textit{causal shortcut} (104) i.e., incorrect attribution of cause in the trace; and \textit{version/API hallucination} (21) i.e., substituting train-time
API knowledge for observed behavior.

\textbf{Evaluation protocol.}
Models receive the unedited, ANSI-stripped terminal context and must return an
answer in a constrained key-value schema (e.g.,
\texttt{outcome=\textlangle ok\textbar fail\textrangle}).
Scoring is fully rule-based and no LLM judge is used.
Details shared in \S\ref{app:terminal-details}.

\section{Experimental Setup}
\label{sec:expt_setup}

We evaluate several models, both frontier closed and open-weight, with both thinking and non-thinking variants when applicable.\footnote{We use API credits from external providers (Baseten, OpenAI, and Anthropic) for open and closed models.}
\textbf{Non-thinking:} GPT-4o-mini, GPT-5.4-mini, GPT-4o, Sonnet~4.6,
Opus~4.6; DeepSeek-V3.1, GLM-5, Qwen-3-30B, Kimi~K2.6.
\textbf{Thinking:} o3-mini, o4-mini, o3, GPT-5.5, Sonnet~4.6~(T),
Opus~4.6~(T). All models with temperature are queried at $\tau=0$. Thinking models are by default queried at a \texttt{medium} level, with reasoning ablations for selected models per domain. For \gridname, each (level, serializer) pair is evaluated over 10 fixed seeds for nondeterministic levels (P1 and P3). For \chessname, each probe is run over 50 different board positions. For \terminalname, each probe is evaluated with the full runtime context plus the probe question, up to 60k characters with middle-truncation on longer agent trajectories. Unless otherwise specified, models are allocated 256 tokens for their answer and 16k tokens for thinking. 

\section{\gridname Results}
\label{sec:gridworld_results}

\begin{table*}[ht]
\centering
\caption{%
\textbf{Micro-averaged hallucination rates by probe category and serializer for \gridname}, with models ordered by overall hallucination. (T)\,=\,medium thinking. Columns are grouped into (left) probe categories — \textbf{P} (Perceptual),
\textbf{M} (Memory), \textbf{C} (Causal), \textbf{U} (Uncertainty),
\textbf{X} (Cross-category); and (right) input serializers — \textbf{Symbolic},
\textbf{Grid}, and \textbf{Memory}. The \textbf{Overall} column reports global
average hallucination across all conditions. Cell color reflects relative global hallucination rate
(green\,=\,low, red\,=\,high); \textbf{bold} marks the column minimum.
}
\label{tab:grid_combined_results}
\small
\setlength{\tabcolsep}{3.2pt}
\begin{tabular}{l r rrrrr rrr}
\toprule
\textbf{Model}
& & \multicolumn{5}{c}{\textbf{Probe Category}}
& \multicolumn{3}{c}{\textbf{Serializer}} \\
\cmidrule(lr){3-7}\cmidrule(lr){8-10}
& \textbf{Overall}
& \textbf{P} & \textbf{M} & \textbf{C} & \textbf{U} & \textbf{X}
& \textbf{Sym} & \textbf{Grid} & \textbf{Mem} \\
\midrule
GPT-5.5 (T)
  & \cellcolor{pastelgreen!89!pastelred}\textbf{5.0\%}
  & \cellcolor{pastelgreen!99!pastelred}0.4\%
  & \cellcolor{pastelgreen!88!pastelred}5.9\%
  & \cellcolor{pastelgreen!47!pastelred}25.3\%
  & \cellcolor{pastelgreen!100!pastelred}0.1\%
  & \cellcolor{pastelgreen!94!pastelred}\textbf{2.7\%}
  & \cellcolor{pastelgreen!100!pastelred}\textbf{1.1\%}
  & \cellcolor{pastelgreen!95!pastelred}\textbf{3.1\%}
  & \cellcolor{pastelgreen!47!pastelred}21.3\% \\

Claude Opus 4.6
  & \cellcolor{pastelgreen!88!pastelred}5.9\%
  & \cellcolor{pastelgreen!92!pastelred}3.7\%
  & \cellcolor{pastelgreen!92!pastelred}\textbf{3.8\%}
  & \cellcolor{pastelgreen!58!pastelred}20.1\%
  & \cellcolor{pastelgreen!93!pastelred}3.4\%
  & \cellcolor{pastelgreen!90!pastelred}4.6\%
  & \cellcolor{pastelgreen!93!pastelred}3.8\%
  & \cellcolor{pastelgreen!90!pastelred}5.1\%
  & \cellcolor{pastelgreen!63!pastelred}\textbf{15.5\%} \\

Claude Opus 4.6 (T)
  & \cellcolor{pastelgreen!87!pastelred}6.4\%
  & \cellcolor{pastelgreen!96!pastelred}1.7\%
  & \cellcolor{pastelgreen!92!pastelred}3.8\%
  & \cellcolor{pastelgreen!47!pastelred}25.2\%
  & \cellcolor{pastelgreen!93!pastelred}3.4\%
  & \cellcolor{pastelgreen!90!pastelred}4.7\%
  & \cellcolor{pastelgreen!93!pastelred}3.7\%
  & \cellcolor{pastelgreen!90!pastelred}4.9\%
  & \cellcolor{pastelgreen!52!pastelred}19.6\% \\

o3
  & \cellcolor{pastelgreen!86!pastelred}6.8\%
  & \cellcolor{pastelgreen!99!pastelred}\textbf{0.3\%}
  & \cellcolor{pastelgreen!88!pastelred}5.5\%
  & \cellcolor{pastelgreen!45!pastelred}26.0\%
  & \cellcolor{pastelgreen!99!pastelred}0.3\%
  & \cellcolor{pastelgreen!88!pastelred}5.7\%
  & \cellcolor{pastelgreen!96!pastelred}2.7\%
  & \cellcolor{pastelgreen!87!pastelred}6.2\%
  & \cellcolor{pastelgreen!49!pastelred}20.6\% \\

Claude Sonnet 4.6
  & \cellcolor{pastelgreen!85!pastelred}7.2\%
  & \cellcolor{pastelgreen!94!pastelred}2.8\%
  & \cellcolor{pastelgreen!89!pastelred}5.0\%
  & \cellcolor{pastelgreen!59!pastelred}\textbf{19.7\%}
  & \cellcolor{pastelgreen!98!pastelred}1.1\%
  & \cellcolor{pastelgreen!84!pastelred}7.6\%
  & \cellcolor{pastelgreen!90!pastelred}4.8\%
  & \cellcolor{pastelgreen!84!pastelred}7.2\%
  & \cellcolor{pastelgreen!60!pastelred}16.6\% \\

o4-mini
  & \cellcolor{pastelgreen!84!pastelred}7.4\%
  & \cellcolor{pastelgreen!98!pastelred}0.8\%
  & \cellcolor{pastelgreen!91!pastelred}4.5\%
  & \cellcolor{pastelgreen!35!pastelred}30.6\%
  & \cellcolor{pastelgreen!98!pastelred}0.9\%
  & \cellcolor{pastelgreen!88!pastelred}5.8\%
  & \cellcolor{pastelgreen!95!pastelred}2.9\%
  & \cellcolor{pastelgreen!86!pastelred}6.4\%
  & \cellcolor{pastelgreen!41!pastelred}23.6\% \\

Claude Sonnet 4.6 (T)
  & \cellcolor{pastelgreen!84!pastelred}7.7\%
  & \cellcolor{pastelgreen!98!pastelred}1.1\%
  & \cellcolor{pastelgreen!87!pastelred}6.2\%
  & \cellcolor{pastelgreen!49!pastelred}24.0\%
  & \cellcolor{pastelgreen!98!pastelred}1.1\%
  & \cellcolor{pastelgreen!84!pastelred}7.5\%
  & \cellcolor{pastelgreen!90!pastelred}5.0\%
  & \cellcolor{pastelgreen!87!pastelred}6.1\%
  & \cellcolor{pastelgreen!50!pastelred}20.4\% \\

o3-mini
  & \cellcolor{pastelgreen!83!pastelred}8.1\%
  & \cellcolor{pastelgreen!99!pastelred}0.4\%
  & \cellcolor{pastelgreen!90!pastelred}4.9\%
  & \cellcolor{pastelgreen!50!pastelred}23.8\%
  & \cellcolor{pastelgreen!88!pastelred}5.9\%
  & \cellcolor{pastelgreen!84!pastelred}7.6\%
  & \cellcolor{pastelgreen!90!pastelred}4.9\%
  & \cellcolor{pastelgreen!84!pastelred}7.2\%
  & \cellcolor{pastelgreen!50!pastelred}20.2\% \\

Kimi K2.6
  & \cellcolor{pastelgreen!80!pastelred}9.3\%
  & \cellcolor{pastelgreen!89!pastelred}5.4\%
  & \cellcolor{pastelgreen!87!pastelred}6.2\%
  & \cellcolor{pastelgreen!0!pastelred}47.4\%
  & \cellcolor{pastelgreen!96!pastelred}1.9\%
  & \cellcolor{pastelgreen!81!pastelred}8.8\%
  & \cellcolor{pastelgreen!92!pastelred}4.3\%
  & \cellcolor{pastelgreen!77!pastelred}10.1\%
  & \cellcolor{pastelgreen!10!pastelred}35.5\% \\

GPT-4o
  & \cellcolor{pastelgreen!73!pastelred}12.6\%
  & \cellcolor{pastelgreen!79!pastelred}10.0\%
  & \cellcolor{pastelgreen!64!pastelred}17.2\%
  & \cellcolor{pastelgreen!39!pastelred}29.0\%
  & \cellcolor{pastelgreen!99!pastelred}0.5\%
  & \cellcolor{pastelgreen!76!pastelred}11.3\%
  & \cellcolor{pastelgreen!85!pastelred}6.8\%
  & \cellcolor{pastelgreen!67!pastelred}13.6\%
  & \cellcolor{pastelgreen!30!pastelred}27.9\% \\

GLM-5
  & \cellcolor{pastelgreen!70!pastelred}14.3\%
  & \cellcolor{pastelgreen!58!pastelred}19.8\%
  & \cellcolor{pastelgreen!67!pastelred}15.8\%
  & \cellcolor{pastelgreen!44!pastelred}26.4\%
  & \cellcolor{pastelgreen!100!pastelred}\textbf{0.0\%}
  & \cellcolor{pastelgreen!72!pastelred}13.4\%
  & \cellcolor{pastelgreen!81!pastelred}8.5\%
  & \cellcolor{pastelgreen!57!pastelred}17.7\%
  & \cellcolor{pastelgreen!38!pastelred}24.9\% \\

DeepSeek-V3-0324
  & \cellcolor{pastelgreen!68!pastelred}15.4\%
  & \cellcolor{pastelgreen!50!pastelred}23.5\%
  & \cellcolor{pastelgreen!70!pastelred}14.3\%
  & \cellcolor{pastelgreen!32!pastelred}32.2\%
  & \cellcolor{pastelgreen!90!pastelred}4.6\%
  & \cellcolor{pastelgreen!73!pastelred}12.7\%
  & \cellcolor{pastelgreen!77!pastelred}9.9\%
  & \cellcolor{pastelgreen!58!pastelred}17.3\%
  & \cellcolor{pastelgreen!28!pastelred}28.6\% \\

GPT-5.4-mini
  & \cellcolor{pastelgreen!55!pastelred}21.5\%
  & \cellcolor{pastelgreen!58!pastelred}19.8\%
  & \cellcolor{pastelgreen!44!pastelred}26.6\%
  & \cellcolor{pastelgreen!53!pastelred}22.2\%
  & \cellcolor{pastelgreen!88!pastelred}5.7\%
  & \cellcolor{pastelgreen!49!pastelred}24.0\%
  & \cellcolor{pastelgreen!63!pastelred}15.4\%
  & \cellcolor{pastelgreen!27!pastelred}29.0\%
  & \cellcolor{pastelgreen!42!pastelred}23.4\% \\

Qwen-3-30B
  & \cellcolor{pastelgreen!51!pastelred}23.3\%
  & \cellcolor{pastelgreen!53!pastelred}22.4\%
  & \cellcolor{pastelgreen!42!pastelred}27.7\%
  & \cellcolor{pastelgreen!33!pastelred}31.6\%
  & \cellcolor{pastelgreen!72!pastelred}13.2\%
  & \cellcolor{pastelgreen!51!pastelred}23.0\%
  & \cellcolor{pastelgreen!56!pastelred}18.1\%
  & \cellcolor{pastelgreen!33!pastelred}27.0\%
  & \cellcolor{pastelgreen!23!pastelred}30.8\% \\

GPT-4o-mini
  & \cellcolor{pastelgreen!41!pastelred}28.1\%
  & \cellcolor{pastelgreen!26!pastelred}34.9\%
  & \cellcolor{pastelgreen!18!pastelred}39.0\%
  & \cellcolor{pastelgreen!35!pastelred}30.9\%
  & \cellcolor{pastelgreen!89!pastelred}5.1\%
  & \cellcolor{pastelgreen!39!pastelred}28.7\%
  & \cellcolor{pastelgreen!54!pastelred}18.8\%
  & \cellcolor{pastelgreen!0!pastelred}39.5\%
  & \cellcolor{pastelgreen!22!pastelred}31.2\% \\
\bottomrule
\end{tabular}
\end{table*}

\subsection{Category-Level Summary}
\label{sec:results-tier}

Table~\ref{tab:grid_combined_results} reports probe-level hallucination rates by category for each model.
A consistent hierarchy holds for most models: $\text{U} < \text{P} < \text{M} < \text{C}$,
with X showing high variance.
Presence probes approach 100\% accuracy for strong models, while the causal category is difficult even for frontier models (Claude Opus~4.6: 20.1\%, GPT-5.5: 25.3\%).
Two frontier clusters emerge: reasoning models (GPT-5.5, Claude Opus/Sonnet~4.6,
o3, o4-mini, o3-mini) achieve 5.0-8.1\% overall, while non-reasoning frontier models
(GPT-4o) and most open-weight models cluster at 12.6-15.4\%.
Kimi~K2.6 shows a striking anomaly: a low overall rate of 9.3\% driven almost
entirely by the C-category at 47.4\%, while remaining relatively low on P and U.
This breakdown highlights that overall hallucination rate conflates
different failure modes, motivating category-level decomposition.
 \subsection{Serializer as an Independent Variable}
\label{sec:serializer-results}

The format in which the environment state is presented has a notable effect on
hallucination, as seen in \autoref{tab:grid_combined_results}. Grid serialization is harder overall, but the damage is probe-specific:
allocentric rotation (P3) collapses to near 0\% under grid for all models, while
presence probes remain near-ceiling in both formats. Cross-observation attribute tracking (M2) is catastrophic under grid even for strong models, yet count probes are
differentially preserved -- o3-mini maintains near-100\% count accuracy under grid
where non-thinking GPT models drop to 13--25\%. This serializer and probe type
$\times$ model interaction cannot be decomposed by existing benchmarks, and has
a direct practical implication: the format in which an agent receives environment state is an independent driver of hallucination.

\subsection{Thinking Models}
\label{sec:thinking}

Comparing thinking and non-thinking variants within the same model family reveals a
consistent and counterintuitive pattern.
Extended thinking slightly improves perceptual accuracy (Sonnet~4.6 P:
$2.8\% \to 1.1\%$; Opus~4.6 P: $3.7\% \to 1.7\%$) but \emph{worsens} causal
hallucination (Sonnet~4.6 C: $19.7\% \to 24.0\%$; Opus~4.6 C:
$20.1\% \to 25.2\%$), with negligible difference on uncertainty. We hypothesize that this may be caused by perceptual probes having a clear answer that is derivable by carefully reading the context through an extended CoT, whereas causal probes require open-ended forward simulation; additional reasoning provides more opportunity to confabulate plausible intermediate states rather than trusting direct observations. This echoes results in VLMs where extended reasoning can increase hallucination \citep{liu2025thinkingseeingassessingamplified}. See \S\ref{appendix:gridworld-reasoning-ablation} for full reasoning ablation.

\subsection{Hard Subset}
\label{sec:hard}
The Hard subset comprises 12 (level, serializer) pairs where $\geq$\,5 models
achieve $\geq$\,20\% hallucination rate (full results in
\S\ref{appendix:gridworld-hard}), skewing toward memory-serializer causal levels
and grid-serializer perceptual levels -- confirming serializer format as a
first-order difficulty driver.
Frontier models achieve 15.5-21.3\% mean hallucination rates on this subset;
thinking models provide no consistent advantage (Sonnet~4.6: 17.9\% with thinking
vs.\ 20.0\% without), consistent with the causal thinking paradox in
\S\ref{sec:thinking}.
Kimi~K2.6 is the clearest outlier at 53.9\%, driven by near-total failure on the
C1b variants (95.0\% and 100.0\%), and C6 (Flood-Fire Escape) is among the hardest levels, where adversarial testimony compounds causal reasoning under a
time constraint (Sonnet~4.6: 56.1\%). Overall, we see that the Hard subset of \gridname is effective and challenges even frontier models.  

\subsection{Failure Mode Analysis}
\label{sec:failure-modes}

\textbf{Over-trusting textual information.}
Models consistently privilege written testimony over direct observation. In M4
(Unreliable Narrator), even strong models partially defer to a signpost that contradicts
what they can directly observe. C1a with-board vs.\ no-board sharpens
this: Opus~4.6 hallucinates \emph{more} with an \emph{accurate} notice board
present (42.0\% vs.\ 28.4\% without), suggesting that written rules (whether accurate or not) can derail
implicit causal reasoning rather than support it. 

\textbf{Failure to track multi-step state changes.}
Memory levels reveal that models struggle to maintain accurate world state across observations. In M1 (River Field), models fallback to the
stale notice board rather than computing current object positions from elapsed steps. In M2 (Witness Stand), cross-chamber attribute tracking collapses
under grid serialization even when presence detection remains intact, isolating the
failure to the binding of properties to objects across time. 

\textbf{Difficulty reasoning about future states.}
Causal levels requiring forward simulation are the hardest. C3 (Flood Room) demands prediction of which tiles will
be passable at a future timestep. Hallucination rates reveal a striking
within-frontier split: Opus~4.6 and Sonnet~4.6 achieve 0\%, while o3 reaches
43.3\% and o4-mini 33.3\% -- reflecting 
different causal reasoning behavior 
and exposing distinct failure modes across model families. 
C6 (Flood-Fire Escape) compounds forward simulation with adversarial testimony, and is among the hardest levels for frontier models (Sonnet~4.6: 56.1\%). Overall, 
\gridname allows us to stress-test models effectively by isolating features such as the grid serialization, forward-simulation dynamics, and adversarial testimony.

\subsection{In-Navigation Setting (InNav)}
To study \gridname in a more interactive setting and mirror real-world agent deployment, we evaluate models in an In-Navigation (\textsc{InNav}) setting where they answer probes while actively navigating the environment. By comparing \textsc{InNav} to a \textsc{CtrlStatic} condition with identical trajectories, we can examine two effects: epistemic grounding from action-perception feedback vs. cognitive load from navigation demands. We find that navigation often reduces hallucination for reasoning models (e.g., up to $-13.2\%$), especially in causal and uncertainty categories, but can increase errors in memory or long-horizon tasks, where cognitive load dominates. Full details are in \S\ref{appendix:in-navigation}. Clearly, hallucination depends on more than just static reasoning ability, but on how models interact with their environment. This demonstrates that \gridname is also useful for evaluating agents in more realistic, embodied contexts.

\section{\chessname Results}
\label{sec:chess_results}

\begin{table*}[ht]
\centering
\caption{\textbf{\chessname hallucination rates by probe category under two serializers: excluding a FEN and including an incorrect FEN.} Columns: Overall (all 7 probes, $\sim$348 questions); P\,=\,Perceptual (\texttt{can\_capture}, \texttt{defended}, \texttt{hanging}); C\,=\,Causal (\texttt{hypothetical\_in\_check}, \texttt{after\_move\_undefended\_count}); M\,=\,Memory (\texttt{hidden\_side\_capture\_stats}); X\,=\,Compound (\texttt{san\_legal\_move}). Color marks \textit{relative} hallucination rate (green\,=\,low, red\,=\,high); \textbf{bold} marks column minimum.}
\label{tab:chess-results-combined}
\scriptsize
\setlength{\tabcolsep}{2.5pt}
\resizebox{\textwidth}{!}{%
\begin{tabular}{lrrrrrrrrrr}
\toprule
\textbf{Model}
& \multicolumn{5}{c}{\textbf{No FEN}}
& \multicolumn{5}{c}{\textbf{Incorrect FEN}} \\
\cmidrule(lr){2-6}\cmidrule(lr){7-11}
& \textbf{Overall} & \textbf{P} & \textbf{C} & \textbf{M} & \textbf{X}
& \textbf{Overall} & \textbf{P} & \textbf{C} & \textbf{M} & \textbf{X} \\
\midrule
GPT-5.5 (T)
& \cellcolor{pastelgreen!100!pastelred}\textbf{0.3\%} & \cellcolor{pastelgreen!100!pastelred}\textbf{0.0\%} & \cellcolor{pastelgreen!98!pastelred}\textbf{1.0\%} & \cellcolor{pastelgreen!100!pastelred}\textbf{0.0\%} & \cellcolor{pastelgreen!100!pastelred}\textbf{0.0\%}
& \cellcolor{pastelgreen!94!pastelred}\textbf{4.0\%} & \cellcolor{pastelgreen!95!pastelred}\textbf{3.3\%} & \cellcolor{pastelgreen!86!pastelred}9.0\% & \cellcolor{pastelgreen!100!pastelred}\textbf{0.0\%} & \cellcolor{pastelgreen!100!pastelred}\textbf{0.0\%} \\
o3
& \cellcolor{pastelgreen!97!pastelred}2.0\% & \cellcolor{pastelgreen!99!pastelred}0.7\% & \cellcolor{pastelgreen!97!pastelred}2.0\% & \cellcolor{pastelgreen!97!pastelred}2.1\% & \cellcolor{pastelgreen!91!pastelred}6.0\%
& \cellcolor{pastelgreen!91!pastelred}6.0\% & \cellcolor{pastelgreen!93!pastelred}4.7\% & \cellcolor{pastelgreen!83!pastelred}11.0\% & \cellcolor{pastelgreen!100!pastelred}\textbf{0.0\%} & \cellcolor{pastelgreen!91!pastelred}6.0\% \\
GPT-5.4
& \cellcolor{pastelgreen!96!pastelred}2.6\% & \cellcolor{pastelgreen!100!pastelred}\textbf{0.0\%} & \cellcolor{pastelgreen!95!pastelred}3.0\% & \cellcolor{pastelgreen!87!pastelred}8.3\% & \cellcolor{pastelgreen!94!pastelred}4.0\%
& \cellcolor{pastelgreen!94!pastelred}4.0\% & \cellcolor{pastelgreen!93!pastelred}4.7\% & \cellcolor{pastelgreen!94!pastelred}\textbf{4.0\%} & \cellcolor{pastelgreen!94!pastelred}4.2\% & \cellcolor{pastelgreen!97!pastelred}2.0\% \\
o4-mini
& \cellcolor{pastelgreen!91!pastelred}5.7\% & \cellcolor{pastelgreen!97!pastelred}2.0\% & \cellcolor{pastelgreen!89!pastelred}7.0\% & \cellcolor{pastelgreen!68!pastelred}20.8\% & \cellcolor{pastelgreen!100!pastelred}\textbf{0.0\%}
& \cellcolor{pastelgreen!87!pastelred}8.3\% & \cellcolor{pastelgreen!92!pastelred}5.3\% & \cellcolor{pastelgreen!82!pastelred}12.0\% & \cellcolor{pastelgreen!78!pastelred}14.6\% & \cellcolor{pastelgreen!94!pastelred}4.0\% \\
o3-mini
& \cellcolor{pastelgreen!71!pastelred}18.7\% & \cellcolor{pastelgreen!98!pastelred}1.3\% & \cellcolor{pastelgreen!40!pastelred}39.0\% & \cellcolor{pastelgreen!68!pastelred}20.8\% & \cellcolor{pastelgreen!57!pastelred}28.0\%
& \cellcolor{pastelgreen!62!pastelred}24.7\% & \cellcolor{pastelgreen!82!pastelred}12.0\% & \cellcolor{pastelgreen!31!pastelred}45.0\% & \cellcolor{pastelgreen!55!pastelred}29.2\% & \cellcolor{pastelgreen!72!pastelred}18.0\% \\
Claude Sonnet 4.6 (T)
& \cellcolor{pastelgreen!70!pastelred}19.8\% & \cellcolor{pastelgreen!84!pastelred}10.7\% & \cellcolor{pastelgreen!32!pastelred}44.0\% & \cellcolor{pastelgreen!87!pastelred}8.3\% & \cellcolor{pastelgreen!85!pastelred}10.0\%
& \cellcolor{pastelgreen!60!pastelred}26.1\% & \cellcolor{pastelgreen!71!pastelred}18.7\% & \cellcolor{pastelgreen!29!pastelred}46.0\% & \cellcolor{pastelgreen!87!pastelred}8.3\% & \cellcolor{pastelgreen!60!pastelred}26.0\% \\
GPT-5.4-mini (T)
& \cellcolor{pastelgreen!66!pastelred}22.4\% & \cellcolor{pastelgreen!89!pastelred}7.3\% & \cellcolor{pastelgreen!80!pastelred}13.0\% & \cellcolor{pastelgreen!0!pastelred}75.0\% & \cellcolor{pastelgreen!55!pastelred}36.0\%
& \cellcolor{pastelgreen!66!pastelred}22.1\% & \cellcolor{pastelgreen!87!pastelred}8.7\% & \cellcolor{pastelgreen!71!pastelred}19.0\% & \cellcolor{pastelgreen!0!pastelred}75.0\% & \cellcolor{pastelgreen!72!pastelred}18.0\% \\
Claude Opus 4.6 (T)
& \cellcolor{pastelgreen!62!pastelred}24.7\% & \cellcolor{pastelgreen!65!pastelred}22.7\% & \cellcolor{pastelgreen!26!pastelred}48.0\% & \cellcolor{pastelgreen!97!pastelred}2.1\% & \cellcolor{pastelgreen!91!pastelred}6.0\%
& \cellcolor{pastelgreen!60!pastelred}25.9\% & \cellcolor{pastelgreen!60!pastelred}26.0\% & \cellcolor{pastelgreen!29!pastelred}46.0\% & \cellcolor{pastelgreen!90!pastelred}6.2\% & \cellcolor{pastelgreen!94!pastelred}4.0\% \\
Claude Opus 4.6
& \cellcolor{pastelgreen!49!pastelred}33.3\% & \cellcolor{pastelgreen!57!pastelred}28.0\% & \cellcolor{pastelgreen!23!pastelred}50.0\% & \cellcolor{pastelgreen!55!pastelred}29.2\% & \cellcolor{pastelgreen!69!pastelred}20.0\%
& \cellcolor{pastelgreen!49!pastelred}33.0\% & \cellcolor{pastelgreen!55!pastelred}29.3\% & \cellcolor{pastelgreen!18!pastelred}53.0\% & \cellcolor{pastelgreen!62!pastelred}25.0\% & \cellcolor{pastelgreen!82!pastelred}12.0\% \\
Claude Sonnet 4.6
& \cellcolor{pastelgreen!39!pastelred}39.7\% & \cellcolor{pastelgreen!49!pastelred}33.3\% & \cellcolor{pastelgreen!25!pastelred}49.0\% & \cellcolor{pastelgreen!52!pastelred}31.3\% & \cellcolor{pastelgreen!26!pastelred}48.0\%
& \cellcolor{pastelgreen!39!pastelred}39.7\% & \cellcolor{pastelgreen!45!pastelred}36.0\% & \cellcolor{pastelgreen!25!pastelred}49.0\% & \cellcolor{pastelgreen!55!pastelred}29.2\% & \cellcolor{pastelgreen!35!pastelred}42.0\% \\
Kimi-K2.6
& \cellcolor{pastelgreen!32!pastelred}44.5\% & \cellcolor{pastelgreen!47!pastelred}34.7\% & \cellcolor{pastelgreen!48!pastelred}34.0\% & \cellcolor{pastelgreen!17!pastelred}54.2\% & \cellcolor{pastelgreen!0!pastelred}86.0\%
& \cellcolor{pastelgreen!35!pastelred}42.2\% & \cellcolor{pastelgreen!35!pastelred}42.0\% & \cellcolor{pastelgreen!46!pastelred}35.0\% & \cellcolor{pastelgreen!13!pastelred}56.2\% & \cellcolor{pastelgreen!32!pastelred}44.0\% \\
GPT-5.5 (NT)
& \cellcolor{pastelgreen!26!pastelred}48.3\% & \cellcolor{pastelgreen!27!pastelred}47.3\% & \cellcolor{pastelgreen!34!pastelred}43.0\% & \cellcolor{pastelgreen!0!pastelred}85.4\% & \cellcolor{pastelgreen!60!pastelred}26.0\%
& \cellcolor{pastelgreen!25!pastelred}48.9\% & \cellcolor{pastelgreen!26!pastelred}48.0\% & \cellcolor{pastelgreen!20!pastelred}52.0\% & \cellcolor{pastelgreen!0!pastelred}87.5\% & \cellcolor{pastelgreen!88!pastelred}8.0\% \\
DeepSeek-V3.1
& \cellcolor{pastelgreen!18!pastelred}53.4\% & \cellcolor{pastelgreen!26!pastelred}48.0\% & \cellcolor{pastelgreen!57!pastelred}28.0\% & \cellcolor{pastelgreen!0!pastelred}91.7\% & \cellcolor{pastelgreen!0!pastelred}84.0\%
& \cellcolor{pastelgreen!23!pastelred}50.0\% & \cellcolor{pastelgreen!34!pastelred}42.7\% & \cellcolor{pastelgreen!51!pastelred}32.0\% & \cellcolor{pastelgreen!0!pastelred}95.8\% & \cellcolor{pastelgreen!2!pastelred}64.0\% \\
GLM-5
& \cellcolor{pastelgreen!12!pastelred}56.9\% & \cellcolor{pastelgreen!31!pastelred}44.7\% & \cellcolor{pastelgreen!34!pastelred}43.0\% & \cellcolor{pastelgreen!0!pastelred}85.4\% & \cellcolor{pastelgreen!0!pastelred}94.0\%
& \cellcolor{pastelgreen!27!pastelred}47.4\% & \cellcolor{pastelgreen!32!pastelred}44.0\% & \cellcolor{pastelgreen!43!pastelred}37.0\% & \cellcolor{pastelgreen!0!pastelred}89.6\% & \cellcolor{pastelgreen!42!pastelred}38.0\% \\
Qwen-3-30B
& \cellcolor{pastelgreen!8!pastelred}59.8\% & \cellcolor{pastelgreen!29!pastelred}46.0\% & \cellcolor{pastelgreen!28!pastelred}47.0\% & \cellcolor{pastelgreen!0!pastelred}95.8\% & \cellcolor{pastelgreen!0!pastelred}92.0\%
& \cellcolor{pastelgreen!8!pastelred}59.5\% & \cellcolor{pastelgreen!28!pastelred}46.7\% & \cellcolor{pastelgreen!35!pastelred}42.0\% & \cellcolor{pastelgreen!0!pastelred}95.8\% & \cellcolor{pastelgreen!0!pastelred}98.0\% \\
GPT-5.4-mini (NT)
& \cellcolor{pastelgreen!6!pastelred}60.9\% & \cellcolor{pastelgreen!24!pastelred}49.3\% & \cellcolor{pastelgreen!20!pastelred}52.0\% & \cellcolor{pastelgreen!0!pastelred}83.3\% & \cellcolor{pastelgreen!0!pastelred}92.0\%
& \cellcolor{pastelgreen!16!pastelred}54.3\% & \cellcolor{pastelgreen!28!pastelred}46.7\% & \cellcolor{pastelgreen!15!pastelred}55.0\% & \cellcolor{pastelgreen!0!pastelred}87.5\% & \cellcolor{pastelgreen!32!pastelred}44.0\% \\
GPT-4o
& \cellcolor{pastelgreen!3!pastelred}62.9\% & \cellcolor{pastelgreen!21!pastelred}51.3\% & \cellcolor{pastelgreen!18!pastelred}53.0\% & \cellcolor{pastelgreen!0!pastelred}91.7\% & \cellcolor{pastelgreen!0!pastelred}90.0\%
& \cellcolor{pastelgreen!11!pastelred}57.8\% & \cellcolor{pastelgreen!22!pastelred}50.7\% & \cellcolor{pastelgreen!8!pastelred}60.0\% & \cellcolor{pastelgreen!0!pastelred}87.5\% & \cellcolor{pastelgreen!29!pastelred}46.0\% \\
GPT-4o-mini
& \cellcolor{pastelgreen!3!pastelred}62.9\% & \cellcolor{pastelgreen!21!pastelred}51.3\% & \cellcolor{pastelgreen!20!pastelred}52.0\% & \cellcolor{pastelgreen!0!pastelred}85.4\% & \cellcolor{pastelgreen!0!pastelred}98.0\%
& \cellcolor{pastelgreen!3!pastelred}62.9\% & \cellcolor{pastelgreen!15!pastelred}55.3\% & \cellcolor{pastelgreen!23!pastelred}50.0\% & \cellcolor{pastelgreen!0!pastelred}79.2\% & \cellcolor{pastelgreen!0!pastelred}96.0\% \\
\bottomrule
\end{tabular}%
}
\end{table*}


\textbf{Category-level summary.}
Non-reasoning models and those without extended reasoning cluster tightly on P
and C: \texttt{gpt-4o}, \texttt{gpt-4o-mini}, and \texttt{gpt-5.4-mini} without
reasoning are nearly indistinguishable on those two categories with $\approx$50\%
hallucination. All three collapse on both long-sequence categories:
M and X reach avg. hallucination >80\%. The same broad pattern holds for the open-weight models:
Qwen-3-30B, GLM-5, and DeepSeek-V3.1. Adding a reasoning budget to
\texttt{gpt-5.4-mini} cuts hallucination sharply on P
(49.3\%\(\to\)7.3\%) and C (52.0\%\(\to\)13.0\%), but leaves substantial
long-horizon failures: M remains 75.0\% and X remains 36.0\%. Clearly, our probe categories allow us to separate failure modes 
rather than collapsing them into one aggregate score.

\textbf{Memory and compound probes as frontier stress tests.}
At P and C, the four strongest models are near-saturated: \texttt{gpt-5.5},
\texttt{o3}, \texttt{gpt-5.4}, and \texttt{o4-mini} all achieve hallucination
rates $<$ 10\%. The long-sequence categories separate them from weaker models, but
also reveal meaningful differences within the frontier group. \texttt{gpt-5.5}
solves both M and X at 0.0\%; \texttt{o3} remains near-saturated
(M\,2.1\%, X\,6.0\%); \texttt{gpt-5.4} is similarly strong
(M\,8.3\%, X\,4.0\%). By contrast, \texttt{o4-mini} solves the compound SAN
probe (X\,0.0\%) but is weaker on the pure memory-counting probe (M\,20.8\%),
while \texttt{o3-mini} has low P (1.3\%) but much higher C (39.0\%), M
(20.8\%), and X (28.0\%). Claude thinking models show the opposite: Opus 4.6 (T) and Sonnet 4.6 (T) are strong on M/X
(M\,2.1\%, X\,6.0\%; M\,8.3\%, X\,10.0\%) despite high causal hallucination (48.0\% and 44.0\%).



\textbf{Incorrect FEN metadata} tests robustness against corrupted auxiliary states. While overall model ordering is largely preserved, the perturbation exposes brittleness; even top models like \texttt{gpt-5.5} and \texttt{o3} show a 3-4\% increase in hallucination. The effect is more mixed for weaker
models: some improve slightly because the corrupted FEN is not the dominant failure source, while others remain dominated by M/X failures. The FEN ablation acts as a serializer-robustness probe, distinguishing models that reason from the board itself from those distracted by inconsistent textual states.


\chessname illustrates the diagnostic value of a structured probe taxonomy
over aggregate scores. The weakest non-reasoning models cluster tightly under No FEN (Overall 57-63\%) while the P/C/M/X decomposition immediately locates the dominant failures in
long-sequence state tracking and compound SAN reconstruction. Model-specific
anomalies only surface at the probe level. The
incorrect-FEN condition adds a robustness test, revealing whether
models over-trust corrupted auxiliary metadata even with a visible board. This validates the benchmark design: the ground truths are
algorithmically checkable, difficulty is tunable via game length, and the taxonomy reveals particular failure modes.

\section{\terminalname Results}
\label{sec:terminal-results}
 

\begin{table}[ht]
\centering
\caption{%
  \textbf{\terminalname hallucination rates by probe category.}
  (T)\,=\,extended thinking.
  Categories: C\,=\,Causal, X\,=\,Cross-tier compound, M\,=\,Memory,
  P\,=\,Perceptual, U\,=\,Uncertainty.
  Cell color reflects relative hallucination (green\,=\,low, red\,=\,high);
  \textbf{bold} marks column min.
}
\label{tab:terminal-results}
\small
\setlength{\tabcolsep}{5pt}
\begin{tabular}{lrrrrrr}
\toprule
\textbf{Model}
  & \textbf{Overall}
  & \textbf{C}
  & \textbf{X}
  & \textbf{M}
  & \textbf{P}
  & \textbf{U} \\
\midrule
GPT-5.5 (T)
  & \cellcolor{pastelgreen!92!pastelred}\textbf{ 5.9\%}
  & \cellcolor{pastelgreen!99!pastelred}\textbf{ 0.9\%}
  & \cellcolor{pastelgreen!97!pastelred}\textbf{ 1.9\%}
  & \cellcolor{pastelgreen!96!pastelred}\textbf{ 2.8\%}
  & \cellcolor{pastelgreen!99!pastelred}\textbf{ 0.9\%}
  & \cellcolor{pastelgreen!69!pastelred}\textbf{23.1\%} \\
o3 (T)
  & \cellcolor{pastelgreen!86!pastelred} 10.2\%
  & \cellcolor{pastelgreen!94!pastelred}  4.7\%
  & \cellcolor{pastelgreen!87!pastelred}  9.4\%
  & \cellcolor{pastelgreen!93!pastelred}  5.6\%
  & \cellcolor{pastelgreen!89!pastelred}  8.6\%
  & \cellcolor{pastelgreen!69!pastelred}\textbf{23.1\%} \\
o4-mini (T)
  & \cellcolor{pastelgreen!81!pastelred} 14.2\%
  & \cellcolor{pastelgreen!91!pastelred}  6.6\%
  & \cellcolor{pastelgreen!89!pastelred}  8.5\%
  & \cellcolor{pastelgreen!85!pastelred} 11.1\%
  & \cellcolor{pastelgreen!90!pastelred}  7.6\%
  & \cellcolor{pastelgreen!50!pastelred} 37.5\% \\
Claude Opus 4.6 (T)
  & \cellcolor{pastelgreen!79!pastelred} 15.7\%
  & \cellcolor{pastelgreen!86!pastelred} 10.4\%
  & \cellcolor{pastelgreen!87!pastelred}  9.4\%
  & \cellcolor{pastelgreen!86!pastelred} 10.2\%
  & \cellcolor{pastelgreen!87!pastelred}  9.5\%
  & \cellcolor{pastelgreen!48!pastelred} 39.4\% \\
Claude Sonnet 4.6 (T)
  & \cellcolor{pastelgreen!76!pastelred} 17.8\%
  & \cellcolor{pastelgreen!87!pastelred}  9.4\%
  & \cellcolor{pastelgreen!81!pastelred} 14.1\%
  & \cellcolor{pastelgreen!85!pastelred} 11.1\%
  & \cellcolor{pastelgreen!76!pastelred} 18.1\%
  & \cellcolor{pastelgreen!51!pastelred} 36.5\% \\
o3-mini (T)
  & \cellcolor{pastelgreen!72!pastelred} 21.0\%
  & \cellcolor{pastelgreen!81!pastelred} 14.1\%
  & \cellcolor{pastelgreen!79!pastelred} 16.0\%
  & \cellcolor{pastelgreen!81!pastelred} 13.9\%
  & \cellcolor{pastelgreen!81!pastelred} 14.3\%
  & \cellcolor{pastelgreen!37!pastelred} 47.1\% \\
GLM-5
  & \cellcolor{pastelgreen!65!pastelred} 26.3\%
  & \cellcolor{pastelgreen!72!pastelred} 20.8\%
  & \cellcolor{pastelgreen!71!pastelred} 21.7\%
  & \cellcolor{pastelgreen!72!pastelred} 21.3\%
  & \cellcolor{pastelgreen!54!pastelred} 34.3\%
  & \cellcolor{pastelgreen!55!pastelred} 33.7\% \\
GPT-4o
  & \cellcolor{pastelgreen!49!pastelred} 38.4\%
  & \cellcolor{pastelgreen!52!pastelred} 35.9\%
  & \cellcolor{pastelgreen!47!pastelred} 39.6\%
  & \cellcolor{pastelgreen!54!pastelred} 34.3\%
  & \cellcolor{pastelgreen!53!pastelred} 35.2\%
  & \cellcolor{pastelgreen!37!pastelred} 47.1\% \\
DeepSeek V3.1
  & \cellcolor{pastelgreen!39!pastelred} 45.6\%
  & \cellcolor{pastelgreen!35!pastelred} 49.1\%
  & \cellcolor{pastelgreen!40!pastelred} 45.3\%
  & \cellcolor{pastelgreen!54!pastelred} 34.3\%
  & \cellcolor{pastelgreen!42!pastelred} 43.8\%
  & \cellcolor{pastelgreen!26!pastelred} 55.8\% \\
GPT-5.4-mini
  & \cellcolor{pastelgreen!33!pastelred} 50.3\%
  & \cellcolor{pastelgreen!43!pastelred} 42.4\%
  & \cellcolor{pastelgreen!38!pastelred} 46.2\%
  & \cellcolor{pastelgreen!42!pastelred} 43.5\%
  & \cellcolor{pastelgreen!26!pastelred} 55.2\%
  & \cellcolor{pastelgreen!14!pastelred} 64.4\% \\
Qwen3-30B
  & \cellcolor{pastelgreen!32!pastelred} 51.0\%
  & \cellcolor{pastelgreen!45!pastelred} 41.5\%
  & \cellcolor{pastelgreen!41!pastelred} 44.3\%
  & \cellcolor{pastelgreen!28!pastelred} 53.7\%
  & \cellcolor{pastelgreen!30!pastelred} 52.4\%
  & \cellcolor{pastelgreen!15!pastelred} 63.5\% \\
GPT-4o-mini
  & \cellcolor{pastelgreen!31!pastelred} 51.8\%
  & \cellcolor{pastelgreen!47!pastelred} 39.6\%
  & \cellcolor{pastelgreen!38!pastelred} 46.2\%
  & \cellcolor{pastelgreen!24!pastelred} 57.4\%
  & \cellcolor{pastelgreen!18!pastelred} 61.9\%
  & \cellcolor{pastelgreen!28!pastelred} 53.8\% \\
Kimi-K2.6
  & \cellcolor{pastelgreen!25!pastelred} 56.5\%
  & \cellcolor{pastelgreen!25!pastelred} 56.6\%
  & \cellcolor{pastelgreen!37!pastelred} 47.2\%
  & \cellcolor{pastelgreen!35!pastelred} 49.1\%
  & \cellcolor{pastelgreen!24!pastelred} 57.1\%
  & \cellcolor{pastelgreen!3!pastelred}  73.1\% \\
\bottomrule
\end{tabular}
\end{table}
 
\textbf{Overall accuracy.}
The terminal domain is the most difficult in \textsc{HalluWorld}: even GPT-5.5
achieves only 94.1\% overall accuracy, and no model saturates any individual probe
category.
Two tiers emerge, mirroring the gridworld finding: thinking/reasoning models achieve 5.9--21.0\% hallucination rates overall, while standard models cluster at 26.3--56.5\%.
As in the gridworld evaluation, this gap partly reflects a compute budget disparity:
thinking models were given \texttt{max\_output\_tokens=16\,384} while standard models
were capped at 256.
The performance difference should therefore be interpreted as an upper bound on the
underlying capability difference, not a controlled comparison.
 
\textbf{Probe category hierarchy.}
A consistent ordering holds across most models (by accuracy):
$\text{Perceptual} \approx \text{Causal} > \text{Cross-Tier} \approx \text{Memory} \gg \text{Uncertainty}$.
Perceptual and causal accuracy are near-ceiling for the strongest models - correctly
reading a value directly present in a 50k-character terminal
context is tractable once long-context retrieval is reliable.
Uncertainty probes are the hardest: even GPT-5.5 and o3 cap at 76.9\%
accuracy, meaning the best models hallucinate $\approx$ 1/4 answers when the correct response is \textit{``cannot determine''}.
This contrasts with the controlled gridworld U-category, where uncertainty is often easier, and suggests that epistemic abstention becomes substantially harder in realistic terminal contexts.
 
\textbf{Notable anomalies.}
GPT-4o-mini shows a unique inverted memory profile: its memory hallucination rate (57.4\%) substantially exceeds its causal rate
(39.6\%), suggesting a specific failure in long-horizon state tracking. 
Kimi-K2.6 exhibits the highest uncertainty hallucination in the terminal domain
(73.1\%).
Since U-probes specifically target the \textit{uncertainty overclaim} failure mode, this shows Kimi's high tendency towards confident confabulation under epistemic limits here.


 

\textbf{Thinking depth ablation.}
The thinking paradox identified in \gridname\ (\S\ref{sec:thinking})
has a mixed instantiation in \terminalname. Looking at \autoref{tab:terminal-ablation}, for Sonnet~4.6, accuracy increases monotonically from 79.2\% (no thinking)
to 86.2\% (max), with every probe category benefiting from deeper reasoning.
GPT-5.5 shows a different pattern: any reasoning budget yields a large jump
($+$8.5~pp, 86.4\%~$\to$~94.9\%), but the gains are non-monotonic:
\texttt{medium} is slightly below \texttt{low}, \texttt{high} is best, and \texttt{xhigh} slightly regresses relative to \texttt{high}. GPT-5.5 shows a non-monotonic pattern on uncertainty: \texttt{low}, \texttt{medium}, and \texttt{xhigh} reasoning
underperform the \texttt{no}-reasoning setting, while \texttt{high} reasoning reaches the best U accuracy
(84.3\%). 
This suggests that, for the most part, reasoning interferes
specifically with epistemic abstention rather than with factual retrieval or causal
inference.
Uncertainty remains the hardest category regardless of reasoning depth for both
models, with all configurations plateauing in the 60--85\% range.
Full ablation results are in \S\ref{app:terminal-ablation}. Additional analysis of hallucination rate as a function of trajectory depth is provided in \autoref{app:terminal-hallu-depth}.

\section{Conclusion \& Future Work}
\label{sec:discussion}
Our benchmark, \benchmarkname, demonstrates that hallucination is fundamentally a world-modeling failure that decomposes into distinct categories (perception, memory, causality, uncertainty), many of which remain unsolved for frontier models. By grounding evaluation in explicit, controllable environments, \benchmarkname enables reproducible and fine-grained diagnosis of when and why hallucinations occur. \benchmarkname turns hallucination from a vague metric into an experimentally tractable problem, supporting targeted interventions and deeper analysis beyond aggregate scores.

There are several directions for future work. First, our current probing measures explicit or stated beliefs, but models may have a wider distribution in their internal representations \citep{shrivastava2025linearly}. Mechanistic techniques such as causal tracing \citep{prakashlanguage,meng2022locating} could provide one lens to observe these internal beliefs. 
Second, current probes use canonical locations co-designed with the constructed environment as locations where cognitive load is likely higher. 
More general formulation of probe placement 
could use active learning approaches \citep{xia2024hallucination} to dynamically discover states in a given environment where the agent would face maximal cognitive load. 
Further, one could more explicitly incorporate and assess multi-source conflict resolution. Lastly, it would be interesting to investigate hallucinations in small model and data limited regimes, e.g., at the BabyLM-scale \citep{warstadt2023findings,hu2024findings,feng2026babyscaleinvestigatingmodels,long2024babyview,feng-etal-2024-child,hu2025production,zeng2026bringingbilingualbabylminvestigating}.

\section*{Acknowledgments}
We gratefully acknowledge Modal, the National Science Foundation ACCESS Program, Lambda Labs' Research Grant Program, and NVIDIA's Academic Grant Program for providing compute resources that enabled this work. EL was supported by the National Sciences and Engineering Research Council of Canada (NSERC), [funding reference number 578085], as well as the SoftBank-ARM Fellowship.\\

\bibliographystyle{plainnat}
\bibliography{references}


\newpage

\appendix

\section{\benchmarkname Qualitative Examples}
\label{appendix:qualitative_analysis_illustration}

\subsection{\gridname}
\label{appendix:halluworldgrid_qualitative_analysis_illustration}


\newcommand{\cmark}{\textcolor{green!70!black}{\ding{51}}} 
\newcommand{\xmark}{\textcolor{red!70!black}{\ding{55}}}    

\begin{table*}[!ht]
    \small
    \centering
    \setlength{\tabcolsep}{6pt}
    \begin{tabular}{lp{0.82\textwidth}}
    \toprule
    \textbf{Component} & \textbf{Details} \\ 
    \midrule

    \textbf{World:} P1 Dense Array & 
    \textbf{Grid View (Canonical Location 1):} \\
    \textbf{Serializer:} Grid & 
    {\ttfamily\footnotesize
    \begin{tabular}{@{}l llllllll lllll@{}}
               & L6 & L5 & L4 & L3 & L2 & L1 & 0  & R1 & R2 & R3 & R4 & R5 & R6 \\
    ahead 0 & ?? & \#\# & .. & .. & .. & .. & @. & .. & .. & .. & .. & .. & .. \\
    ahead 1 & ?? & \#\# & bB & bB & bB & bB & bB & bB & bB & bB & gB & .. & .. \\
    ahead 2 & ?? & \#\# & bB & rB & bB & bB & bB & bB & bB & bB & bB & .. & .. \\
    ahead 3 & ?? & \#\# & bB & bB & bB & bB & bB & bB & bB & bB & bB & .. & .. \\
    ahead 4 & ?? & \#\# & .. & .. & .. & .. & .. & .. & .. & .. & .. & .. & .. \\
    ahead 5 & ?? & \#\# & rK & rK & rK & .. & rK & rK & rK & rK & rK & .. & .. \\
    ahead 6 & ?? & \#\# & rK & rK & rK & rK & rK & rK & rK & bK & rK & .. & .. \\
    ahead 7 & ?? & \#\# & rK & rK & rK & rK & rK & rK & rK & rK & rK & .. & .. \\
    \end{tabular}} \\
    \textbf{Agent:} Facing North & \textbf{Legend:} K=key, B=ball, D=door, @=agent, \#=wall, .=floor \\
    \textbf{Mode:} CtrlStatic & \textbf{Colors:} r=red, b=blue, g=green, y=yellow \\
    \midrule

    \textbf{Probes (4 total)} & 
    \textbf{Probe 1 (Presence):} ``Is there a green ball in your current field of view?'' \\
    & \quad $\rightarrow$ \textbf{Ground Truth:} \texttt{yes} (green ball at R4, ahead 1) \\[0.3em]
    & \textbf{Probe 2 (Count):} ``How many blue balls do you see?'' \\
    & \quad $\rightarrow$ \textbf{Ground Truth:} \texttt{14} \\[0.3em]
    & \textbf{Probe 3 (Attribute):} ``What color is the ball at R1, ahead 2?'' \\
    & \quad $\rightarrow$ \textbf{Ground Truth:} \texttt{red} \\[0.3em]
    & \textbf{Probe 4 (Location):} ``Where is the blue key relative to you?'' \\
    & \quad $\rightarrow$ \textbf{Ground Truth:} \texttt{\{"steps\_ahead": 6, "lateral": 3\}} \\
    \midrule

    \textbf{Model Responses} & 
    \textbf{GPT-4o:} \cmark{} P1: ``yes'' | \xmark{} P2: ``15'' (overcount) | \xmark{} P3: ``blue'' | \cmark{} P4: Correct \\[0.3em]
    & \textbf{GPT-5.5:} \cmark{} P1: ``yes'' | \cmark{} P2: ``14'' | \cmark{} P3: ``red'' | \xmark{} P4: (off by 1 step) \\[0.3em]
    & \textbf{o3-mini:} \cmark{} P1: ``yes'' | \xmark{} P2: ``13'' (undercount) | \xmark{} P3: ``blue'' | \cmark{} P4: Correct \\[0.3em]
    & \textbf{gpt-4o-mini:} \xmark{} P1: ``no'' | \xmark{} P2: ``10'' | \cmark{} P3: ``red'' | \xmark{} P4: (lateral wrong) \\
    \bottomrule
    \end{tabular}
    \caption{\textbf{\gridname Perception Probes (P1 Dense Array):} Example showing world state (Grid serializer), four probe types, and model responses. \xmark~indicates hallucination, \cmark~indicates correct answer.}
    \label{tab:qual-example-p1}
\end{table*}

\begin{table*}[!ht]
    \small
    \centering
    \setlength{\tabcolsep}{6pt}
    \begin{tabular}{lp{0.82\textwidth}}
    \toprule
    \textbf{Component} & \textbf{Details} \\ 
    \midrule
    \textbf{World:} C1a Persistent & \textbf{Memory Serialization (Timestep 8):} \\
    \textbf{Serializer:} Memory & \textit{``You started in a room with a grey boulder. You moved it north, revealing a grey key underneath. You picked up the grey key. You moved 3 steps north through a corridor. You reached a locked grey door. You toggled it open with your grey key. You entered a new room with a red ball visible to the east.''}  \\[0.5em]
    \textbf{Mode:} InNav & \textbf{Current FOV:} \texttt{[Ahead: floor | Left: wall | Right: red ball (rB) | Behind: grey open door]} \\
    \midrule

    \textbf{Probes (3 total)} & \textbf{Probe 1 (Persistence):} ``What was hidden under the boulder at the start?'' $\rightarrow$ \texttt{grey key} \\
    & \textbf{Probe 2 (Presence):} ``Is there a red ball visible?'' $\rightarrow$ \texttt{yes} \\
    & \textbf{Probe 3 (Causal):} ``Why were you able to open the grey door?'' $\rightarrow$ \texttt{had grey key} \\
    \midrule

    \textbf{Model Responses} & \textbf{GPT-4o:} P1 (Mem): \cmark{} | P2 (Percep): \xmark~InNav ``no'' | P3 (Causal): \xmark~InNav ``pushed boulder'' \\
    \textbf{(InNav vs CtrlStatic)} & \textbf{GPT-5.5:} P1 (Mem): \cmark{} | P2 (Percep): \cmark{} | P3 (Causal): \cmark{} \\
    & \textbf{o3-mini:} P1 (Mem): \cmark{} | P2 (Percep): \xmark~InNav ``no'' | P3 (Causal): \xmark~InNav ``door was unlocked'' \\
    \bottomrule
    \end{tabular}
    \caption{\textbf{\gridname Memory/Causal Probes (C1a Persistent Chain):} InNav mode navigation amplifies hallucinations in GPT-4o and o3-mini (demonstrating \textbf{cognitive load}), whereas GPT-5.5 maintains accuracy (\textbf{epistemic grounding}).}
    \label{tab:qual-example-c1a}
\end{table*}

\subsection{\chessname}


\begin{table*}[!ht]
    \small
    \centering
    \setlength{\tabcolsep}{6pt}
    \begin{tabular}{lp{0.7\textwidth}}
    \toprule
    \textbf{Component} & \textbf{Details} \\
    \midrule

    \textbf{World:} Chess position &
    \textbf{Board View (no FEN shown):} \\
    \textbf{Serializer:} Board grid &
    {\fontfamily{fvm}\selectfont\footnotesize
    \begin{tabular}{@{}lcccccccc@{}}
    8 & . & n & . & . & . & r & k & . \\
    7 & . & q & . & . & . & . & . & . \\
    6 & . & . & . & p & P & . & p & . \\
    5 & . & p & p & . & . & Q & . & p \\
    4 & . & . & . & . & . & . & . & . \\
    3 & . & . & . & . & B & . & b & K \\
    2 & P & . & . & . & P & P & . & P \\
    1 & R & . & . & . & . & R & . & . \\
      & a & b & c & d & e & f & g & h
    \end{tabular}} \\
    \textbf{Side to move:} White &
    \textbf{Legend:} standard chess coordinates; ``.'' denotes an empty square, lowercase denotes a black piece, and uppercase a white piece. \\
    \midrule

    \textbf{Probes (5 total)} &
    \textbf{P / Capture:} ``Can white legally capture the black pawn on b5 on the next move?'' \\
    & \quad $\rightarrow$ \textbf{Ground Truth:} \texttt{no} \\[0.3em]
    & \textbf{P / Defense:} ``The black pawn on c5 is attacked by the opponent. Is it also defended by another piece of its own side?'' \\
    & \quad $\rightarrow$ \textbf{Ground Truth:} \texttt{yes} \\[0.3em]
    & \textbf{P / Hanging:} ``Is the black pawn on d6 hanging?'' \\
    & \quad $\rightarrow$ \textbf{Ground Truth:} \texttt{no} \\[0.3em]
    & \textbf{C / Hypothetical Check:} ``If white plays \texttt{f5f6}, would the side to move be in check?'' \\
    & \quad $\rightarrow$ \textbf{Ground Truth:} \texttt{no} \\[0.3em]
    & \textbf{C / Post-rollout Count:} ``If white plays \texttt{a1c1} and black plays \texttt{b7a6}, how many white pieces lost all friendly defenders?'' \\
    & \quad $\rightarrow$ \textbf{Ground Truth:} \texttt{1} \\
    \midrule

    \textbf{Qualitative takeaway} &
    These probes isolate board-grounded hallucination from long-context memory. The P probes require static relational reading of the visible piece graph, while the C probes require applying one or two legal moves and recomputing a derived property. \\
    \bottomrule
    \end{tabular}
    \caption{\textbf{\chessname Perceptual/Causal Probes:} Example board used for static board-reading probes and short-horizon causal rollout. Ground truth is computed directly from the chess engine.}
    \label{tab:qual-example-chess-board}
\end{table*}

\newpage

\subsection{\terminalname}

\begingroup
    \small
    \setlength{\tabcolsep}{6pt}
    \begin{longtable}{@{}p{0.15\textwidth}p{0.80\textwidth}@{}}
    \caption{\textbf{\terminalname Runtime Probes (3d-model-format-legacy):}
    Example showing a late compile loop where the model must reconcile terminal
    output, command history, filesystem snapshots, and an inconsistent working
    directory signal.}
    \label{tab:qual-example-terminal-3d-model}\\
    \toprule
    \textbf{Component} & \textbf{Details} \\
    \midrule
    \endfirsthead

    \toprule
    \textbf{Component} & \textbf{Details} \\
    \midrule
    \endhead

    \midrule
    \multicolumn{2}{r}{\footnotesize Continued on next page} \\
    \endfoot

    \bottomrule
    \endlastfoot

    \textbf{Task:} 3d-model-format-legacy &
    \textbf{Runtime Terminal Evidence (step 0258):} \\
    \textbf{Serializer:} Terminal trace + path snapshots &
    {\ttfamily\footnotesize
    \setlength{\tabcolsep}{3pt}
    \begin{tabular}{@{}p{0.08\linewidth}@{\hspace{4pt}}p{0.30\linewidth}@{\hspace{4pt}}p{0.22\linewidth}@{\hspace{4pt}}p{0.28\linewidth}@{}}
    Cmd & Operation & /app/MdfLib mtime & listing\_sha256 \\
    0249 & timeline memset edit & 1777902356 & 661d... \\
    0250 & header memset edit & 1777902356 & 48dd... \\
    0251 & delete windows.h from stdafx.cpp & 1777902356 & 48dd... \\
    0252 & g++ compile & 1777902356 & 48dd... \\
    0253 & header memset edit & 1777902371 & e0cf... \\
    0254 & mesh memset edit & 1777902372 & e0cf... \\
    0255 & timeline memset edit & 1777902372 & e0cf... \\
    0256 & header memset edit & 1777902372 & 3dfc... \\
    0257 & delete windows.h from stdafx.cpp & 1777902373 & 3dfc... \\
    0258 & g++ compile & 1777902373 & 3dfc... \\
    \end{tabular}} \\[0.4em]
    \textbf{Trigger Command:} &
    \texttt{g++ -o MdfFormat MdfFormat.cpp stdafx.cpp -I.} \\
    \textbf{Trace Signal:} &
    Compile output still reports \texttt{MdfFormat.cpp:543:55} and
    \texttt{stdafx.h:11:10: fatal error: windows.h}; captured prompt shows
    \texttt{/app/MdfLib\#}, while relative snapshots for \texttt{MdfFormat.cpp}
    and \texttt{stdafx.cpp} under \texttt{/app} are nonexistent. \\
    \textbf{Mode:} Runtime probe & \textbf{Scope:} through\_trigger \\
    \midrule

    \textbf{Probes (5 total)} &
    \textbf{Probe 1 (Perceptual):} ``Which command indices changed
    \texttt{listing\_sha256} while leaving \texttt{mtime\_epoch} unchanged?''
    $\rightarrow$ \texttt{idx=0250,0256} \\[0.3em]
    & \textbf{Probe 2 (Memory):} ``Which earlier reported compile issue was
    never directly targeted by any \texttt{sed} pattern in commands 0219--0257?''
    $\rightarrow$ \texttt{untouched=line543\_cast} \\[0.3em]
    & \textbf{Probe 3 (Causal):} ``If command 0257 deleted
    \texttt{\#include <windows.h>} from \texttt{stdafx.h} instead of
    \texttt{stdafx.cpp}, what would happen to that fatal diagnostic?''
    $\rightarrow$ \texttt{diag=disappear} \\[0.3em]
    & \textbf{Probe 4 (Uncertainty):} ``Were the relative snapshots
    \texttt{MdfFormat.cpp} and \texttt{stdafx.cpp} taken in the same directory
    context that the compile commands used?''
    $\rightarrow$ \texttt{status=contradicted} \\[0.3em]
    & \textbf{Probe 5 (Cross-Category):} ``Which recurring edit type is associated
    with every observed \texttt{/app/MdfLib} listing-hash change and unrelated
    to the \texttt{windows.h} include chain?''
    $\rightarrow$ \texttt{edit=header\_memset\_fix} \\
    \midrule

    \textbf{Evaluation Note} &
    The extracted probe JSONs store grounded questions and gold answers, but
    this checkout does not include per-probe \texttt{probe\_eval} model-answer
    records. Aggregate results over all 529 probes are available separately:
    GPT-5.5 94.1\%, o3 89.8\%, GPT-4o 61.6\%, GPT-4o-mini 48.2\%. \\
    \end{longtable}
\endgroup

\begingroup
    \small
    \setlength{\tabcolsep}{6pt}
    \begin{longtable}{@{}p{0.15\textwidth}p{0.80\textwidth}@{}}
    \caption{\textbf{\terminalname Memory/Causal Probes (hf-train-lora-adapter):}
    Example showing how terminal traces distinguish syntax repair, API repair,
    stale metadata, and unsupported claims about training.}
    \label{tab:qual-example-terminal-lora}\\
    \toprule
    \textbf{Component} & \textbf{Details} \\
    \midrule
    \endfirsthead

    \toprule
    \textbf{Component} & \textbf{Details} \\
    \midrule
    \endhead

    \midrule
    \multicolumn{2}{r}{\footnotesize Continued on next page} \\
    \endfoot

    \bottomrule
    \endlastfoot

    \textbf{Task:} hf-train-lora-adapter &
    \textbf{Runtime Terminal Evidence (step 0008):} \\
    \textbf{Serializer:} Terminal trace + file snapshots &
    {\ttfamily\footnotesize
    \setlength{\tabcolsep}{3pt}
    \begin{tabular}{@{}p{0.08\linewidth}@{\hspace{4pt}}p{0.29\linewidth}@{\hspace{4pt}}p{0.30\linewidth}@{\hspace{4pt}}p{0.23\linewidth}@{}}
    Cmd & Event & train\_lora.py state & Output artifact state \\
    0002 & echo script with literal \textbackslash n & size 1094, alpha=16 & none \\
    0003 & pip install peft==0.15.0 & unchanged & none \\
    0004 & python /app/train\_lora.py & SyntaxError & none \\
    0005 & sed converts literal \textbackslash n & size 1060 & none \\
    0006 & python /app/train\_lora.py & TypeError: unexpected alpha & none \\
    0008 & remove alpha and rerun & size 1051, no alpha in config & adapter + results JSON \\
    \end{tabular}} \\[0.4em]
    \textbf{Trigger Command:} &
    \texttt{python /app/train\_lora.py} \\
    \textbf{Final Source Signal:} &
    \texttt{LoraConfig(...)} no longer contains \texttt{alpha}, but
    \texttt{results['lora\_config']} still records \texttt{\{'r': 8, 'alpha': 16,
    'epochs': 3\}}. The script loads train/val JSON only to count samples, then
    saves the adapter without an optimizer, backward pass, or training loop. \\
    \textbf{Mode:} Runtime probe & \textbf{Scope:} through\_trigger \\
    \midrule

    \textbf{Probes (5 total)} &
    \textbf{Probe 1 (Perceptual):} ``Compare the executable
    \texttt{LoraConfig(...)} call with the later metadata. Which mismatch class
    applies?'' $\rightarrow$ \texttt{mismatch=metadata\_extra\_alpha} \\[0.3em]
    & \textbf{Probe 2 (Memory):} ``Immediately before command 8 began, how had
    \texttt{/app/train\_lora.py} changed relative to the original file?''
    $\rightarrow$ \texttt{delta=newline\_and\_alpha\_fix} \\[0.3em]
    & \textbf{Probe 3 (Causal):} ``If after command 5 the code had replaced
    \texttt{alpha=16} with \texttt{lora\_alpha=16}, what would the next run most
    likely have done?'' $\rightarrow$ \texttt{outcome=would\_succeed} \\[0.3em]
    & \textbf{Probe 4 (Uncertainty):} ``Do final adapter weights contain
    gradient-based training updates from \texttt{/app/data/train.json}?''
    $\rightarrow$ \texttt{status=contradicted} \\[0.3em]
    & \textbf{Probe 5 (Cross-Category):} ``Which created output is known to reflect
    dataset contents rather than only model/config defaults?''
    $\rightarrow$ \texttt{depends\_on\_dataset=results\_only} \\
    \midrule

    \textbf{Evaluation Note} &
    This example targets hallucinations that confuse successful artifact creation
    with real training, or conflate stale metadata with executable configuration.
    Per-probe model responses can be appended from \texttt{runs/probe\_eval}
    result JSONs when those artifacts are restored. \\
    \end{longtable}
\endgroup

\section{\chessname Probe Types}
\label{appendix:chess_all_probes}
\begin{table}[H]
\centering
\small
\caption{All \chessname\ probes. Cat.\ = probe category
(P\,=\,Perceptual, C\,=\,Causal, M\,=\,Memory).
Board = whether the ASCII board is shown; History = whether a move
sequence is provided. All probes scored 0/1; 50 episodes each.}
\label{tab:chess-probes}
\resizebox{\textwidth}{!}{
\begin{tabular}{llccl}
\toprule
\textbf{Probe} & \textbf{Cat.} & \textbf{Board} & \textbf{History}
  & \textbf{Core challenge} \\
\midrule
\texttt{chess\_can\_capture}
  & P & \checkmark & --
  & Legal capture detection; negatives biased toward pinned-piece pseudo-legals \\
\texttt{chess\_defended}
  & P & \checkmark & --
  & Defender detection on attacked squares only; ``yes'' answers load-bearing \\
\texttt{chess\_hanging}
  & P & \checkmark & --
  & Attacked-and-undefended detection; kings excluded \\
\midrule
\texttt{chess\_hypothetical\_in\_check}
  & C & \checkmark & --
  & Apply one specified move; evaluate check on successor state \\
\texttt{chess\_after\_move\_undefended\_count}
  & C & \checkmark & --
  & Count pieces losing all defenders after 1--2 hypothetical moves \\
\midrule
\texttt{chess\_hidden\_side\_capture\_stats}
  & M & -- & \checkmark
  & Maintain running capture count over 25--88 moves; no board shown \\
\texttt{chess\_san\_legal\_move}
  & M & -- & \checkmark
  & Produce a legal move after 40+ move prefix; terminal position
    guarantees \\
  & & & & few qualifying moves \\
\bottomrule
\end{tabular}
}
\end{table}

\begin{table*}[!h]
    \small
    \centering
    \setlength{\tabcolsep}{6pt}
    \begin{tabular}{lp{0.68\textwidth}}
    \toprule
    \textbf{Component} & \textbf{Details} \\
    \midrule

    \textbf{Probe family} & \textbf{Long-horizon move tracking without an observational shortcut} \\
    \textbf{Serializer} & Move-history text; the model must reconstruct or track the hidden board state from a sequence of legal chess moves. \\
    \midrule

    \textbf{M Probe: Hidden Capture Stats} &
    \textbf{Observation excerpt:} \\
    & {\ttfamily\footnotesize
    Moves played (UCI with SAN): e3c5 (Bxc5), g3f2 (Bxf2), f5h5 (Qxh5), g6h5 (gxh5), f1f2 (Rxf2), d6c5 (dxc5), f2f8 (Rxf8+), \ldots, d4f3 (Nf3)} \\[0.4em]
    & \textbf{Question:} ``How many black pieces were removed by captures during the sequence above?'' \\
    & \quad $\rightarrow$ \textbf{Ground Truth:} \texttt{6} \\
    \midrule

    \textbf{X Probe: SAN Legal Move} &
    \textbf{Observation excerpt:} \\
    & {\ttfamily\footnotesize
    Moves played: 1. b3 Nh6 2. e3 g6 3. a4 e5 4. Na3 Na6 5. Bxa6 bxa6 \ldots 52. Kf2 Kh6 53. Ke3 Re8+ 54. Kd2 Re4} \\[0.4em]
    & \textbf{Question:} ``From this position, reply with exactly one legal move for the side to move, in UCI format.'' \\
    & \quad $\rightarrow$ \textbf{Ground Truth:} \texttt{\{d2c1, d2c2, d2c3, d2d1, d2d3\}} \\
    \midrule

    \textbf{Qualitative takeaway} &
    The memory probe requires counting events over a long explicit trajectory. The SAN probe is compound: it requires reconstructing an implicit board from Standard Algebraic Notation, applying chess transition rules across many turns, and then producing a legal continuation. This explains why the chess benchmark remains discriminative after simple board-reading saturates: the compound probe tests whether a model maintains a faithful internal world state. \\
    \bottomrule
    \end{tabular}
    \caption{\textbf{\chessname Memory and Compound Probes:} Examples of long-horizon chess probes. The M probe tests event tracking over a move sequence; the X probe combines memory, causal transition rules, and valid action generation from SAN-only input.}
    \label{tab:qual-example-chess-memory-compound}
\end{table*}

\section{List of all \gridname Levels}
\label{appendix:gridworld-all-levels}
{\small
\begin{longtable}{@{}lp{2.8cm}p{1.3cm}p{1.1cm}p{5.4cm}c@{}}
\caption{%
  All 33 \gridname levels.
  \textbf{Probes:} Pr\,=\,Presence, Ct\,=\,Count, St\,=\,State/Attribute,
  Lo\,=\,Location, Ca\,=\,Causal, Un\,=\,Uncertainty.
  \textbf{Ser.:} Sym\,=\,Symbolic, Grd\,=\,Grid, Mem\,=\,Memory, Mix\,=\,Mixed.
  $\bullet$\,=\,included in HalluWorld-Hard subset.
}
\label{tab:level-reference} \\
\toprule
\textbf{ID} & \textbf{Name} & \textbf{Probes} & \textbf{Ser.} & \textbf{Core challenge} & \textbf{Hard} \\
\midrule
\endfirsthead
\multicolumn{6}{l}{\small\textit{(continued)}} \\
\toprule
\textbf{ID} & \textbf{Name} & \textbf{Probes} & \textbf{Ser.} & \textbf{Core challenge} & \textbf{Hard} \\
\midrule
\endhead
\midrule
\multicolumn{6}{r}{\small\textit{continued on next page}} \\
\endfoot
\bottomrule
\endlastfoot

\multicolumn{6}{l}{\textit{Perceptual (P)}} \\
\midrule
P1      & Dense Array          & Pr, Ct, St & Sym, Grd
  & Dense regular grid with deliberate violations (color swap, absence, state mismatch); tests pattern-completion failure
  & Grd\,$\bullet$ \\[2pt]
P2      & Corridor Gauntlet    & Pr, Ct, Lo & Sym, Grd
  & Narrow corridor; objects at varying depths and lateral offsets; tests spatial ordering precision
  & Grd\,$\bullet$ \\[2pt]
P3      & Rotation Challenge   & Lo         & Sym, Grd
  & Fixed scene, random agent spawn direction; tests egocentric vs.\ allocentric frame conversion
  & --- \\[2pt]
P4a     & Harder Array         & Pr, Ct, St & Sym, Grd
  & Denser, larger P1 variant with more object types; tests compound counting under density
  & Grd\,$\bullet$ \\[2pt]
P4b     & Delta Perception     & Pr, Ct, St & Sym, Grd
  & Two observations separated by agent movement; tests FOV-shift appearance and disappearance detection
  & --- \\[2pt]
P5      & Object Permanence    & Pr, Ct, St & Sym, Grd
  & Object moved out of FOV between observations; tests out-of-view state tracking
  & --- \\
\midrule

\multicolumn{6}{l}{\textit{Memory (M)}} \\
\midrule
M1\,(N=3) & River Field (early)  & Pr, Ct, St & Mem
  & River carries objects east at fixed rate; stale notice board shows t=0 state; 3 steps elapsed
  & --- \\[2pt]
M1\,(N=6) & River Field (mid)    & Pr, Ct, St & Mem
  & Same as M1 at 6 steps; object positions maximally ambiguous
  & $\bullet$ \\[2pt]
M1\,(N=9) & River Field (late)   & Pr, Ct, St & Mem
  & Same as M1 at 9 steps; one object has dried, testing temporal decay tracking
  & $\bullet$ \\[2pt]
M2      & Witness Stand        & Pr, Ct, St & Sym, Grd
  & Five sequential chamber observations; tests cross-observation source attribution and recency interference
  & --- \\[2pt]
M3      & Incident Report      & Pr, Ct, St & Sym, Grd
  & Two room snapshots with 3 silent changes (color swap, state change, removal); tests change enumeration
  & --- \\[2pt]
M4      & Unreliable Narrator  & Pr, Ct, St & Mem
  & Signpost makes 3 false claims about the room alongside direct observation; tests sign vs.\ observation conflict
  & --- \\
\midrule

\multicolumn{6}{l}{\textit{Causal (C)}} \\
\midrule
C1a         & Persistent Chain         & Pr, Ca & Mem
  & Three-step chain; all effects \textbf{persistent} (one-shot triggers); notice board states rule; tests continuous-bias error
  & $\bullet$ \\[2pt]
C1a\,(nb)   & Persistent Chain (no board) & Pr, Ca & Mem
  & C1a without notice board; model must infer persistence from mechanics alone
  & $\bullet$ \\[2pt]
C1b         & Continuous Chain         & Pr, Ca & Mem
  & Identical layout to C1a; all effects \textbf{continuous} (gates re-lock on key removal); tests persistent-bias error
  & $\bullet$ \\[2pt]
C1b\,(nb)   & Continuous Chain (no board) & Pr, Ca & Mem
  & C1b without notice board; model must infer continuity from mechanics alone
  & $\bullet$ \\[2pt]
C2          & Fire Crossing            & Pr, St, Ca & Mem
  & Wet ball extinguishes fire barrier; dry decoy ball cannot; tests wet-condition state tracking under time pressure
  & --- \\[2pt]
C3          & Flood Room               & Pr, St, Ca & Mem
  & Floodtiles advance one row per step; agent must reach elevated goal before path is cut off; tests step-precise forward simulation
  & --- \\[2pt]
C4          & Forking Paths            & Pr, Ca & Mem
  & Three paths to goal: iron key (valid), lit torch on wood door (valid), water crossing (red herring—no boat); tests solution-space exploration
  & --- \\[2pt]
C5a         & Adversarial Board        & Pr, Ca & Mem
  & C1a layout; notice board \textbf{lies} (claims gate is continuous); annotation is correct; pits adversarial testimony against structural inference
  & $\bullet$ \\[2pt]
C6          & Flood--Fire Escape       & Pr, St, Ca & Mem
  & Flood extinguishes fire at step 4 creating a timed action window; lying notice board gives wrong timings; tests cross-mechanic composition with adversarial testimony
  & $\bullet$ \\
\midrule

\multicolumn{6}{l}{\textit{Uncertainty (U)}} \\
\midrule
U1          & Fog of War               & Pr, Un     & Sym
  & Hub with four never-observed sealed rooms described on notice board; tests epistemic abstention vs.\ confident confabulation
  & --- \\[2pt]
U2\,(high)  & Oracle Problem (80\%)    & Un         & Mix
  & Two signposts claim room contents; notice board states 80\% signpost accuracy; tests confidence scaling to stated reliability
  & --- \\[2pt]
U2\,(mid)   & Oracle Problem (50\%)    & Un         & Mix
  & Same setup at 50\% stated accuracy
  & --- \\[2pt]
U2\,(low)   & Oracle Problem (20\%)    & Un         & Mix
  & Same setup at 20\% stated accuracy
  & --- \\[2pt]
U4          & The Amnesiac             & St, Ca, Un & Sym
  & Room with evidence of prior activity but no stored observations; tests deducible vs.\ genuinely unanswerable causal attribution
  & --- \\
\midrule

\multicolumn{6}{l}{\textit{Cross-Category Compound (X)}} \\
\midrule
X1          & Facility Tour            & Pr, Ct, St & Sym
  & 3-zone facility tour; baseline multi-room test of cross-zone intrusion
  & --- \\[2pt]
X2          & Facility Tour+           & Pr, Ct, St & Sym, Grd
  & 5-zone tour; lying signpost claims incorrect details about a prior zone
  & --- \\[2pt]
X3          & Facility Tour{++}        & Pr, Ct, St & Sym, Grd
  & 7-zone tour; earlier zones degrade more under recency bias
  & --- \\[2pt]
X4          & Compound Witness         & Pr, Ct, St & Sym
  & 5-zone tour with embedded Witness Stand sub-sequence mid-tour
  & --- \\[2pt]
X5          & Cascading Testimony      & Pr, Ct, St & Sym, Grd
  & 7-zone tour; later signposts contradict the agent's own earlier direct observations
  & Grd\,$\bullet$ \\[2pt]
X6          & Return Visit             & Pr, Ct, St & Sym, Grd
  & 5-zone tour + revisit; Zone A modified on return; tests world-model update after backtracking
  & --- \\[2pt]
X7          & Dragon Keep              & Pr, Ct, St, Ca & Sym, Grd
  & 8-zone RPG tour; unreliable NPCs, detour, backtracking, goal requires integrating information across all zones
  & --- \\
\end{longtable}
}

\section{\gridname Environmental Mechanics}
\label{appendix:gridworld-mechanics}
Beyond standard MiniGrid objects, we implement the following mechanics to support probing of memory and causal hallucinations.
 
\paragraph{Tiles.}
\texttt{RiverTile(direction, speed)} carries objects in a fixed direction at a fixed rate
per step.
\texttt{FireTile} is impassable and is extinguished by any object with
\texttt{wet\_turns\_remaining\,$>$\,0}.
\texttt{FloodTile(rise\_step)} becomes impassable at a specified step count.
\texttt{PressurePlate(effect)} supports two effect types: \emph{continuous} (gate open while
plate is weighted) and \emph{trigger} (one-shot, permanent).
\texttt{DarkZone} tiles are passable but block the agent's field of view.
 
\paragraph{Object states.}
Objects carry a \texttt{condition} attribute (\textsc{dry} $|$ \textsc{wet} $|$ \textsc{soaked})
with \texttt{wet\_turns\_remaining} decrementing each step off-river and resetting on re-entry.
This state is the primary mechanic for C2 and C6.
 
\paragraph{Adversarial testimony objects.}
\texttt{NoticeBoardObject(text)} renders stale written information that is always visible in
the serializer. It is accurate in M1, inaccurate in M4, and actively wrong in C5a and C6,
making it the primary source of adversarial testimony across tiers.
\texttt{SignpostObject(text, accurate)} supports controlled false information with stated
accuracy rates, used in U2.

\section{\terminalname: Further Details}
\label{app:terminal-details}
\terminalname is constructed from a collection of over 100 Terminal-Bench~\cite{terminal-bench} tasks. We intentionally chose an older and smaller agent, GPT-4o-mini, to attempt these tasks and generate trajectories. This choice reflects the hypothesis that weaker agents tend to be less methodical and less thorough, producing longer, noisier, and more failure-prone interaction traces. Such trajectories create richer settings in which hallucination-relevant evidence may be hidden, stale, partially contradicted, or distributed across many terminal steps, allowing us to test whether later and stronger models can still recover the correct answer from the available evidence.

To construct the benchmark, we developed a runtime probe injector that observes the agent’s terminal session during task execution. The injector records each command, working directory, terminal output, pre- and post-command pane state, and relevant file snapshots. From these observations, it constructs evidence contexts and generates targeted probe questions whose answers are grounded in the recorded trajectory. Crucially, the probes are generated from the runtime trace rather than shown to the original task-solving agent, so they do not alter the agent’s behavior or contaminate the trajectory.

The resulting probes are designed to test several hallucination failure modes in agentic settings, including stale memory, perceptual mistakes, causal over-attribution, uncertainty overclaiming, and cross-category reasoning failures that require connecting commands, files, outputs, and task state. Each probe has a structured answer schema and an associated golden answer, making evaluation mostly automatic while keeping the questions narrow and auditable. After extraction, probes are filtered and curated for answerability and difficulty, yielding a compact benchmark subset of high-quality examples.

\begin{table}
\centering
\small
\caption{%
  Thinking depth ablation on \terminalname\ (probe accuracy, higher is better).
  Probe categories: Ca\,=\,Causal, X\,=\,Cross-category, M\,=\,Memory,
  P\,=\,Perceptual, U\,=\,Uncertainty.
  \textit{Medium} rows are the main leaderboard entries.%
}
\label{tab:terminal-ablation}
\begin{tabular}{llccccccc}
\toprule
\textbf{Model} & \textbf{Budget} & \textbf{N}
  & \textbf{Overall} & \textbf{Ca} & \textbf{X}
  & \textbf{M} & \textbf{P} & \textbf{U} \\
\midrule
\multirow{5}{*}{Claude Sonnet 4.6}
  & none   & 529 & 79.2\% & 89.6\% & 84.9\% & 88.0\% & 74.3\% & 58.7\% \\
  & low    & 529 & 80.3\% & 87.7\% & 82.1\% & 88.9\% & 78.1\% & 64.4\% \\
  & \textit{medium} & 529 & \textit{82.2\%} & \textit{90.6\%} & \textit{85.8\%}
    & \textit{88.9\%} & \textit{81.9\%} & \textit{63.5\%} \\
  & high   & 529 & 83.9\% & 90.6\% & 86.8\% & 90.7\% & 89.5\% & 61.5\% \\
  & max    & 529 & 86.2\% & 95.3\% & 88.7\% & 91.7\% & 90.5\% & 64.4\% \\
\midrule
\multirow{5}{*}{GPT-5.5}
  & none   & 529 & 86.4\% & 93.4\% & 91.5\% & 84.3\% & 81.9\% & 80.8\% \\
  & low    & 529 & 94.9\% & 99.1\% & 99.1\% & 99.1\% & 99.0\% & 77.9\% \\
  & \textit{medium} & 529 & \textit{94.1\%} & \textit{99.1\%} & \textit{98.1\%}
    & \textit{97.2\%} & \textit{99.0\%} & \textit{76.9\%} \\
  & high   & 527$\dagger$ & 96.6\% & 100.0\% & 99.1\% & 99.1\% & 100.0\% & 84.3\% \\
  & xhigh  & 527$\dagger$ & 95.3\% & 99.1\%  & 99.1\% & 99.1\% & 100.0\% & 78.6\% \\
\bottomrule
\end{tabular}
\end{table}

\begin{figure}
    \centering
    \begin{subfigure}{\textwidth}
        \centering
        \includegraphics[width=\linewidth]{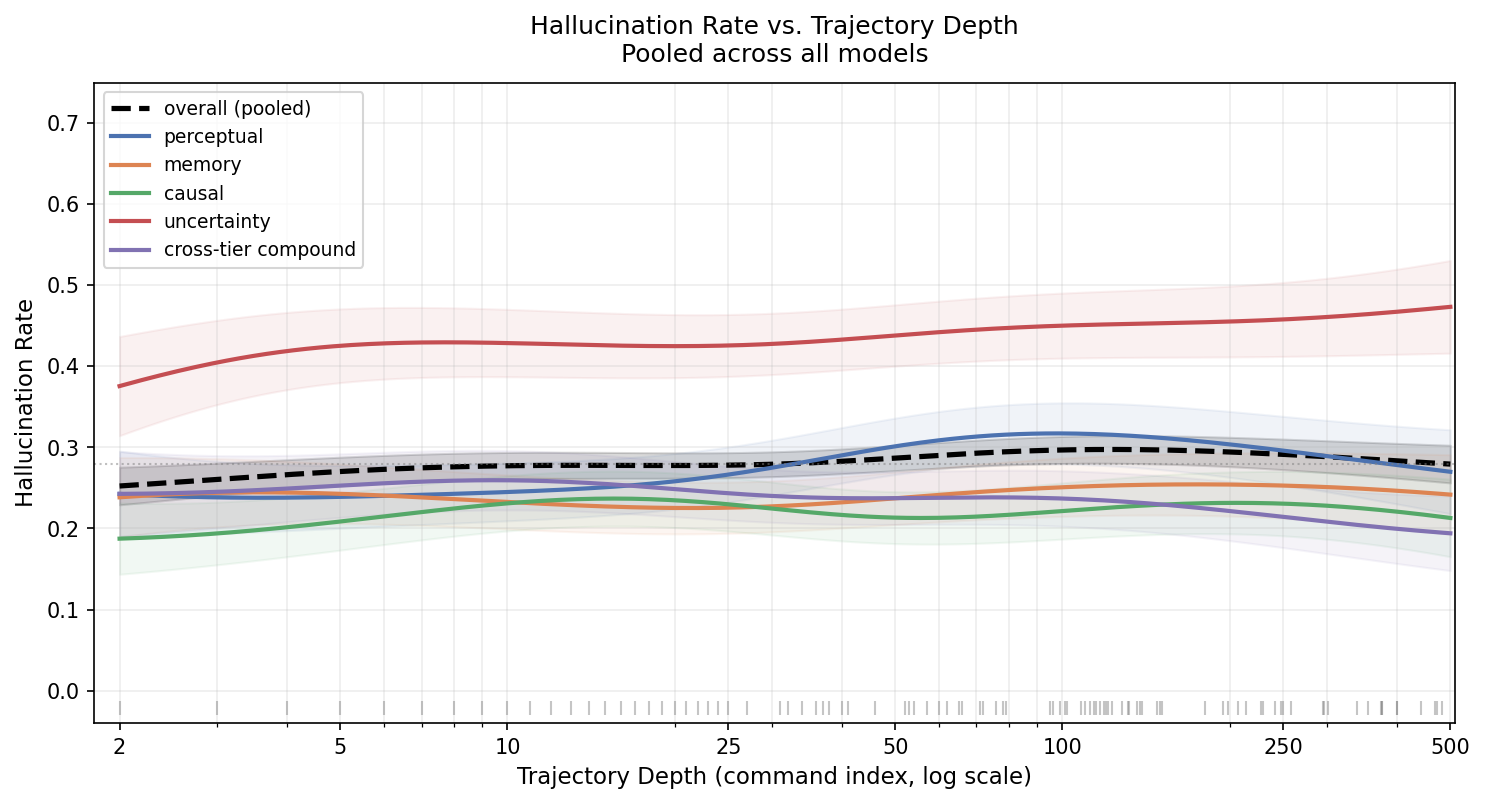}
        \caption{Hallucination Rate vs. Trajectory Depth pooled across all models}
    \end{subfigure}
    \vspace{0.5em}
    \vspace{0.5em}
    \begin{subfigure}{\textwidth}
        \centering
        \includegraphics[width=\linewidth]{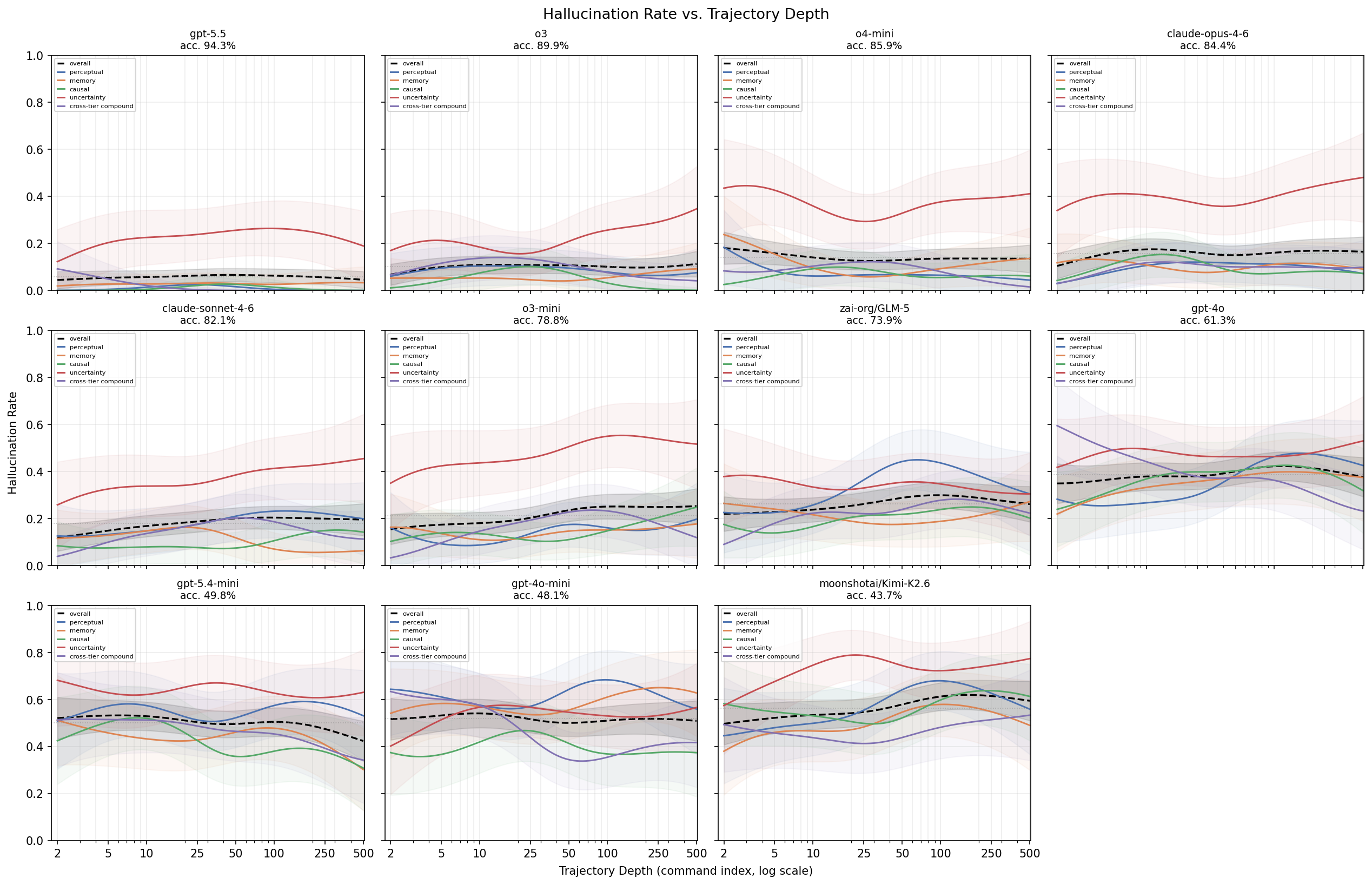}
        \caption{Hallucination Rate vs. Trajectory Depth per model}
    \end{subfigure}
    \vspace{0.5em}
    \caption{\textbf{\terminalname Hallucination Rate vs. Trajectory Depth}. Trajectory depth is defined as the index of the trigger command, i.e., the number of shell commands executed before the probe injection. Hallucination rates are smoothed using a Gaussian kernel in log-depth space. Context is truncated to 60k characters using middle truncation (retaining the start and end of trajectories). Uncertainty probes (red) show the strongest depth sensitivity, with increasing rates at larger depths, while most other probe types remain stable or increase mildly.}
    \label{app:terminal-hallu-depth}
\end{figure}

\section{\terminalname Thinking Depth Ablation}
\label{app:terminal-ablation}

Table~\ref{tab:terminal-ablation} reports probe accuracy by category across
thinking/reasoning depth for Sonnet~4.6 and GPT-5.5, while Figure \ref{app:terminal-hallu-depth} visualizes the hallucination rate by trajectory depth.
The \textit{medium} rows match the main leaderboard entries exactly.
$\dagger$~Two probes permanently excluded from GPT-5.5 high/xhigh runs due to
content policy rejection on \texttt{sql-injection-attack} probes at high reasoning
effort ($N{=}527$).

\section{\gridname Reasoning Ablation}
\label{appendix:gridworld-reasoning-ablation}

From Table \ref{app:tab:reasoning-budget}, it appears that more reasoning does not necessarily correlate with lower hallucination. Interestingly, it mainly appears to increase hallucination for Causal while having negligible improvement on the other probe categories, aligning with the analysis in \S\ref{sec:thinking}.

\begin{table}[ht]
\centering
\caption{%
  \textbf{Effect of reasoning budget on hallucination rate by category for \gridname.}
  Categories: \textbf{P}\,=\,Perceptual, \textbf{M}\,=\,Memory,
  \textbf{C}\,=\,Causal, \textbf{U}\,=\,Uncertainty, \textbf{X}\,=\,Cross-category.
  For GPT-5.5, ``None'' indicates reasoning disabled; the default evaluation
  uses medium budget. For Claude Sonnet~4.6, the non-thinking model serves
  as the None baseline.
}
\label{app:tab:reasoning-budget}
\small
\setlength{\tabcolsep}{5pt}
\begin{tabular}{llrrrrrrr}
\toprule
\textbf{Model} & \textbf{Budget}
  & \textbf{Overall}
  & \textbf{P}
  & \textbf{M}
  & \textbf{C}
  & \textbf{U}
  & \textbf{X} \\
\midrule
\multirow{4}{*}{GPT-5.5}
  & None   & \cellcolor{pastelgreen!82!pastelred}\textbf{4.8\%} & \cellcolor{pastelgreen!99!pastelred}0.3\% & \cellcolor{pastelgreen!81!pastelred}5.2\% & \cellcolor{pastelgreen!10!pastelred}24.5\% & \cellcolor{pastelgreen!97!pastelred}0.8\% & \cellcolor{pastelgreen!91!pastelred}\textbf{2.5\%} \\
  & Low    & \cellcolor{pastelgreen!82!pastelred}5.0\%          & \cellcolor{pastelgreen!99!pastelred}\textbf{0.2\%} & \cellcolor{pastelgreen!80!pastelred}5.5\% & \cellcolor{pastelgreen!4!pastelred}25.9\%  & \cellcolor{pastelgreen!100!pastelred}\textbf{0.1\%} & \cellcolor{pastelgreen!90!pastelred}2.6\% \\
  & Medium & \cellcolor{pastelgreen!82!pastelred}5.0\%          & \cellcolor{pastelgreen!99!pastelred}0.4\% & \cellcolor{pastelgreen!78!pastelred}5.9\% & \cellcolor{pastelgreen!7!pastelred}25.3\%  & \cellcolor{pastelgreen!100!pastelred}\textbf{0.1\%} & \cellcolor{pastelgreen!90!pastelred}2.7\% \\
  & High   & \cellcolor{pastelgreen!82!pastelred}\textbf{4.8\%} & \cellcolor{pastelgreen!99!pastelred}0.3\% & \cellcolor{pastelgreen!80!pastelred}5.5\% & \cellcolor{pastelgreen!9!pastelred}24.7\%  & \cellcolor{pastelgreen!99!pastelred}0.3\% & \cellcolor{pastelgreen!91!pastelred}\textbf{2.5\%} \\
\midrule
\multirow{5}{*}{Claude Sonnet 4.6}
  & None   & \cellcolor{pastelgreen!73!pastelred}7.2\% & \cellcolor{pastelgreen!90!pastelred}2.8\% & \cellcolor{pastelgreen!82!pastelred}\textbf{5.0\%} & \cellcolor{pastelgreen!27!pastelred}\textbf{19.7\%} & \cellcolor{pastelgreen!96!pastelred}1.1\% & \cellcolor{pastelgreen!72!pastelred}7.6\% \\
  & Low    & \cellcolor{pastelgreen!70!pastelred}8.1\% & \cellcolor{pastelgreen!90!pastelred}2.8\% & \cellcolor{pastelgreen!73!pastelred}7.2\% & \cellcolor{pastelgreen!8!pastelred}25.0\%  & \cellcolor{pastelgreen!95!pastelred}1.3\% & \cellcolor{pastelgreen!73!pastelred}\textbf{7.3\%} \\
  & Medium & \cellcolor{pastelgreen!72!pastelred}7.7\% & \cellcolor{pastelgreen!96!pastelred}1.1\% & \cellcolor{pastelgreen!77!pastelred}6.2\% & \cellcolor{pastelgreen!11!pastelred}24.0\% & \cellcolor{pastelgreen!96!pastelred}1.1\% & \cellcolor{pastelgreen!72!pastelred}7.5\% \\
  & High   & \cellcolor{pastelgreen!70!pastelred}8.0\% & \cellcolor{pastelgreen!96!pastelred}\textbf{1.0\%} & \cellcolor{pastelgreen!76!pastelred}6.4\% & \cellcolor{pastelgreen!4!pastelred}25.9\%  & \cellcolor{pastelgreen!95!pastelred}1.3\% & \cellcolor{pastelgreen!72!pastelred}7.7\% \\
  & Max    & \cellcolor{pastelgreen!70!pastelred}8.1\% & \cellcolor{pastelgreen!91!pastelred}2.5\% & \cellcolor{pastelgreen!78!pastelred}5.9\% & \cellcolor{pastelgreen!0!pastelred}27.1\%  & \cellcolor{pastelgreen!96!pastelred}1.2\% & \cellcolor{pastelgreen!72!pastelred}7.5\% \\
\bottomrule
\end{tabular}
\end{table}

\section{\gridname Hard Subset Results}
\label{appendix:gridworld-hard}

\begin{table}[H]
\centering
\caption{%
  \textbf{\gridname Hard subset results.}
  The 12 hardest (level, serializer) pairs where $\geq$\,5 models achieve
  $\geq$\,20\% hallucination rate (see \S\ref{appendix:gridworld-all-levels}).
  Models sorted by mean hard-subset hallucination rate (ascending).
  Cell color reflects relative hallucination rate (green\,=\,low, red\,=\,high).
  \textbf{Bold} = column minimum.
  \textbf{Columns:} C1a\,nb, C1a, C1b, C1b\,nb, C5a, C6,
  M1\textsubscript{6}, M1\textsubscript{9}, P1, P2, P4, X5.
}
\label{tab:hard-subset}
\footnotesize
\setlength{\tabcolsep}{3pt}
\begin{adjustbox}{max width=\linewidth}
\begin{tabular}{lrccccccccccccc}
\toprule
\textbf{Model} & \textbf{Mean}
  & \rotatebox{60}{\textbf{C1a nb}}
  & \rotatebox{60}{\textbf{C1a}}
  & \rotatebox{60}{\textbf{C1b}}
  & \rotatebox{60}{\textbf{C1b nb}}
  & \rotatebox{60}{\textbf{C5a}}
  & \rotatebox{60}{\textbf{C6}}
  & \rotatebox{60}{\textbf{M1\textsubscript{6}}}
  & \rotatebox{60}{\textbf{M1\textsubscript{9}}}
  & \rotatebox{60}{\textbf{P1}}
  & \rotatebox{60}{\textbf{P2}}
  & \rotatebox{60}{\textbf{P4}}
  & \rotatebox{60}{\textbf{X5}} \\
\midrule
o3
  & \cellcolor{pastelgreen!85!pastelred}\textbf{15.5\%}
  & \cellcolor{pastelgreen!81!pastelred}\textbf{18.8}
  & \cellcolor{pastelgreen!86!pastelred} 13.7
  & \cellcolor{pastelgreen!74!pastelred} 26.2
  & \cellcolor{pastelgreen!74!pastelred} 26.2
  & \cellcolor{pastelgreen!69!pastelred} 31.2
  & \cellcolor{pastelgreen!69!pastelred} 30.8
  & \cellcolor{pastelgreen!96!pastelred}  4.3
  & \cellcolor{pastelgreen!86!pastelred} 14.3
  & \cellcolor{pastelgreen!100!pastelred}\textbf{ 0.0}
  & \cellcolor{pastelgreen!95!pastelred}  5.4
  & \cellcolor{pastelgreen!100!pastelred}\textbf{ 0.0}
  & \cellcolor{pastelgreen!85!pastelred} 15.4 \\
o3-mini
  & \cellcolor{pastelgreen!83!pastelred} 17.0\%
  & \cellcolor{pastelgreen!63!pastelred} 37.5
  & \cellcolor{pastelgreen!68!pastelred} 32.5
  & \cellcolor{pastelgreen!79!pastelred} 21.5
  & \cellcolor{pastelgreen!79!pastelred} 21.3
  & \cellcolor{pastelgreen!76!pastelred}\textbf{23.8}
  & \cellcolor{pastelgreen!72!pastelred} 28.3
  & \cellcolor{pastelgreen!99!pastelred}  1.4
  & \cellcolor{pastelgreen!84!pastelred} 15.7
  & \cellcolor{pastelgreen!98!pastelred}  2.5
  & \cellcolor{pastelgreen!100!pastelred}\textbf{ 0.0}
  & \cellcolor{pastelgreen!100!pastelred}  0.0
  & \cellcolor{pastelgreen!81!pastelred} 19.3 \\
GPT-5.5 (T)
  & \cellcolor{pastelgreen!83!pastelred} 17.1\%
  & \cellcolor{pastelgreen!64!pastelred} 36.3
  & \cellcolor{pastelgreen!63!pastelred} 37.5
  & \cellcolor{pastelgreen!83!pastelred} 17.5
  & \cellcolor{pastelgreen!83!pastelred} 17.5
  & \cellcolor{pastelgreen!75!pastelred} 25.0
  & \cellcolor{pastelgreen!75!pastelred}\textbf{25.0}
  & \cellcolor{pastelgreen!90!pastelred} 10.0
  & \cellcolor{pastelgreen!77!pastelred} 22.9
  & \cellcolor{pastelgreen!100!pastelred}  0.0
  & \cellcolor{pastelgreen!94!pastelred}  5.7
  & \cellcolor{pastelgreen!100!pastelred}  0.0
  & \cellcolor{pastelgreen!92!pastelred}\textbf{ 8.0} \\
Claude Opus 4.6 (T)
  & \cellcolor{pastelgreen!82!pastelred} 17.8\%
  & \cellcolor{pastelgreen!72!pastelred} 27.8
  & \cellcolor{pastelgreen!54!pastelred} 46.1
  & \cellcolor{pastelgreen!84!pastelred} 15.6
  & \cellcolor{pastelgreen!91!pastelred}  8.9
  & \cellcolor{pastelgreen!63!pastelred} 37.5
  & \cellcolor{pastelgreen!60!pastelred} 39.6
  & \cellcolor{pastelgreen!100!pastelred}\textbf{ 0.0}
  & \cellcolor{pastelgreen!93!pastelred}  7.0
  & \cellcolor{pastelgreen!95!pastelred}  5.0
  & \cellcolor{pastelgreen!98!pastelred}  2.0n
  & \cellcolor{pastelgreen!89!pastelred} 11.1
  & \cellcolor{pastelgreen!87!pastelred} 12.8 \\
Claude Sonnet 4.6 (T)
  & \cellcolor{pastelgreen!82!pastelred} 17.9\%
  & \cellcolor{pastelgreen!69!pastelred} 31.2
  & \cellcolor{pastelgreen!69!pastelred} 31.2
  & \cellcolor{pastelgreen!91!pastelred}  8.8
  & \cellcolor{pastelgreen!93!pastelred}  7.5
  & \cellcolor{pastelgreen!50!pastelred} 50.0
  & \cellcolor{pastelgreen!62!pastelred} 38.5
  & \cellcolor{pastelgreen!82!pastelred} 17.7
  & \cellcolor{pastelgreen!99!pastelred}  1.5
  & \cellcolor{pastelgreen!93!pastelred}  7.5
  & \cellcolor{pastelgreen!98!pastelred}  2.0
  & \cellcolor{pastelgreen!100!pastelred}  0.0
  & \cellcolor{pastelgreen!81!pastelred} 18.9 \\
Claude Opus 4.6
  & \cellcolor{pastelgreen!81!pastelred} 19.5\%
  & \cellcolor{pastelgreen!72!pastelred} 28.4
  & \cellcolor{pastelgreen!58!pastelred} 42.0
  & \cellcolor{pastelgreen!80!pastelred} 20.0
  & \cellcolor{pastelgreen!89!pastelred} 11.3
  & \cellcolor{pastelgreen!65!pastelred} 35.0
  & \cellcolor{pastelgreen!53!pastelred} 46.7
  & \cellcolor{pastelgreen!100!pastelred}  0.0
  & \cellcolor{pastelgreen!95!pastelred}  5.0
  & \cellcolor{pastelgreen!90!pastelred} 10.0
  & \cellcolor{pastelgreen!98!pastelred}  2.4
  & \cellcolor{pastelgreen!80!pastelred} 20.0
  & \cellcolor{pastelgreen!87!pastelred} 12.8 \\
Claude Sonnet 4.6
  & \cellcolor{pastelgreen!80!pastelred} 20.0\%
  & \cellcolor{pastelgreen!75!pastelred} 25.0
  & \cellcolor{pastelgreen!64!pastelred} 35.7
  & \cellcolor{pastelgreen!89!pastelred} 11.4
  & \cellcolor{pastelgreen!93!pastelred}\textbf{ 7.1}
  & \cellcolor{pastelgreen!55!pastelred} 45.0
  & \cellcolor{pastelgreen!44!pastelred} 56.1
  & \cellcolor{pastelgreen!79!pastelred} 20.8
  & \cellcolor{pastelgreen!100!pastelred}\textbf{ 0.0}
  & \cellcolor{pastelgreen!88!pastelred} 12.5
  & \cellcolor{pastelgreen!95!pastelred}  5.3
  & \cellcolor{pastelgreen!100!pastelred}  0.0
  & \cellcolor{pastelgreen!79!pastelred} 21.4 \\
o4-mini
  & \cellcolor{pastelgreen!79!pastelred} 21.3\%
  & \cellcolor{pastelgreen!49!pastelred} 51.2
  & \cellcolor{pastelgreen!55!pastelred} 45.0
  & \cellcolor{pastelgreen!74!pastelred} 26.2
  & \cellcolor{pastelgreen!74!pastelred} 26.2
  & \cellcolor{pastelgreen!69!pastelred} 31.2
  & \cellcolor{pastelgreen!69!pastelred} 30.8
  & \cellcolor{pastelgreen!99!pastelred}  1.4
  & \cellcolor{pastelgreen!89!pastelred} 11.4
  & \cellcolor{pastelgreen!93!pastelred}  7.5
  & \cellcolor{pastelgreen!92!pastelred}  8.3
  & \cellcolor{pastelgreen!100!pastelred}  0.0
  & \cellcolor{pastelgreen!84!pastelred} 15.7 \\

GPT-4o
  & \cellcolor{pastelgreen!69!pastelred} 31.1\%
  & \cellcolor{pastelgreen!44!pastelred} 56.2
  & \cellcolor{pastelgreen!50!pastelred} 50.0
  & \cellcolor{pastelgreen!88!pastelred} 12.5
  & \cellcolor{pastelgreen!91!pastelred}  8.8
  & \cellcolor{pastelgreen!63!pastelred} 37.5
  & \cellcolor{pastelgreen!58!pastelred} 42.5
  & \cellcolor{pastelgreen!81!pastelred} 18.6
  & \cellcolor{pastelgreen!71!pastelred} 28.6
  & \cellcolor{pastelgreen!78!pastelred} 22.5
  & \cellcolor{pastelgreen!47!pastelred} 53.3
  & \cellcolor{pastelgreen!88!pastelred} 12.2
  & \cellcolor{pastelgreen!70!pastelred} 30.3 \\
GLM-5
  & \cellcolor{pastelgreen!67!pastelred} 33.5\%
  & \cellcolor{pastelgreen!43!pastelred} 57.5
  & \cellcolor{pastelgreen!46!pastelred} 53.8
  & \cellcolor{pastelgreen!93!pastelred}\textbf{ 7.5}
  & \cellcolor{pastelgreen!90!pastelred} 10.0
  & \cellcolor{pastelgreen!63!pastelred} 37.5
  & \cellcolor{pastelgreen!60!pastelred} 40.0
  & \cellcolor{pastelgreen!79!pastelred} 21.4
  & \cellcolor{pastelgreen!84!pastelred} 15.7
  & \cellcolor{pastelgreen!73!pastelred} 27.5
  & \cellcolor{pastelgreen!59!pastelred} 41.3
  & \cellcolor{pastelgreen!46!pastelred} 54.4
  & \cellcolor{pastelgreen!65!pastelred} 35.4 \\
GPT-5.4-mini
  & \cellcolor{pastelgreen!66!pastelred} 34.1\%
  & \cellcolor{pastelgreen!80!pastelred} 20.0
  & \cellcolor{pastelgreen!88!pastelred}\textbf{12.5}
  & \cellcolor{pastelgreen!84!pastelred} 16.2
  & \cellcolor{pastelgreen!81!pastelred} 18.8
  & \cellcolor{pastelgreen!64!pastelred} 36.3
  & \cellcolor{pastelgreen!60!pastelred} 40.0
  & \cellcolor{pastelgreen!69!pastelred} 31.4
  & \cellcolor{pastelgreen!86!pastelred} 14.3
  & \cellcolor{pastelgreen!65!pastelred} 35.0
  & \cellcolor{pastelgreen!8!pastelred} 92.3
  & \cellcolor{pastelgreen!52!pastelred} 47.8
  & \cellcolor{pastelgreen!56!pastelred} 44.3 \\
DeepSeek-V3-0324
  & \cellcolor{pastelgreen!65!pastelred} 34.6\%
  & \cellcolor{pastelgreen!39!pastelred} 61.3
  & \cellcolor{pastelgreen!39!pastelred} 61.3
  & \cellcolor{pastelgreen!89!pastelred} 11.3
  & \cellcolor{pastelgreen!89!pastelred} 11.3
  & \cellcolor{pastelgreen!51!pastelred} 48.8
  & \cellcolor{pastelgreen!68!pastelred} 32.5
  & \cellcolor{pastelgreen!71!pastelred} 29.4
  & \cellcolor{pastelgreen!90!pastelred} 10.0
  & \cellcolor{pastelgreen!65!pastelred} 35.0
  & \cellcolor{pastelgreen!55!pastelred} 45.1
  & \cellcolor{pastelgreen!56!pastelred} 44.4
  & \cellcolor{pastelgreen!75!pastelred} 24.9 \\
Qwen-3-30B
  & \cellcolor{pastelgreen!62!pastelred} 37.8\%
  & \cellcolor{pastelgreen!75!pastelred} 25.0
  & \cellcolor{pastelgreen!63!pastelred} 37.5
  & \cellcolor{pastelgreen!63!pastelred} 37.5
  & \cellcolor{pastelgreen!50!pastelred} 50.0
  & \cellcolor{pastelgreen!63!pastelred} 37.5
  & \cellcolor{pastelgreen!58!pastelred} 41.7
  & \cellcolor{pastelgreen!71!pastelred} 28.6
  & \cellcolor{pastelgreen!86!pastelred} 14.3
  & \cellcolor{pastelgreen!80!pastelred} 20.0
  & \cellcolor{pastelgreen!31!pastelred} 69.5
  & \cellcolor{pastelgreen!44!pastelred} 55.6
  & \cellcolor{pastelgreen!63!pastelred} 36.9 \\
GPT-4o-mini
  & \cellcolor{pastelgreen!60!pastelred} 39.7\%
  & \cellcolor{pastelgreen!71!pastelred} 28.7
  & \cellcolor{pastelgreen!91!pastelred}  8.8
  & \cellcolor{pastelgreen!58!pastelred} 42.5
  & \cellcolor{pastelgreen!60!pastelred} 40.0
  & \cellcolor{pastelgreen!50!pastelred} 50.0
  & \cellcolor{pastelgreen!68!pastelred} 31.7
  & \cellcolor{pastelgreen!71!pastelred} 28.6
  & \cellcolor{pastelgreen!84!pastelred} 15.7
  & \cellcolor{pastelgreen!70!pastelred} 30.0
  & \cellcolor{pastelgreen!17!pastelred} 83.3
  & \cellcolor{pastelgreen!33!pastelred} 66.7
  & \cellcolor{pastelgreen!49!pastelred} 50.9 \\
Kimi K2.6
  & \cellcolor{pastelgreen!46!pastelred} 53.9\%
  & \cellcolor{pastelgreen!19!pastelred} 80.8
  & \cellcolor{pastelgreen!32!pastelred} 68.2
  & \cellcolor{pastelgreen!5!pastelred} 95.0
  & \cellcolor{pastelgreen!0!pastelred}100.0
  & \cellcolor{pastelgreen!40!pastelred} 60.0
  & \cellcolor{pastelgreen!25!pastelred} 75.0
  & \cellcolor{pastelgreen!63!pastelred} 36.7
  & \cellcolor{pastelgreen!67!pastelred} 33.3
  & \cellcolor{pastelgreen!75!pastelred} 25.0
  & \cellcolor{pastelgreen!62!pastelred} 38.2
  & \cellcolor{pastelgreen!89!pastelred} 11.1
  & \cellcolor{pastelgreen!77!pastelred} 23.1 \\
\bottomrule
\end{tabular}
\end{adjustbox}
\end{table}

\section{In-Navigation Experiments for \textsc{\gridname}}
\label{appendix:in-navigation}
In this section, we evaluate models in an \textbf{In-Navigation} (\textsc{InNav}) setting where the model actively navigates the environment while being probed about world state. Technically, \textsc{InNav} mode uses chat completions with conversational turn-based history in the standard messages format: each navigation step (action selection, observation receipt) is appended to the conversation history, and probes are asked as follow-up messages within this ongoing dialogue. This mirrors how agents would be deployed in interactive settings where world related queries arise during task execution, e.g., with a user in the loop conversing with the model to understand what's going on.

To enable controlled comparison without conflating navigation ability with reasoning ability, we use a single high-performance navigation model (GPT-5.4-mini with \texttt{reasoning\_effort=high}) 
to generate trajectories for \textit{all} probe models.

\subsection{Epistemic Grounding vs. Cognitive Load: Does Navigation Help or Hurt Hallucination?}
\label{sec:nav-effect}
Navigation introduces two competing effects: \textbf{epistemic grounding} (action-perception feedback may anchor the agent's internal world model better) versus \textbf{cognitive load} (navigation demands may degrade state tracking). To isolate navigation's net impact, we compare two conditions using identical trajectories: \textbf{In-Navigation} (\textsc{InNav}), where models answer probes while navigating, versus \textbf{Controlled Static} (\textsc{CtrlStatic}), where models observe the same states without navigation context. This paired design controls for trajectory information, isolating whether navigation helps or hurts hallucination rates. Results are in Table \ref{tab:navigation-effect}.

\begin{table}[ht]
\centering
\caption{%
  \textbf{Navigation impact on hallucination rates (overall and by category) for \gridname.} 
  Navigation Effect (NavEff) = $\text{\textsc{InNav}} - \text{\textsc{CtrlStatic}}$ hallucination \%.
  Left: Overall results across all 31 worlds.
  Right: Category-specific NavEff results (P=Perception, C=Causal, M=Memory, U=Uncertainty, X=X-level). 
  Models sorted by overall NavEff. ** means statistically significant (95\% CI excludes zero).
  Cell color reflects relative hallucination rate or effect (green=low/improvement, red=high/degradation); 
  \textbf{bold} marks the column minimum.
}
\label{tab:navigation-effect}
\small
\setlength{\tabcolsep}{3.5pt}
\begin{minipage}[t]{0.48\textwidth}
\centering
\begin{tabular}{lrrr}
\toprule
\textbf{Model}
  & \textbf{InNav}
  & \textbf{CtrlSt}
  & \textbf{NavEff} \\
\midrule
gpt-5.5       & \cellcolor{pastelgreen!92!pastelred}\textbf{27.4} & \cellcolor{pastelgreen!48!pastelred}40.7 & \cellcolor{pastelgreen!83!pastelred}\textbf{$-13.2$}** \\
GLM-5         & \cellcolor{pastelgreen!60!pastelred}37.0 & \cellcolor{pastelgreen!45!pastelred}41.5 & \cellcolor{pastelgreen!45!pastelred}$-4.4$** \\
o3-mini       & \cellcolor{pastelgreen!63!pastelred}36.1 & \cellcolor{pastelgreen!48!pastelred}40.5 & \cellcolor{pastelgreen!45!pastelred}$-4.4$** \\
o4-mini       & \cellcolor{pastelgreen!58!pastelred}37.6 & \cellcolor{pastelgreen!46!pastelred}41.3 & \cellcolor{pastelgreen!43!pastelred}$-3.8$** \\
o3            & \cellcolor{pastelgreen!57!pastelred}37.8 & \cellcolor{pastelgreen!47!pastelred}41.0 & \cellcolor{pastelgreen!40!pastelred}$-3.2$** \\
\midrule
DeepSeek-V3   & \cellcolor{pastelgreen!40!pastelred}43.0 & \cellcolor{pastelgreen!32!pastelred}45.4 & \cellcolor{pastelgreen!37!pastelred}$-2.4$ \\
gpt-4o        & \cellcolor{pastelgreen!54!pastelred}38.9 & \cellcolor{pastelgreen!49!pastelred}40.2 & \cellcolor{pastelgreen!32!pastelred}$-1.3$ \\
gpt-4o-mini   & \cellcolor{pastelgreen!52!pastelred}39.5 & \cellcolor{pastelgreen!51!pastelred}39.8 & \cellcolor{pastelgreen!27!pastelred}$-0.3$ \\
Opus 4.6      & \cellcolor{pastelgreen!61!pastelred}36.8 & \cellcolor{pastelgreen!64!pastelred}\textbf{35.7} & \cellcolor{pastelgreen!21!pastelred}$+1.1$ \\
Sonnet 4.6    & \cellcolor{pastelgreen!59!pastelred}37.3 & \cellcolor{pastelgreen!63!pastelred}36.1 & \cellcolor{pastelgreen!21!pastelred}$+1.2$ \\
Kimi-K2.6     & \cellcolor{pastelgreen!60!pastelred}37.0 & \cellcolor{pastelgreen!64!pastelred}35.8 & \cellcolor{pastelgreen!20!pastelred}$+1.3$ \\
gpt-5.4-mini  & \cellcolor{pastelgreen!50!pastelred}40.1 & \cellcolor{pastelgreen!59!pastelred}37.2 & \cellcolor{pastelgreen!13!pastelred}$+2.9$ \\
\bottomrule
\end{tabular}
\end{minipage}%
\hfill
\begin{minipage}[t]{0.48\textwidth}
\centering
\begin{tabular}{lccccc}
\toprule
\textbf{Model} & \textbf{P} & \textbf{C} & \textbf{M} & \textbf{U} & \textbf{X} \\
\midrule
gpt-5.5       & \cellcolor{pastelgreen!17!pastelred}$+2.0$  & \cellcolor{pastelgreen!95!pastelred}\textbf{$-15.8$} & \cellcolor{pastelgreen!68!pastelred}\textbf{$-9.6$} & \cellcolor{pastelgreen!100!pastelred}\textbf{$-16.9$} & \cellcolor{pastelgreen!84!pastelred}\textbf{$-13.3$} \\
GLM-5         & \cellcolor{pastelgreen!27!pastelred}$-0.3$  & \cellcolor{pastelgreen!45!pastelred}$-4.3$  & \cellcolor{pastelgreen!17!pastelred}$+2.0$ & \cellcolor{pastelgreen!47!pastelred}$-4.9$ & \cellcolor{pastelgreen!81!pastelred}$-12.6$ \\
o3-mini       & \cellcolor{pastelgreen!28!pastelred}$-0.6$  & \cellcolor{pastelgreen!52!pastelred}$-6.0$  & \cellcolor{pastelgreen!40!pastelred}$-3.3$ & \cellcolor{pastelgreen!67!pastelred}$-9.3$ & \cellcolor{pastelgreen!40!pastelred}$-3.1$ \\
o4-mini       & \cellcolor{pastelgreen!35!pastelred}\textbf{$-2.1$}  & \cellcolor{pastelgreen!39!pastelred}$-2.9$  & \cellcolor{pastelgreen!42!pastelred}$-3.6$ & \cellcolor{pastelgreen!55!pastelred}$-6.6$ & \cellcolor{pastelgreen!45!pastelred}$-4.3$ \\
o3            & \cellcolor{pastelgreen!27!pastelred}$-0.3$  & \cellcolor{pastelgreen!31!pastelred}$-1.2$  & \cellcolor{pastelgreen!40!pastelred}$-3.2$ & \cellcolor{pastelgreen!60!pastelred}$-7.7$ & \cellcolor{pastelgreen!45!pastelred}$-4.3$ \\
\midrule
DeepSeek-V3   & \cellcolor{pastelgreen!23!pastelred}$+0.7$  & \cellcolor{pastelgreen!64!pastelred}$-8.7$  & \cellcolor{pastelgreen!0!pastelred}$+10.2$ & \cellcolor{pastelgreen!61!pastelred}$-8.1$ & \cellcolor{pastelgreen!27!pastelred}$-0.1$ \\
gpt-4o        & \cellcolor{pastelgreen!34!pastelred}$-1.8$  & \cellcolor{pastelgreen!51!pastelred}$-5.8$  & \cellcolor{pastelgreen!22!pastelred}$+0.9$ & \cellcolor{pastelgreen!35!pastelred}$-2.1$ & \cellcolor{pastelgreen!12!pastelred}$+3.3$ \\
gpt-4o-mini   & \cellcolor{pastelgreen!9!pastelred}$+3.9$  & \cellcolor{pastelgreen!48!pastelred}$-5.1$  & \cellcolor{pastelgreen!0!pastelred}$+6.1$ & \cellcolor{pastelgreen!47!pastelred}$-4.9$ & \cellcolor{pastelgreen!23!pastelred}$+0.9$ \\
Opus 4.6      & \cellcolor{pastelgreen!2!pastelred}$+5.6$  & \cellcolor{pastelgreen!21!pastelred}$+1.1$  & \cellcolor{pastelgreen!18!pastelred}$+1.9$ & \cellcolor{pastelgreen!53!pastelred}$-6.3$ & \cellcolor{pastelgreen!17!pastelred}$+2.2$ \\
Sonnet 4.6    & \cellcolor{pastelgreen!30!pastelred}$-0.8$  & \cellcolor{pastelgreen!19!pastelred}$+1.7$  & \cellcolor{pastelgreen!9!pastelred}$+4.1$ & \cellcolor{pastelgreen!28!pastelred}$+0.6$ & \cellcolor{pastelgreen!22!pastelred}$+1.1$ \\
Kimi-K2.6     & \cellcolor{pastelgreen!13!pastelred}$+3.1$  & \cellcolor{pastelgreen!30!pastelred}$-0.9$  & \cellcolor{pastelgreen!14!pastelred}$+2.7$ & \cellcolor{pastelgreen!42!pastelred}$-2.6$ & \cellcolor{pastelgreen!8!pastelred}$+4.3$ \\
gpt-5.4-mini  & \cellcolor{pastelgreen!2!pastelred}$+5.5$  & \cellcolor{pastelgreen!28!pastelred}$-0.5$  & \cellcolor{pastelgreen!4!pastelred}$+5.3$ & \cellcolor{pastelgreen!17!pastelred}$+3.2$ & \cellcolor{pastelgreen!15!pastelred}$+2.7$ \\
\bottomrule
\end{tabular}
\end{minipage}
\end{table}

Epistemic grounding dominates overall (left table): 5 of 12 models show statistically significant hallucination \textit{reductions} from navigation (95\% CI excludes zero), while zero models show significant increases. The strongest benefits appear in reasoning-focused models (gpt-5.5: $-13.2\%$, GLM-5: $-4.4\%$, o3-mini: $-4.4\%$), showing reasoning models steadily gain from multiple action-feedback cycles in the world.

Category-specific patterns reveal nuanced effects (right table): \textbf{Causal} worlds show universal grounding benefits (nearly all models negative or near-zero), with gpt-5.5 achieving $-15.8\%$. \textbf{Uncertainty} worlds have largest variation: gpt-5.5 achieves the strongest benefit overall ($-16.9\%$), while most reasoning models benefit substantially ($-4.9$ to $-9.3\%$), but weaker models show mixed patterns (gpt-5.4-mini: $+3.2\%$, Sonnet 4.6: $+0.6\%$). 
\textbf{Perception} and \textbf{Memory} worlds exhibit model-dependent effects: frontier models show diverse patterns (gpt-5.5 Perception: $+2.0\%$ but Memory: $-9.6\%$), while smaller models experience cognitive load on Memory tasks (gpt-4o-mini: $+6.1\%$, gpt-5.4-mini: $+5.3\%$). \textbf{X-level} multi-zone worlds show the largest model variance: GLM-5 achieves $-12.6\%$ (strong grounding), while Kimi-K2.6 shows $+4.3\%$ (cognitive load dominates). These patterns suggest navigation's net impact depends on both task complexity and model reasoning capacity. 
Detailed category analysis appears in \S\ref{appendix:nav-by-category}.
Overall, \textsc{InNav} provides a rich setting to test if models can jointly handle both goal-seeking (navigation) and internal world model updating, and if they can use reasoning to synergize both.

\subsection{Trajectory Depth Analysis: Intrinsic Difficulty vs. Navigation Impact}
\label{sec:depth-analysis}

To understand whether hallucinations increase with trajectory depth, and whether this pattern differs between In-Navigation and Controlled Static modes, we analyze hallucination rates as a function of \textbf{relative trajectory position}. Each trajectory is divided into five quintiles: 1/5 (first 20\% of steps), 2/5 (20--40\%), 3/5 (40--60\%), 4/5 (60--80\%), and 5/5 (final 80--100\%). We compute linear slopes (percentage point increase in hallucination per quintile) for both modes.

\begin{figure}[ht]
\centering
\begin{subfigure}[b]{0.62\textwidth}
    \includegraphics[width=\textwidth]{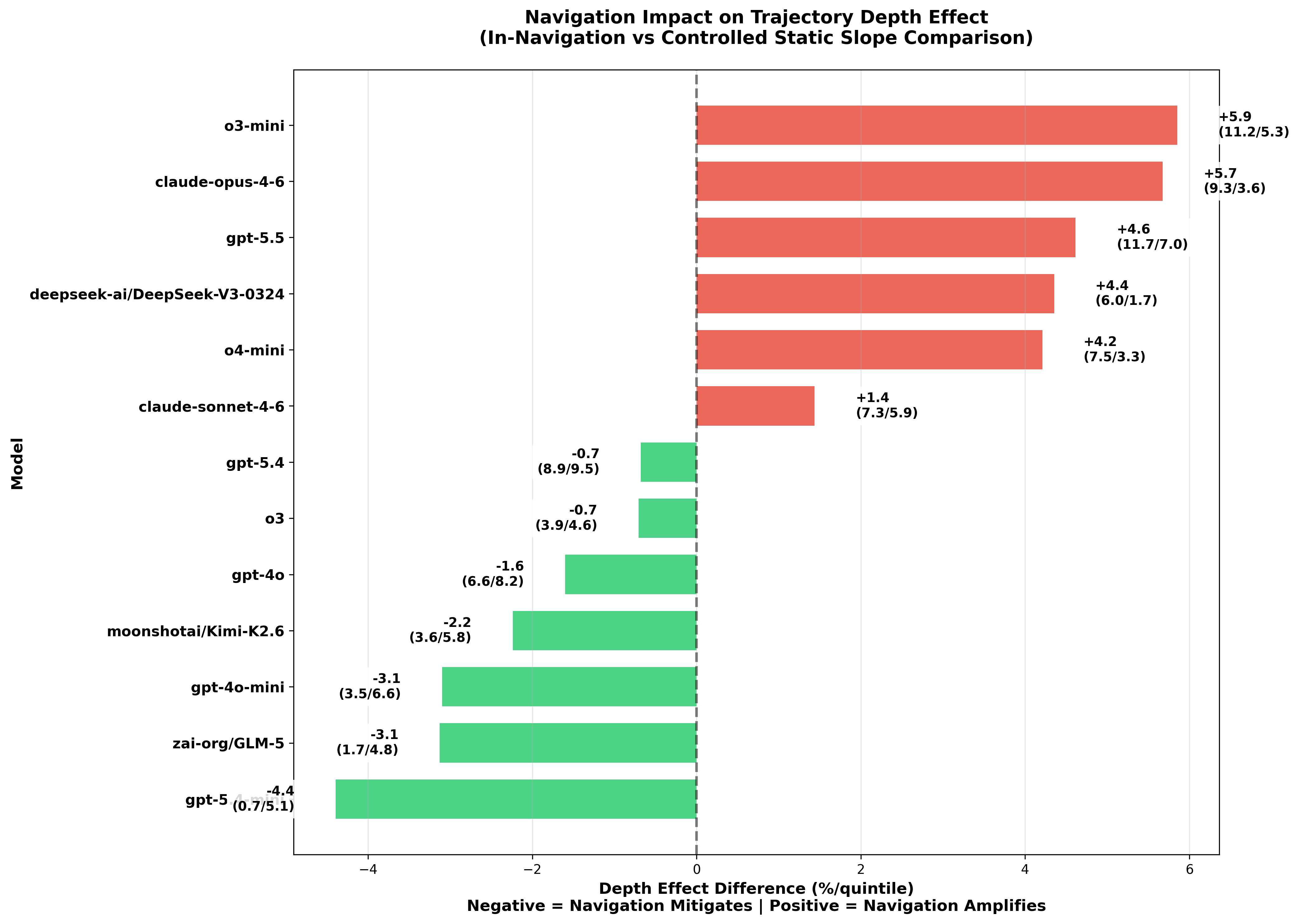}
    \caption{Slope difference ($\text{\textsc{InNav}} - \text{\textsc{CtrlStatic}}$) for \gridname. Red: cognitive load amplifies depth effect. Green: grounding mitigates depth effect. Labels show: Diff (InNav/CtrlStatic) in \%/quintile. Magnitude indicates strength of navigation's impact on depth accumulation.}
    \label{fig:depth-diff}
\end{subfigure}
\hfill
\begin{subfigure}[b]{0.35\textwidth}
    \includegraphics[width=\textwidth]{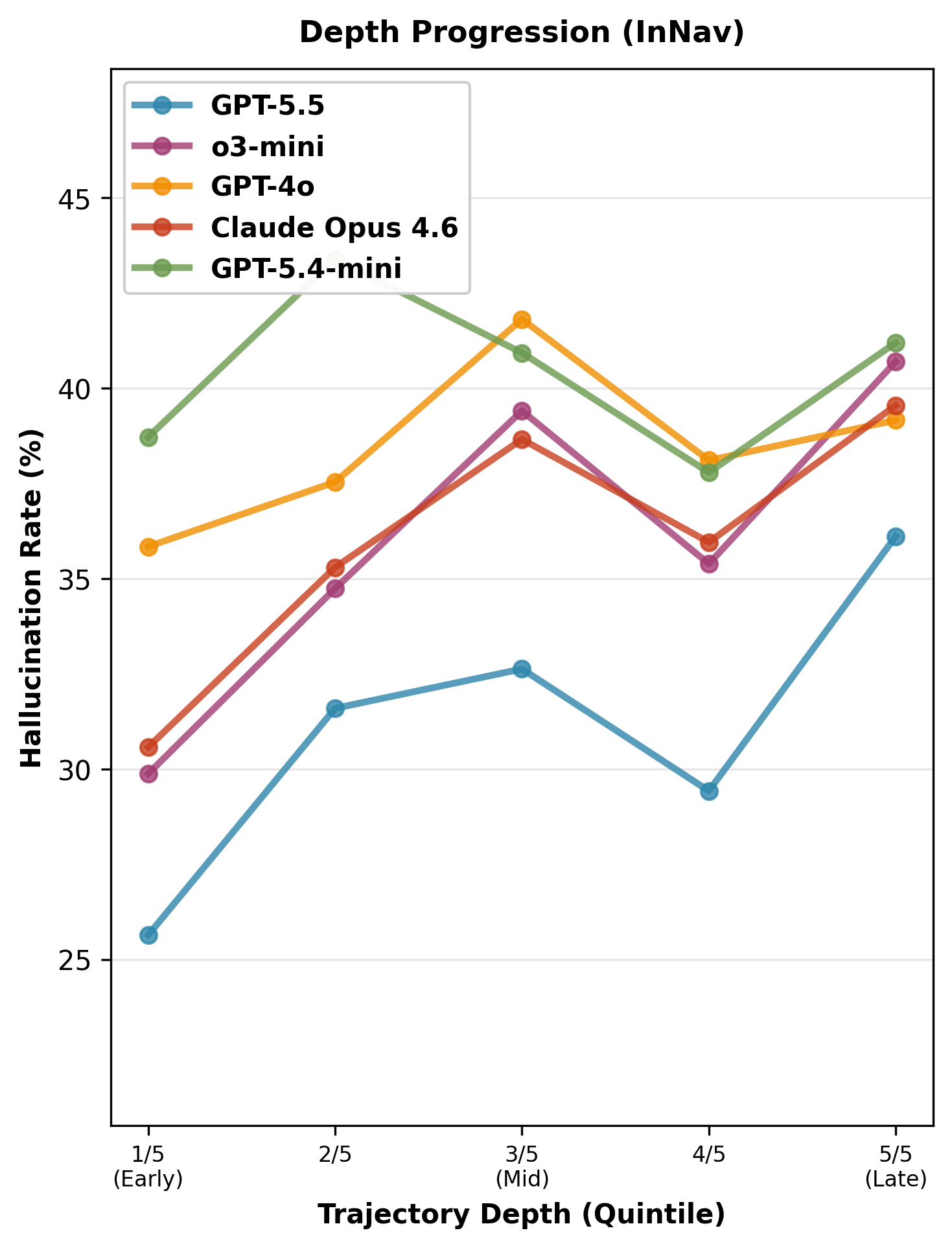}
    \caption{Depth progression for \gridname (select models shown for clarity). \textsc{InNav} hallucinations increase with depth, illustrating how world understanding is harder as models explore deeper.}
    \label{fig:depth-prog}
\end{subfigure}
\caption{\textbf{Navigation impact on trajectory depth effects.} (a) All 13 models show differential slopes between \textsc{InNav} and \textsc{CtrlStatic} modes, revealing model-specific balance between epistemic grounding and cognitive load. Positive difference (red) indicates cognitive load dominates; negative (green) indicates grounding dominates. Frontier reasoning models paradoxically show MORE load accumulation, while efficiency-optimized models show grounding mitigation. (b) Representative models illustrate depth progression patterns: all show increasing hallucination rates over trajectory quintiles, demonstrating that states become intrinsically harder to model as trajectories progress.}
\label{fig:depth-comparison}
\end{figure}

\textbf{Key finding:} All 13 models show positive slopes in \textbf{both} In-Navigation and Controlled Static modes (Figure~\ref{fig:depth-comparison}), indicating that states become intrinsically harder to model as trajectories progress regardless of whether the model is navigating or passively observing. The median slope is $+5.1\%$ per quintile in Controlled Static versus $+6.6\%$ in In-Navigation, corresponding to approximately 20--26\% higher hallucination rates at trajectory endpoints compared to starting positions.

However, the \textit{differential} slopes reveal model-specific navigation strategies (Figure~\ref{fig:depth-comparison}):
\begin{tight_itemize}
\item \textbf{6 models show cognitive load accumulation} (red bars): \textsc{InNav} slopes exceed \textsc{CtrlStatic} by +1.4\% to +5.9\%/quintile (e.g., o3-mini: $+11.2\%$ vs. $+5.3\%$, Opus 4.6: $+9.3\%$ vs. $+3.6\%$), suggesting navigation amplifies depth effects.
\item \textbf{7 models show grounding mitigation} (green bars): \textsc{InNav} slopes are lower than \textsc{CtrlStatic} by 0.7\% to 4.4\%/quintile (e.g., GPT-5.4-mini: $+0.7\%$ vs. $+5.1\%$, GLM-5: $+1.7\%$ vs. $+4.8\%$), indicating navigation mitigates depth effects through epistemic grounding.
\end{tight_itemize}


\textbf{i) Implications for Deployment:} Models cluster into two strategies for long-horizon tasks. Grounding-mitigation models (green bars) actively leverage navigation feedback to counteract complexity accumulation, making them more suitable for extended autonomous operation. Cognitive-load-accumulation models (red bars) show compounding errors over time despite strong static performance, requiring more frequent human intervention or checkpointing in embodied settings.

\textbf{ii) Implications for Evaluation:} Static benchmarks systematically underestimate depth-related failures in navigation-capable models. The universal positive slopes ($+5.1$ to $+6.6\%$ per quintile) mean hallucination rates nearly double from trajectory start to end, yet this effect manifests differently under navigation, some models adapt through epistemic grounding, others deteriorate further through cognitive load. Future embodied AI evaluations must measure performance across trajectory depth to capture these dynamics.


\subsection{Trajectory Depth Analysis by World Category}
\label{sec:depth-by-category}
We observe that the effect of depth on \textsc{InNav} hallucination varies substantially by category (Figure~\ref{fig:cognitive_load_by_category}; we defer detailed analysis to \S\ref{appendix:cognitive_load_by_category}). Perception (-1.0\%/quintile) and Causal worlds (-1.8\%/quintile) show \textit{decreasing} hallucination with trajectory depth, indicating \textbf{epistemic grounding dominates}. Memory and Uncertainty worlds show minimal depth effects (+0.2\%/quintile), with grounding and cognitive load balanced. X-level multi-zone worlds exhibit massive accumulation (+6.5\%/quintile), where \textbf{cognitive load dominates}: long trajectories through many zones overwhelm the LM's working memory. 

\subsection{In-Navigation Serialization Sensitivity Analysis}
\label{sec:serialization-sensitivity}
To check if the better-performing serialization persists in the \textsc{InNav} case, we compare canonical serializers listed in Table \ref{tab:level-reference} with alternative ones. Two categories showed striking inversions: \textbf{Perception} (grid canonical) favored symbolic by 14.3\% on average (5/6 worlds), while \textbf{Uncertainty} (memory canonical) favored grid by 12.8\% on average (4/5 worlds), showing some serializers support better epistemic grounding even though suboptimal for answering probes in isolation. We defer full detail to \S\ref{appendix:serialization-comparison}. 

\subsection{Navigation Effect by World Category}
\label{appendix:nav-by-category}

Table~\ref{tab:nav-category} breaks down the navigation effect by world category. Navigation impact varies substantially by category: Uncertainty worlds show the strongest overall benefit (median $-6.0\%$), suggesting navigation helps models resolve partial observability. X-Levels show large model-dependent effects, with GLM-5 ($-12.6\%$) and GPT-5.5 ($-11.2\%$) benefiting substantially while weaker models show slight costs. Memory worlds exhibit mixed effects (DeepSeek-V3: $+10.2\%$ vs GPT-5.5: $-9.6\%$), suggesting complex architecture-dependent interactions between memory demands and navigation. Perception worlds show minimal navigation effect (median of $-0.3\%$), indicating visual-spatial reasoning operates largely independently of navigation demands.

\begin{table}[ht]
\centering
\caption{%
  \textbf{\gridname Navigation effect by world category.}
  Values show In-Navigation hallucination rate minus Controlled Static rate.
  Negative values indicate navigation helps (reduces hallucinations).
  Categories: P\,=\,Perception, C\,=\,Causal, M\,=\,Memory, U\,=\,Uncertainty, X\,=\,Cross-category.
}
\label{tab:nav-category}
\small
\setlength{\tabcolsep}{4pt}
\begin{tabular}{lrrrrrr}
\toprule
\textbf{Model}
  & \textbf{Overall}
  & \textbf{P}
  & \textbf{C}
  & \textbf{M}
  & \textbf{U}
  & \textbf{X} \\
\midrule
GPT-5.5          & \cellcolor{pastelgreen!100!pastelred}\textbf{$-13.2$} & \cellcolor{pastelgreen!32!pastelred}$+2.0$ & \cellcolor{pastelgreen!98!pastelred}\textbf{$-15.8$} & \cellcolor{pastelgreen!91!pastelred}\textbf{$-9.6$} & \cellcolor{pastelgreen!100!pastelred}\textbf{$-16.9$} & \cellcolor{pastelgreen!100!pastelred}\textbf{$-13.3$} \\
GLM-5            & \cellcolor{pastelgreen!58!pastelred}$-4.4$  & \cellcolor{pastelgreen!49!pastelred}$-0.3$ & \cellcolor{pastelgreen!50!pastelred}$-4.3$  & \cellcolor{pastelgreen!36!pastelred}$+2.0$  & \cellcolor{pastelgreen!49!pastelred}$-4.9$  & \cellcolor{pastelgreen!97!pastelred}$-12.6$ \\
o3-mini          & \cellcolor{pastelgreen!58!pastelred}$-4.4$  & \cellcolor{pastelgreen!51!pastelred}$-0.6$ & \cellcolor{pastelgreen!58!pastelred}$-6.0$  & \cellcolor{pastelgreen!62!pastelred}$-3.3$  & \cellcolor{pastelgreen!68!pastelred}$-9.3$  & \cellcolor{pastelgreen!58!pastelred}$-3.1$ \\
o4-mini          & \cellcolor{pastelgreen!55!pastelred}$-3.8$  & \cellcolor{pastelgreen!60!pastelred}\textbf{$-2.1$} & \cellcolor{pastelgreen!44!pastelred}$-2.9$  & \cellcolor{pastelgreen!64!pastelred}$-3.6$  & \cellcolor{pastelgreen!56!pastelred}$-6.6$  & \cellcolor{pastelgreen!62!pastelred}$-4.3$ \\
o3               & \cellcolor{pastelgreen!52!pastelred}$-3.2$  & \cellcolor{pastelgreen!49!pastelred}$-0.3$ & \cellcolor{pastelgreen!36!pastelred}$-1.2$  & \cellcolor{pastelgreen!62!pastelred}$-3.2$  & \cellcolor{pastelgreen!61!pastelred}$-7.7$  & \cellcolor{pastelgreen!62!pastelred}$-4.3$ \\
DeepSeek-V3-0324 & \cellcolor{pastelgreen!48!pastelred}$-2.4$  & \cellcolor{pastelgreen!41!pastelred}$+0.7$ & \cellcolor{pastelgreen!70!pastelred}$-8.7$  & \cellcolor{pastelgreen!0!pastelred}$+10.2$ & \cellcolor{pastelgreen!63!pastelred}$-8.1$  & \cellcolor{pastelgreen!46!pastelred}$-0.1$ \\
GPT-4o           & \cellcolor{pastelgreen!43!pastelred}$-1.3$  & \cellcolor{pastelgreen!58!pastelred}$-1.8$ & \cellcolor{pastelgreen!57!pastelred}$-5.8$  & \cellcolor{pastelgreen!41!pastelred}$+0.9$  & \cellcolor{pastelgreen!38!pastelred}$-2.1$  & \cellcolor{pastelgreen!30!pastelred}$+3.3$ \\
GPT-4o-mini      & \cellcolor{pastelgreen!38!pastelred}$-0.3$  & \cellcolor{pastelgreen!24!pastelred}$+3.9$ & \cellcolor{pastelgreen!54!pastelred}$-5.1$  & \cellcolor{pastelgreen!18!pastelred}$+6.1$  & \cellcolor{pastelgreen!49!pastelred}$-4.9$  & \cellcolor{pastelgreen!39!pastelred}$+0.9$ \\
Claude Opus 4.6  & \cellcolor{pastelgreen!31!pastelred}$+1.1$  & \cellcolor{pastelgreen!16!pastelred}$+5.6$ & \cellcolor{pastelgreen!25!pastelred}$+1.1$  & \cellcolor{pastelgreen!37!pastelred}$+1.9$  & \cellcolor{pastelgreen!55!pastelred}$-6.3$  & \cellcolor{pastelgreen!34!pastelred}$+2.2$ \\
Claude Sonnet 4.6& \cellcolor{pastelgreen!31!pastelred}$+1.2$  & \cellcolor{pastelgreen!52!pastelred}$-0.8$ & \cellcolor{pastelgreen!22!pastelred}$+1.7$  & \cellcolor{pastelgreen!27!pastelred}$+4.1$  & \cellcolor{pastelgreen!27!pastelred}$+0.6$  & \cellcolor{pastelgreen!38!pastelred}$+1.1$ \\
Kimi K2.6        & \cellcolor{pastelgreen!30!pastelred}$+1.3$  & \cellcolor{pastelgreen!27!pastelred}$+3.1$ & \cellcolor{pastelgreen!34!pastelred}$-0.9$  & \cellcolor{pastelgreen!33!pastelred}$+2.7$  & \cellcolor{pastelgreen!40!pastelred}$-2.6$  & \cellcolor{pastelgreen!26!pastelred}$+4.3$ \\
GPT-5.4-mini     & \cellcolor{pastelgreen!22!pastelred}$+2.9$  & \cellcolor{pastelgreen!17!pastelred}$+5.5$ & \cellcolor{pastelgreen!32!pastelred}$-0.5$  & \cellcolor{pastelgreen!21!pastelred}$+5.3$  & \cellcolor{pastelgreen!16!pastelred}$+3.2$  & \cellcolor{pastelgreen!32!pastelred}$+2.7$ \\
\bottomrule
\end{tabular}
\end{table}

\subsection{More Details: Trajectory Depth Analysis by World Category}
\label{appendix:cognitive_load_by_category}

\begin{figure}[p]
\centering
\begin{subfigure}[b]{0.48\textwidth}
    \includegraphics[width=\textwidth]{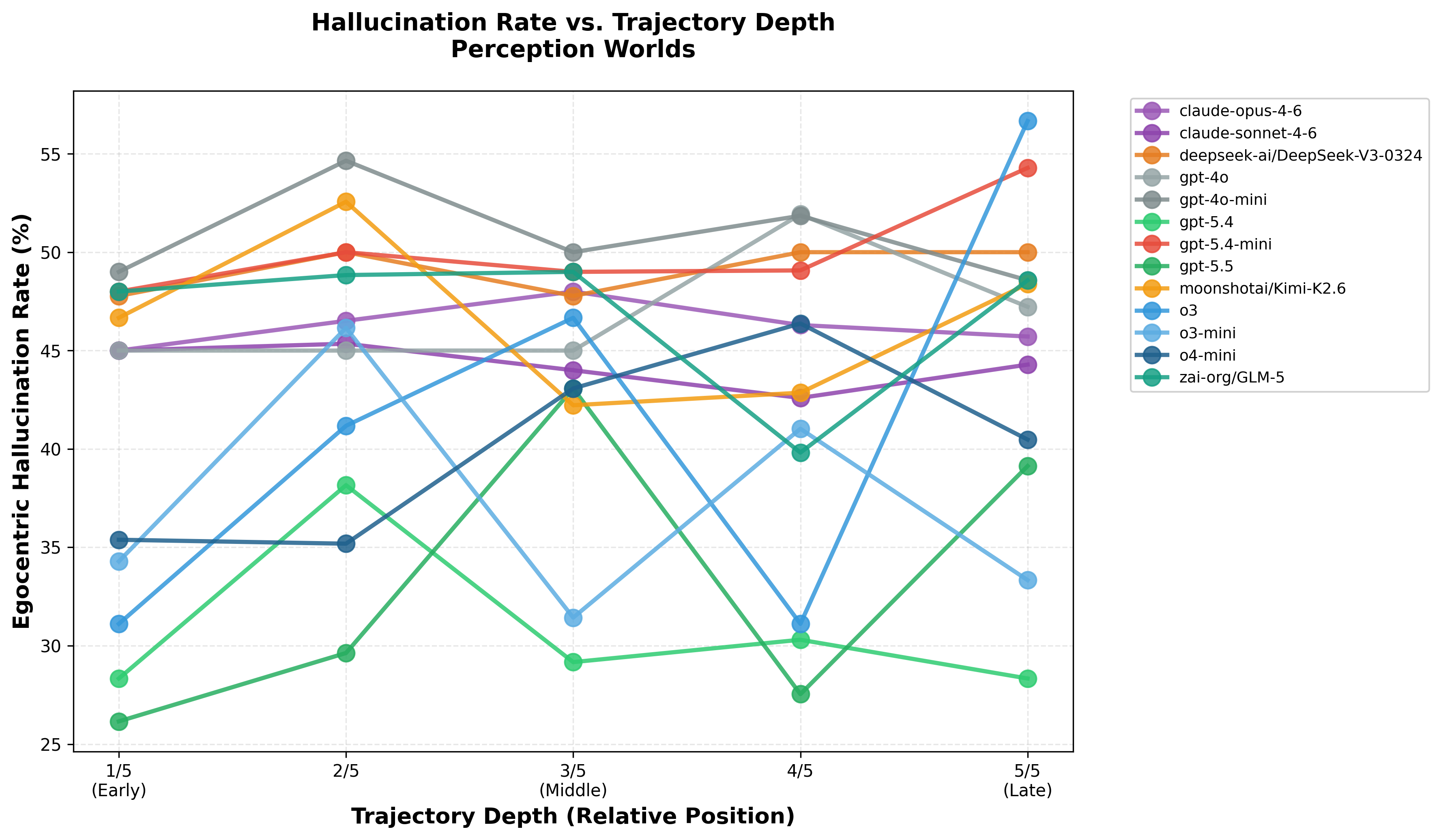}
    \caption{Perception worlds (-1.0\%/quintile)}
    \label{fig:cog_load_perception}
\end{subfigure}
\hfill 
\begin{subfigure}[b]{0.48\textwidth}
    \includegraphics[width=\textwidth]{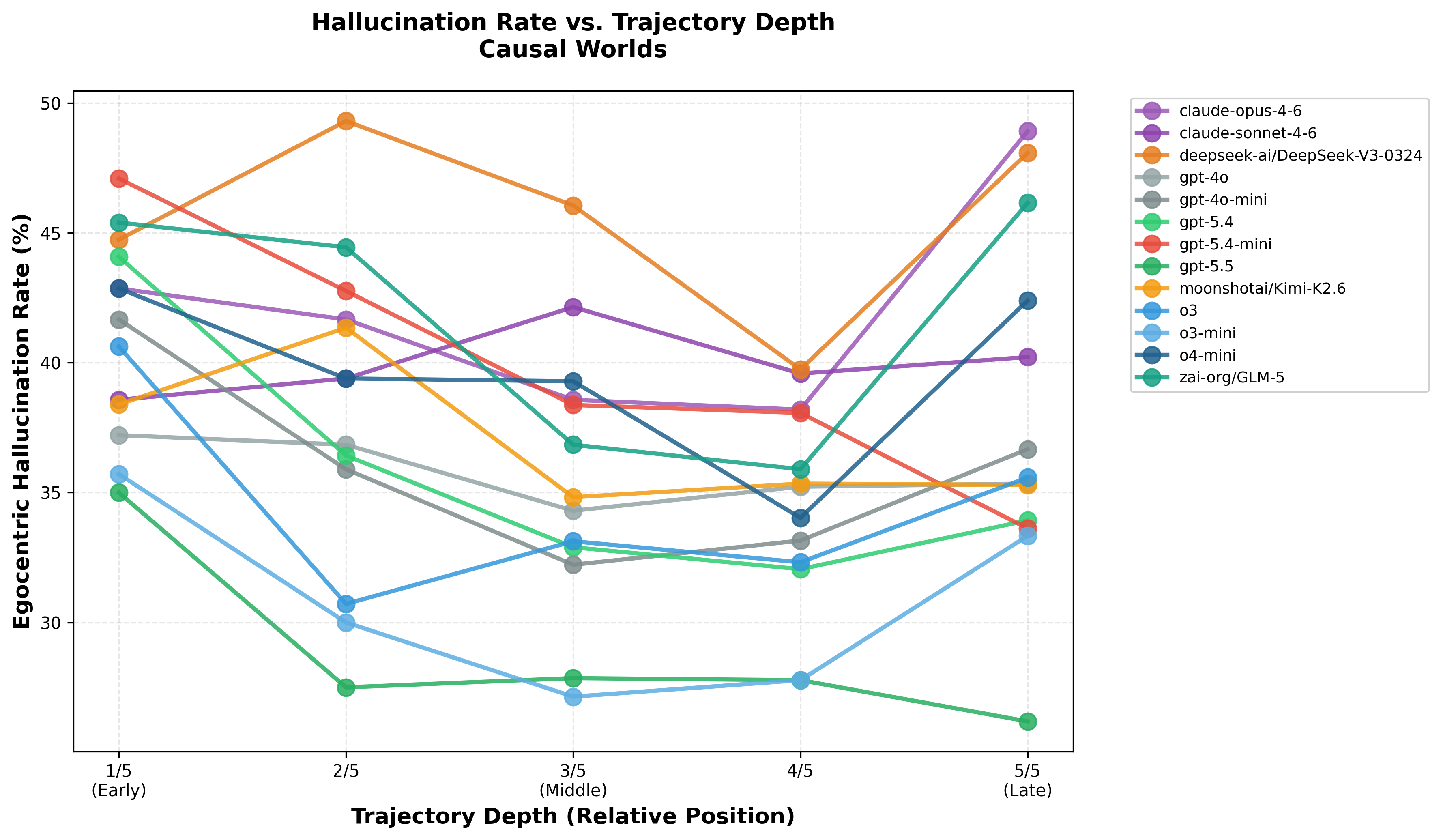}
    \caption{Causal worlds (-1.8\%/quintile)}
    \label{fig:cog_load_causal}
\end{subfigure}

\vspace{0.5cm}

\begin{subfigure}[b]{0.48\textwidth}
    \includegraphics[width=\textwidth]{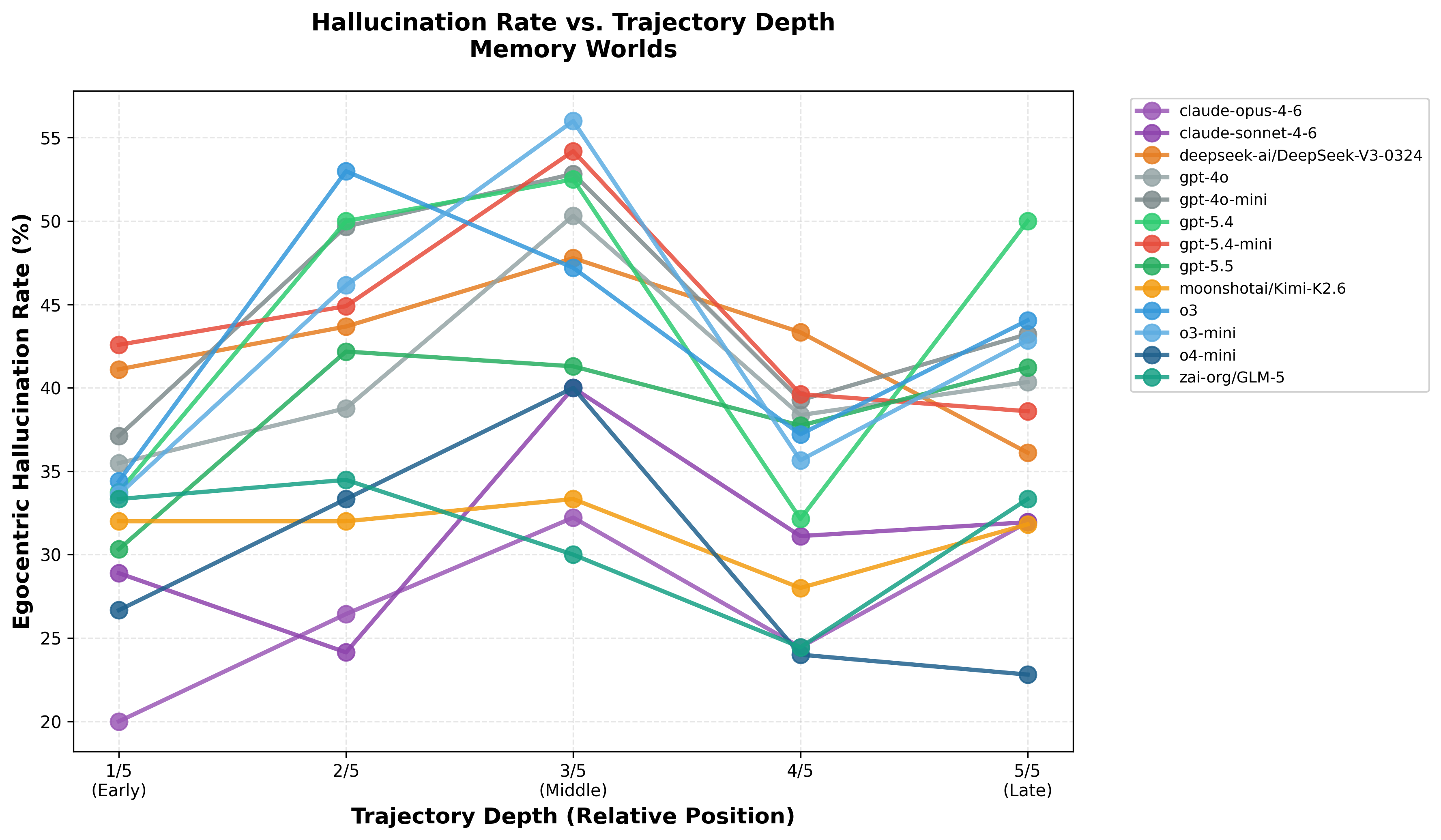}
    \caption{Memory worlds (+0.2\%/quintile)}
    \label{fig:cog_load_memory}
\end{subfigure}
\hfill
\begin{subfigure}[b]{0.48\textwidth}
    \includegraphics[width=\textwidth]{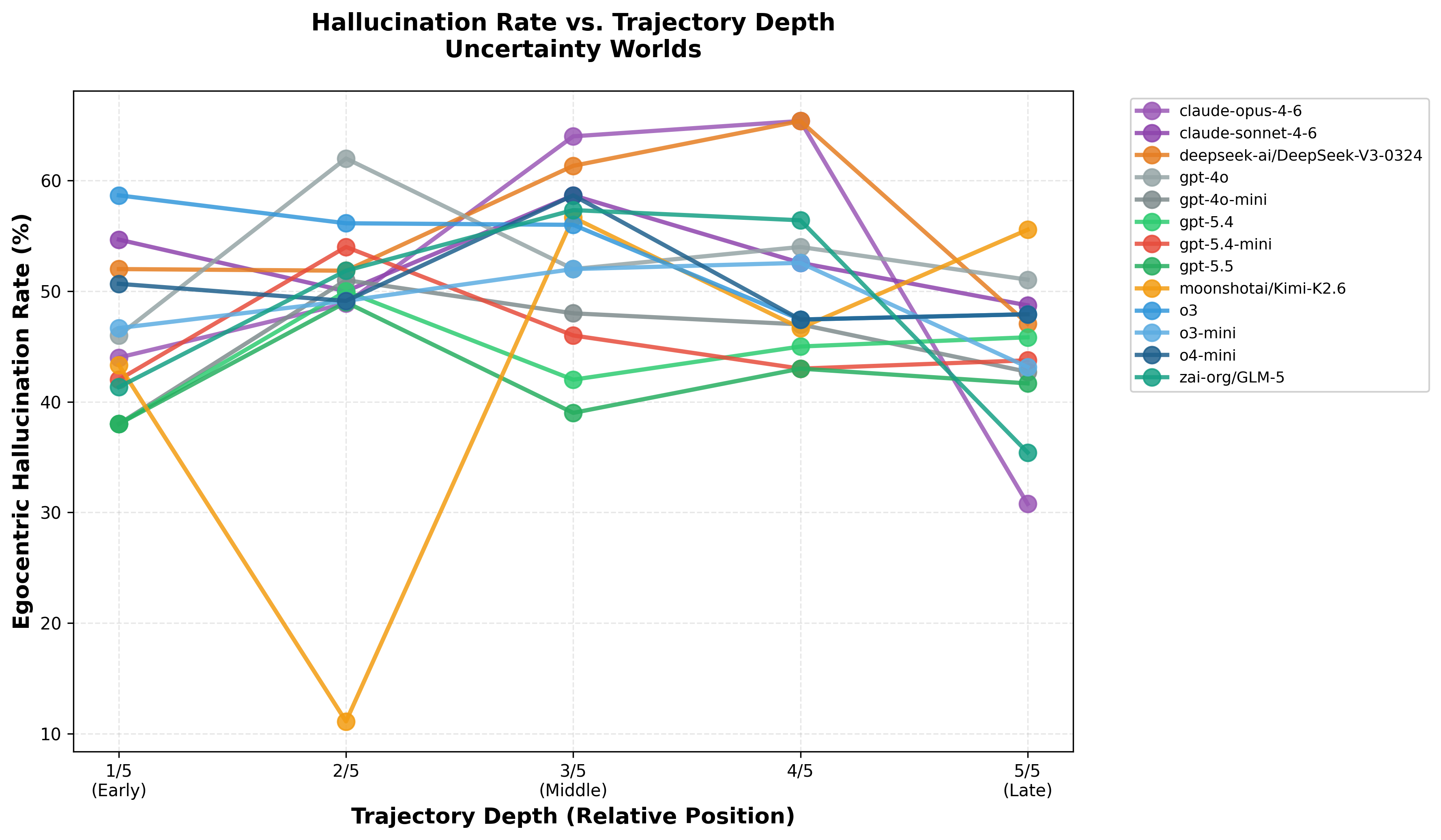}
    \caption{Uncertainty worlds (+0.2\%/quintile)}
    \label{fig:cog_load_uncertainty}
\end{subfigure}
\vspace{0.5cm}
\begin{subfigure}[b]{0.48\textwidth}
    \includegraphics[width=\textwidth]{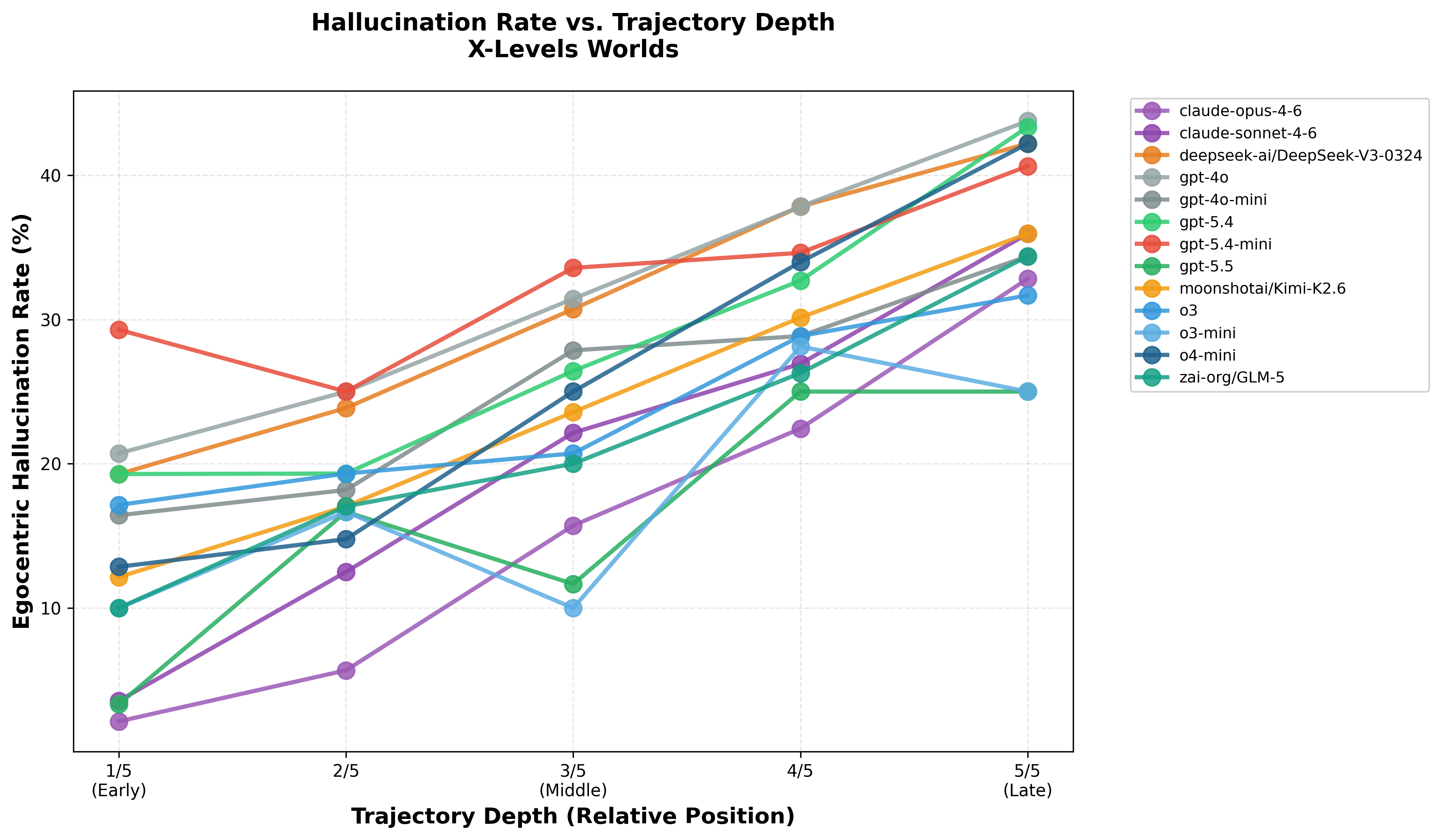}
    \caption{X-level multi-zone worlds (+6.5\%/quintile)}
    \label{fig:cog_load_xlevels}
\end{subfigure}

\caption{\textbf{Trajectory depth effects by world category for \gridname.} Depth dynamics vary considerably across domains, reflecting the balance between epistemic grounding and cognitive load. Perception and Causal worlds show reversed effects (grounding dominates: navigation helps), Memory and Uncertainty worlds show minimal depth effects (balanced), while X-level multi-zone facilities show large cognitive load accumulation (+6.5\% per quintile: cognitive load dominates). This reveals that navigation's impact is category-dependent.}
\label{fig:cognitive_load_by_category}
\end{figure}

To investigate whether the balance between epistemic grounding and cognitive load varies across reasoning domains, we decompose our trajectory depth analysis by world category. Figure~\ref{fig:cognitive_load_by_category} presents \textsc{InNav} hallucination rates versus trajectory depth (quintiles 1/5 through 5/5) separately for Perception, Causal, Memory, Uncertainty, and X-level worlds. Each category exhibits distinct depth dynamics, revealing that navigation's net impact depends critically on the nature of the reasoning task.

\subsubsection{Perception Worlds (-1.0\% per quintile)}
Perception-focused worlds (P1-P5; 5,066 probes across 6 worlds) test spatial reasoning, object tracking, and visual memory. Counter to the overall trend, these worlds show \textbf{decreasing hallucination} as trajectories deepen (average slope: -1.0\% per quintile). This reversed pattern indicates that \textbf{epistemic grounding dominates}: additional observations help models resolve perceptual ambiguities by providing disambiguating evidence about spatial relationships and object properties, reducing uncertainty over time. The effect is particularly pronounced in P4\_harder\_array and P1\_dense\_array, where systematic exploration helps models build accurate spatial representations incrementally. Cognitive load from navigation is outweighed by the grounding benefit of multiple viewpoints.

\subsubsection{Causal Worlds (-1.8\% per quintile)}
Causal reasoning worlds (C1a-C6, C1b\_continuous\_chain; 8,964 probes across 8 worlds) test understanding of cause-effect relationships, state transitions, and action consequences. These worlds exhibit the strongest \textbf{epistemic grounding effect} (average slope: -1.8\% per quintile), with models showing consistent improvement as trajectories progress. This pattern indicates that navigation provides critical disambiguating evidence for causal inference: observing the results of actions (e.g., ``Does pushing this boulder open the path?'', ``Is the fire barrier passable after the lever is pulled?'') allows models to refine their causal models over time. The effect is most pronounced in C2\_fire\_crossing and C3\_flood\_room, where state changes must be observed to answer probes correctly. Grounding from action-perception feedback loops substantially outweighs cognitive load.

\subsubsection{Memory Worlds (+0.2\% per quintile)}
Memory-focused worlds (M1--M4; 6,900 probes across 4 worlds) test retention of past observations, event sequencing, and testimony integration. These worlds show minimal depth effect (average slope: +0.2\% per quintile), nearly flat across trajectory positions. This suggests \textbf{epistemic grounding and cognitive load roughly balance}: while models must track more events as navigation proceeds (increasing cognitive load), additional observations provide compensating context (epistemic grounding). The incremental cost is small. M2\_witness\_stand and M3\_incident\_report show the weakest effects, likely because structured testimony provides explicit memory cues that reduce working memory load while simultaneously grounding responses.

\subsubsection{Uncertainty Worlds (+0.2\% per quintile)}
Uncertainty-focused worlds (U1, U2\_oracle\_*, U4; 4,898 probes across 5 worlds) test reasoning under partial observability, probabilistic inference, and information-seeking behavior. Like Memory worlds, these show minimal depth effect (+0.2\% per quintile). The flat trend suggests models handle increasing trajectory length well: fog-of-war restrictions (U1) and oracle limitations (U2) impose constant rather than accumulating cognitive load, while additional observations provide grounding for probabilistic reasoning. However, U4\_amnesiac shows slightly stronger accumulation (+0.8\%), indicating that memory erasure creates compounding difficulties that overwhelm grounding benefits.

\subsubsection{X-Level Multi-Zone Worlds (+6.5\% per quintile)}
X-level compound worlds (X1--X7; 6,860 probes across 7 worlds) feature multi-room navigation through 3--8 connected zones with complex state tracking requirements. These worlds show \textbf{massive cognitive load accumulation} (average slope: +6.5\% per quintile), more than 5× the overall average. Long trajectories through multiple zones overwhelm working memory, causing dramatic hallucination increases at later trajectory positions. \textbf{Cognitive load dominates epistemic grounding} in these complex multi-zone environments. The effect is strongest in X3\_facility\_7zone (+8.2\% per quintile) and X7\_dragon\_keep (+7.8\%), where traversing many rooms creates compounding memory demands. X1\_facility\_3zone shows the weakest accumulation (+4.1\%), confirming that zone count drives cognitive load.

\subsubsection{Summary and Implications}
These category-specific patterns reveal that depth effects reflect the \textbf{balance between epistemic grounding and cognitive load}, which varies by category. We identify three distinct regimes:

\begin{enumerate}
\item \textbf{Grounding-dominated regimes} (Perception, Causal): Negative slopes (-1.0\% to -1.8\% per quintile) indicate that additional observations provide more benefit (disambiguation of spatial/causal relationships) than cost (working memory load). Navigation actively \textit{helps} state tracking in these domains.

\item \textbf{Balanced regimes} (Memory, Uncertainty): Near-zero slopes (+0.2\% per quintile) indicate that grounding benefits and cognitive load costs roughly cancel. Models handle increasing trajectory length without substantial performance degradation, suggesting efficient resource allocation for these task types.

\item \textbf{Load-dominated regimes} (X-Levels): Large positive slopes (+6.5\% per quintile) indicate that working memory costs from multi-zone navigation overwhelm any grounding benefits. The super-linear growth (5× overall average) suggests compounding interference rather than linear scaling.
\end{enumerate}

This category-dependence has practical implications for agent deployment: models should be deployed with navigation in perceptual/causal tasks (where it helps), can navigate freely in memory/uncertainty tasks (neutral impact), but should minimize navigation steps in multi-zone environments (where it hurts). The patterns also suggest distinct underlying mechanisms: perceptual and causal reasoning benefit from action-perception feedback loops (grounding effect), while spatial navigation across many rooms imposes compounding working memory costs (cognitive load effect). Future work should investigate whether these patterns reflect fundamental differences in how models allocate cognitive resources across task types, and whether architectural interventions (e.g., external memory, spatial memory modules) can mitigate load-dominated regimes.

\subsection{More Details: Serialization Sensitivity Analysis}
\label{appendix:serialization-comparison}

To validate our canonical serialization choices, we systematically compared alternative serializations on worlds where multiple representations were available during \textsc{InNav} evaluations. This analysis addresses a critical methodological question: are canonical serializers universally optimal during navigation, or do certain world types benefit from alternative representations?

\subsubsection{Methodology}

\textbf{Coverage:} We identified 30 worlds tested with multiple serializers:
\begin{itemize}
\item 18 worlds with grid and memory (Causal, Memory, Uncertainty categories)
\item 11 worlds with grid and symbolic (Perception and X-level categories)
\item 1 world with memory and symbolic (M2\_witness\_stand)
\end{itemize}

\textbf{Models:} For each world-serializer pair, we included only models tested on both serializers, yielding 229 model-level comparisons across 30 worlds.

\textbf{Aggregation:} To ensure fair comparison despite different navigation trajectories per serializer, we use hierarchical aggregation:
\begin{enumerate}
\item \textbf{Probe $\rightarrow$ Trace:} Group probes by episode seed, calculate hallucination rate per trace
\item \textbf{Trace $\rightarrow$ World:} Average across traces (equal weight per episode)
\item \textbf{World $\rightarrow$ Serializer:} Compare world-level hallucination rates
\end{enumerate}

This prevents bias from trajectory length differences (grid navigation may take more/fewer steps than memory navigation).

\begin{table}[h]
\centering
\small
\setlength{\tabcolsep}{6pt}
\caption{\textbf{Serialization comparison by world category for \gridname.} Canonical serializers win only half of In-Navigation comparisons, with systematic inversions in Perception and Uncertainty categories. ``Best Non-Canonical'' shows the serializer that outperformed the canonical choice, with average margin in percentage points.}
\label{tab:serialization-summary}
\begin{tabular}{@{}lccclr@{}}
\toprule
\textbf{Category} & \textbf{Canonical} & \textbf{Worlds} & \textbf{InNav Wins} & \textbf{Best Non-Canon.} & \textbf{Avg Margin} \\
\midrule
Perception       & grid     & 6  & 1/6 (16.7\%) & \textbf{symbolic} & 14.3\% \\
Uncertainty      & memory   & 5  & 1/5 (20.0\%) & \textbf{grid}     & 12.8\% \\
Causal           & memory   & 8  & 5/8 (62.5\%) & grid              & 3.6\%  \\
Memory           & mixed    & 4  & 2/4 (50.0\%) & symbolic          & 8.2\%  \\
X-level          & symbolic & 7  & 6/7 (85.7\%) & ---               & ---    \\
\midrule
\textbf{Overall} & ---      & 30 & 15/30 (50.0\%) & ---             & ---    \\
\bottomrule
\end{tabular}
\end{table}

\begin{figure}
\centering
\includegraphics[width=0.75\textwidth]{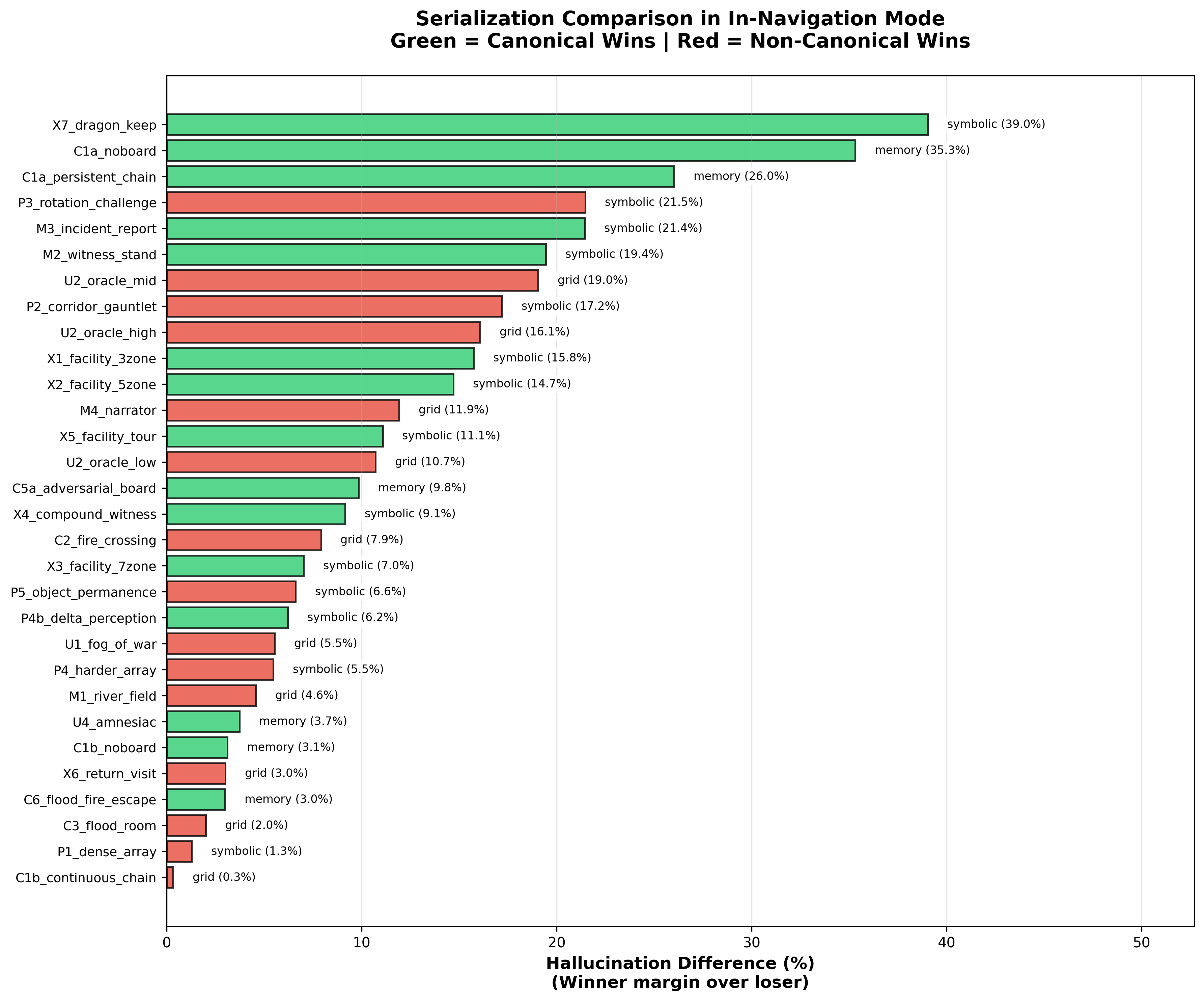}
\caption{\textbf{Serialization comparison in In-Navigation mode for \gridname.} Each bar shows the hallucination difference between the winning and losing serializer, sorted by margin. Green indicates canonical serializer wins (15/30 worlds); red indicates non-canonical wins (15/30 worlds). Notable inversions include Perception worlds (symbolic dominates despite grid being canonical: P3\_rotation\_challenge 21.5\%, P2\_corridor\_gauntlet 17.2\%) and Uncertainty worlds (grid dominates despite memory being canonical: U2\_oracle\_mid 19.0\%, U2\_oracle\_high 16.1\%). X-level worlds validate the canonical symbolic choice (6/7 wins), with X7\_dragon\_keep showing the largest margin (39.0\%).}
\label{fig:serialization-comparison}
\end{figure}

\subsubsection{Overall Results}

Table~\ref{tab:serialization-summary} summarizes results by world category. Canonical serializers won only 50.0\% of \textsc{InNav} comparisons (15 of 30 worlds), revealing that no single serialization is universally optimal during navigation. The pattern varies substantially by category, with systematic inversions in Perception and Uncertainty worlds.

The inversions are particularly striking: in Perception worlds where grid was chosen as canonical based on prior evaluation, symbolic representations reduced hallucination by an average of 14.3 percentage points across models, winning 5 of 6 worlds. Similarly, Uncertainty worlds showed a 12.8 percentage point advantage for grid over the canonical memory choice, winning 4 of 5 worlds. These margins are substantial - comparable to or exceeding the navigation effect itself (Section~\ref{appendix:nav-by-category}) - indicating that serialization choice can be as impactful as the evaluation mode during navigation-based reasoning.

\subsubsection{Category-Specific Analysis}

\paragraph{Perception Worlds: Symbolic Dominance}

Perception worlds (P1--P5) test spatial reasoning, object tracking, and visual memory. Despite grid being canonical, symbolic representations achieved lower hallucination in 5 of 6 worlds during navigation:

\begin{itemize}
\item \textbf{P3\_rotation\_challenge:} Symbolic wins by 21.5\%
\item \textbf{P2\_corridor\_gauntlet:} Symbolic wins by 17.2\%
\item \textbf{P5\_object\_permanence:} Symbolic wins by 6.6\%
\item \textbf{P4\_harder\_array:} Symbolic wins by 5.5\%
\item \textbf{P1\_dense\_array:} Symbolic wins by 1.3\%
\end{itemize}

Only P4b\_delta\_perception favored the canonical grid (6.2\% margin). This systematic preference suggests that during navigation, language-based spatial descriptions (``\textit{Ahead: red ball. Left: blue box. Right: wall.}'') provide more effective grounding than raw ASCII grid layouts for perceptual reasoning. The magnitude of this effect is substantial: the 21.5\% margin on P3\_rotation\_challenge exceeds most model-to-model differences, indicating serialization choice can matter more than model selection for certain tasks.

\paragraph{Uncertainty Worlds: Grid Dominance Inspite of Memory Omniscience}

Uncertainty worlds (U1, U2\_oracle\_*, U4) test reasoning under partial observability with fog-of-war, oracle access limits, and memory erasure. Despite memory being canonical, grid representations achieved lower hallucination in 4 of 5 worlds:

\begin{itemize}
\item \textbf{U2\_oracle\_mid:} Grid wins by 19.0\%
\item \textbf{U2\_oracle\_high:} Grid wins by 16.1\%
\item \textbf{U2\_oracle\_low:} Grid wins by 10.7\%
\item \textbf{U1\_fog\_of\_war:} Grid wins by 5.5\%
\end{itemize}

Only U4\_amnesiac favored memory (3.7\% margin). This pattern suggests that explicit spatial layouts help models reason about unobserved regions more effectively than textual memory descriptions during navigation. Grids may provide clearer spatial invariants (``I know what's at (5,7) even if I can't see it now'') compared to memory's sequential event descriptions.

\paragraph{X-Level Worlds: Symbolic Validated}

X-level multi-zone worlds (X1--X7) test navigation across 3--8 connected rooms with witness testimony and complex spatial reasoning. Symbolic representations (canonical choice) achieved lower hallucination in 6 of 7 worlds:

\begin{itemize}
\item \textbf{X7\_dragon\_keep:} Symbolic wins by 39.0\%
\item \textbf{X1\_facility\_3zone:} Symbolic wins by 15.8\%
\item \textbf{X2\_facility\_5zone:} Symbolic wins by 14.7\%
\item \textbf{X5\_facility\_tour:} Symbolic wins by 11.1\%
\item \textbf{X4\_compound\_witness:} Symbolic wins by 9.1\%
\item \textbf{X3\_facility\_7zone:} Symbolic wins by 7.0\%
\end{itemize}

Only X6\_return\_visit favored grid (3.0\% margin). The strong symbolic advantage validates our canonical choice: reasoning about multi-room environments benefits from higher-level linguistic abstractions that describe room connectivity, zone properties, and witness locations more naturally than low-level grid coordinates.

\paragraph{Causal and Memory Worlds: Mixed Results}

Causal worlds (memory canonical) showed 5/8 wins for memory, with grid competitive on dynamic state-change worlds (C2\_fire\_crossing, C3\_flood\_room). Memory worlds showed mixed patterns, with symbolic winning on testimony-based worlds (M2\_witness\_stand: 19.4\%, M3\_incident\_report: 21.4\%) but grid winning on M4\_narrator (11.9\%).

\subsubsection{Model-Specific Patterns}

Different models show varying serialization preferences (Table~\ref{tab:model-serialization-prefs}). GPT-4o and GPT-4o-mini exhibit the strongest symbolic preference (66.7\% win rate), while Opus 4.6 shows more balanced performance across serializers. O4-mini demonstrates the most uniform distribution (11 symbolic, 9 grid, 9 memory wins), suggesting more serialization-agnostic reasoning.

\begin{table}[h]
\centering
\small
\setlength{\tabcolsep}{5pt}
\caption{\textbf{Model-specific serialization preferences (In-Navigation mode) for \gridname.} Win counts across all world-serializer comparisons. GPT-4o models show strong symbolic preference; Claude models are more balanced.}
\label{tab:model-serialization-prefs}
\begin{tabular}{@{}lccc@{}}
\toprule
\textbf{Model} & \textbf{Symbolic Wins} & \textbf{Grid Wins} & \textbf{Memory Wins} \\
\midrule
gpt-4o                  & 14/21 (66.7\%) & ---            & 7/21 (33.3\%)  \\
gpt-4o-mini             & 14/21 (66.7\%) & ---            & 7/21 (33.3\%)  \\
o3-mini                 & 14/27 (51.9\%) & 8/27 (29.6\%)  & 5/27 (18.5\%)  \\
claude-sonnet-4-6       & 13/29 (44.8\%) & 7/29 (24.1\%)  & 9/29 (31.0\%)  \\
claude-opus-4-6         & 12/29 (41.4\%) & 11/29 (37.9\%) & 6/29 (20.7\%)  \\
o3                      & 13/27 (48.1\%) & 7/27 (25.9\%)  & 7/27 (25.9\%)  \\
o4-mini                 & 11/29 (37.9\%) & 9/29 (31.0\%)  & 9/29 (31.0\%)  \\
\bottomrule
\end{tabular}
\end{table}

\subsubsection{Implications}

These results have both methodological and practical implications:

\textbf{Methodologically}, they validate our multi-serializer evaluation approach: no single serialization is universally optimal during navigation. Perception and Uncertainty categories show systematic inversions where non-canonical choices substantially outperform canonical ones (14.3\% and 12.8\% average margins).

\textbf{Practically}, they suggest task-dependent deployment guidelines: agents should use symbolic serializers for local perceptual reasoning during navigation (P worlds), grid layouts for spatial reasoning under uncertainty (U worlds), and symbolic for complex multi-zone environments (X worlds).

\textbf{Theoretically}, the 50\% canonical win rate reveals that navigation contexts have distinct representational requirements. Symbolic's linguistic clarity appears to help with action-integrated perceptual grounding, while grid's spatial explicitness helps with uncertainty reasoning. Future work should investigate whether these patterns reflect fundamental differences in how models process spatial vs linguistic representations during active navigation.

\section{\gridname Level Editor and Trajectory Recorder Tools}
\label{appendix:gridworld-tools}
To allow frictionless benchmark extensibility for the community, we provide two complementary tools, a \textit{Level Editor} for prototyping/designing worlds and a \textit{Trajectory Recorder} for ground truth annotation and qualitative analysis of trajectories and vulnerability to hallucination in \gridname.

\subsection{Level Editor: Enabling Rapid World Prototyping}
To both visualize as well as modify existing levels (worlds) and craft new ones, we provide a lightweight but intuitive level editor, available both as a localhost Web UI as well as a ncurses based in-terminal editor. 

The Web UI additonally provides sidebars for object selection, placement and grid state preview. This is reminiscent of the world building UI in RTS video games such as Age of Empires II, Warcraft III etc where players can visually place units, terrains and objectives on a grid canvas.

The editor supports all MiniGrid-compatible objects (walls, doors, keys, balls, boxes) extended with our custom mechanics: fire tiles, flood tiles, pressure plates, notice boards, signposts, river tiles, and dark zones. Researchers can interactively place objects on a grid canvas, configure their properties (e.g., door colors, flood activation steps, notice board text with deliberately false claims), and export environments to plain-text \texttt{.txt} format compatible with \texttt{AsciiEnv}. The editor enforces MiniGrid constraints (e.g., wall borders, valid object placements) and automatically generates level metadata (agent start position, view size, see-through-walls flag).

\begin{figure}[ht]
\centering
\begin{subfigure}[b]{0.32\textwidth}
    \centering
    \includegraphics[width=\textwidth]{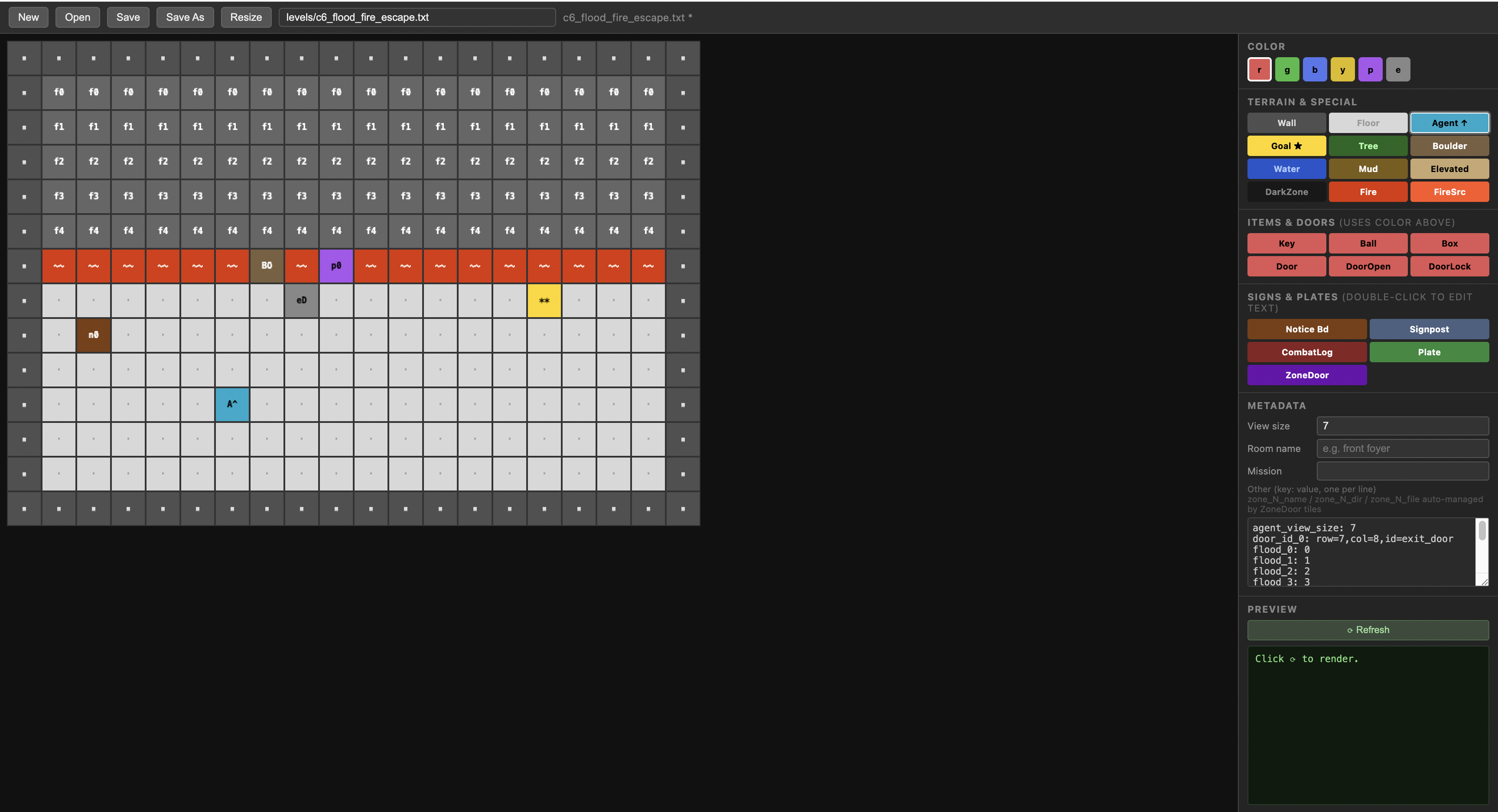}
    \caption{C6 Flood-Fire Escape level on the Web UI, with elements e.g. flood, fire, boulder, plate, door etc.}
    \label{fig:level-editor-web}
\end{subfigure}
\hfill
\begin{subfigure}[b]{0.32\textwidth}
    \centering
    \includegraphics[width=\textwidth]{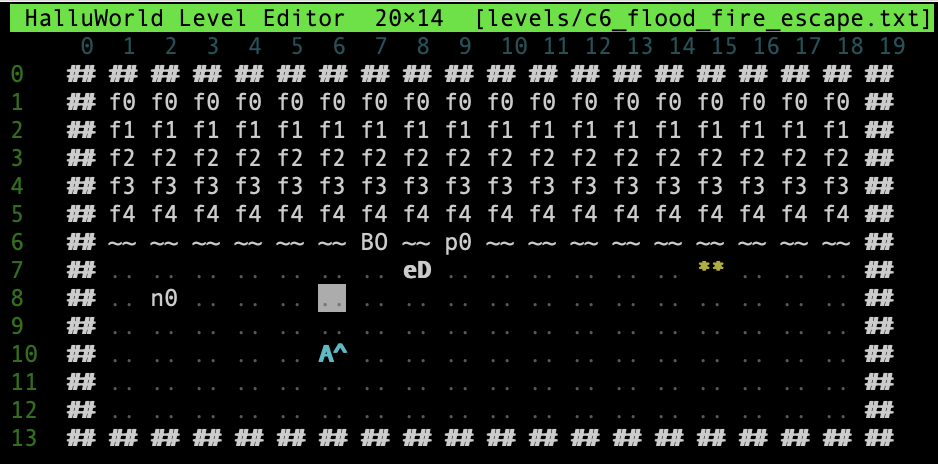}
    \caption{Terminal ncurses editor with keyboard-driven workflow for in-termina editing}
    \label{fig:level-editor-ncurses}
\end{subfigure}
\hfill
\begin{subfigure}[b]{0.32\textwidth}
    \centering
    \includegraphics[width=\textwidth]{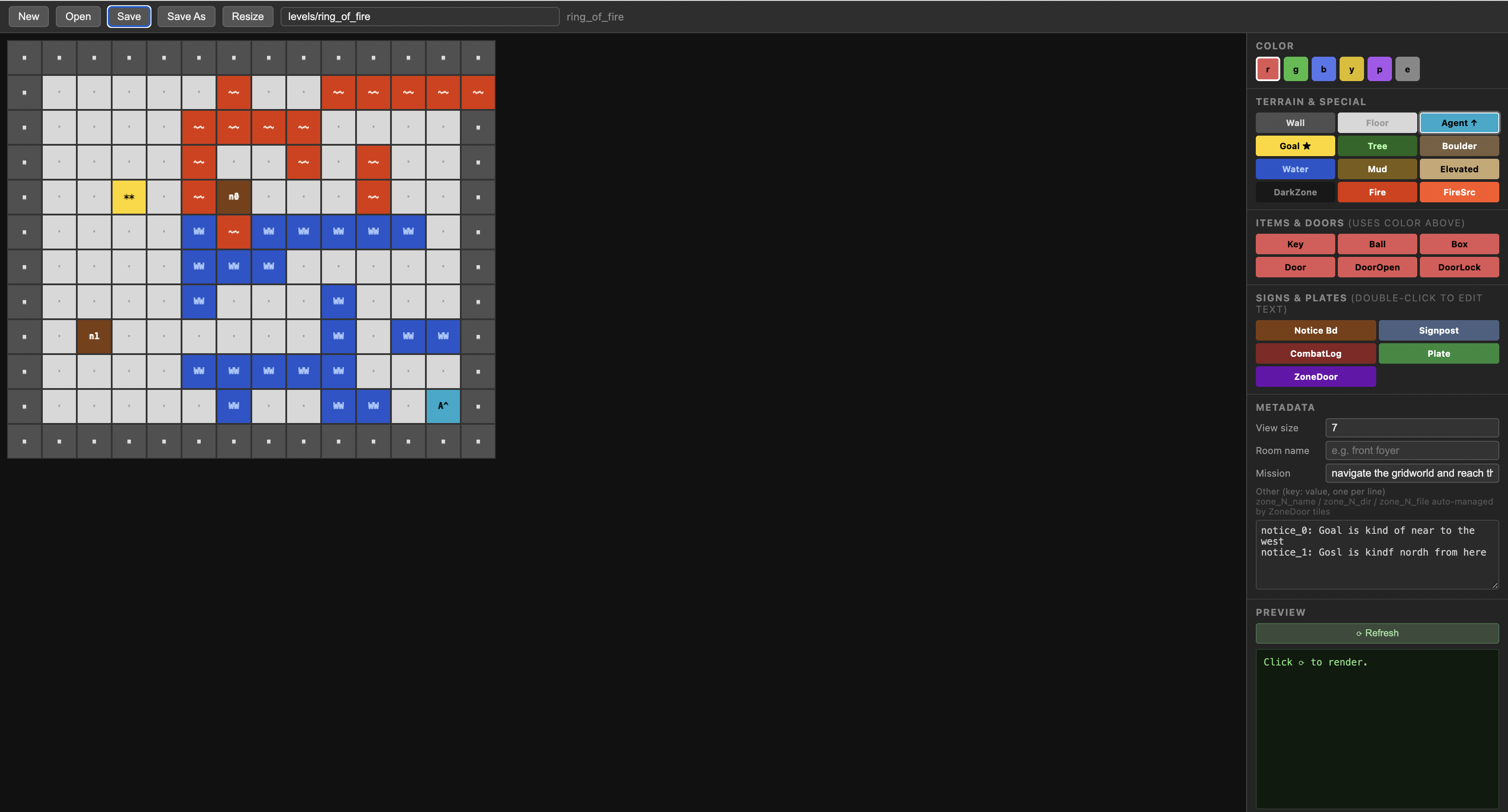}
    \caption{Custom ``Ring of Fire'' level: real path + decoy path with misleading signposts (8 min prototyping).}
    \label{fig:level-editor-custom}
\end{subfigure}
\caption{\textbf{Level Editor interfaces across modalities.} (a) Web-based editor showing C6 Flood Fire Escape's complex mechanics. (b) Terminal ncurses editor for SSH-only environments. (c) Custom created new level demonstrating rapid prototyping. All export to the same plain-text \texttt{.txt} format.}
\label{fig:level-editor-all}
\end{figure}

\paragraph{Rapid prototyping in practice.}
The barrier to creating new test worlds is low: our custom ``Ring of Fire'' level (Figure~\ref{fig:level-editor-custom}), featuring a real path and a decoy path both constrained by water and fire tiles, with signposts that provide misleading navigational clues, was prototyped in under 8 minutes using the web interface. Similarly, the C6 Flood-Fire Escape level (Figure~\ref{fig:level-editor-web}), with timed flood tile advancement, fire mechanics, pushable boulders, pressure plate triggering, and adversarial notice board testimony, demonstrates the editor's support for complex multi-mechanic interactions without requiring manual text file editing.

\subsection{Trajectory Recorder: Ground Truth by Construction}
\label{appendix:trajectory-recorder}

The Trajectory Recorder is a Flask based interactive web application (accessible at \texttt{localhost:5050}) that enables manually navigating through levels to annotate and embed in hallucination probes. Unlike benchmarks requiring post-hoc human labeling or LLM adjudication, our recorder generates naturally generates verifiable ground truth \emph{amidst the process of probe placement}: the human operator walks through the environment using keyboard controls (WASD for movement, G/F for pick/drop, T for toggle), observes the exact simulator state at each step, and plants probes with known-correct answers at moments of interest.

\begin{figure}[ht]                                                        
\centering 
\begin{subfigure}[b]{0.52\textwidth}                                      \centering                                                                \includegraphics[width=\textwidth]{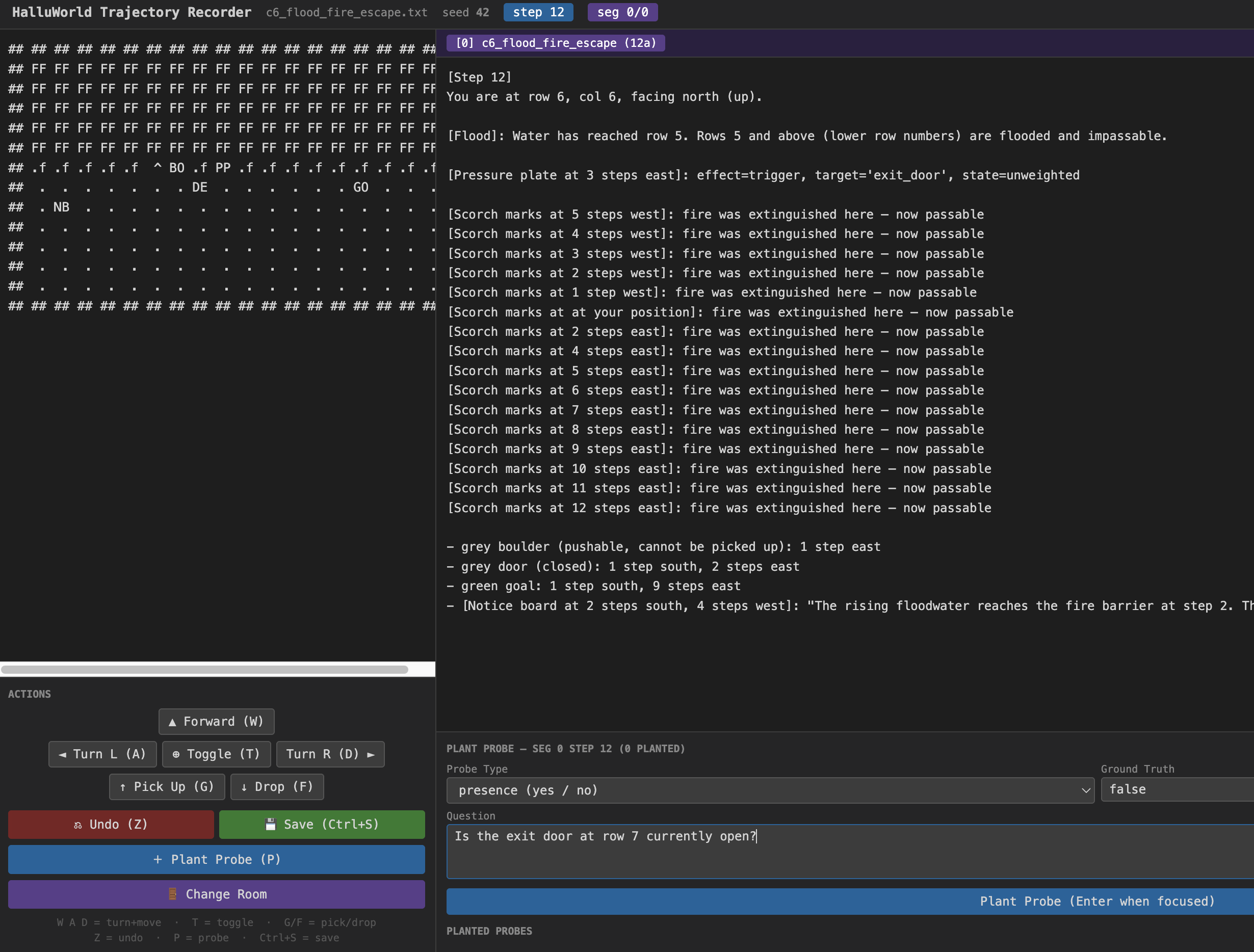}    
\caption{Trajectory Recorder web interface showing grid state (left), observation text (right), and probe planting panel (bottom). Operator has navigated to step~12 in C6 and is planting a probe about the exit door DE being open or not.}            
\label{fig:trajectory-recorder-ui}                                        
\end{subfigure}                                                           \hfill                                                                    \begin{subfigure}[b]{0.42\textwidth}
    \centering
    \footnotesize
    \begin{lstlisting}[language=python,basicstyle=\ttfamily\scriptsize] 
{
  "segments": [
    {
      "level_file": "levels/c6.txt",
      "seed": 42,
      "actions": [2,2,0,2,5,...]
    }
  ],
  "probes": [
    { 
      "segment": 0,
      "step": 12,
      "probe_type": "presence",
      "question": "Is the fire barrier
                   at row 6 currently
                   passable?",
      "ground_truth": "false",
      "metadata": {}
    }
  ]
}
    \end{lstlisting} 
    \caption{Trajectory JSON output showing action sequence and embedded probe. Ground truth (\texttt{false}) is known by operator's observation; fire extinguishes at step~14, not step~12.}
    \label{fig:trajectory-json}
\end{subfigure}         
\caption{\textbf{Trajectory Recorder workflow.} (a) Interactive web UI for manual navigation and probe annotation. (b) Deterministic JSON format consumed by evaluation harness. Ground truth determined by operator's direct observation, not post-hoc adjudication.}     
\label{fig:trajectory-recorder}
\end{figure}

\paragraph{Interactive navigation and probe planting.}
The recorder interface displays two synchronized views (Figure~\ref{fig:trajectory-recorder-ui}): a full grid state panel (left) showing the god's-eye view of all objects, agent position, and hidden state (e.g., whether flood tiles are active, fire extinguished), and an observation text panel (right) showing the agent's field-of-view rendered via \texttt{MemorySerializer}, exactly as models will see it during evaluation. At any trajectory step, the operator can press \texttt{P} to open the probe planting panel, which provides a dropdown for probe type (presence/count/attribute), a text field for the question, and a ground-truth answer field. The operator can verify the correct answer by inspecting the grid state, ensuring probes are neither ambiguous nor inadvertently wrong. Undo functionality (\texttt{Z} key) allows backtracking within the current room segment.

\paragraph{Deterministic replay and output format.}
The recorder saves trajectories as JSON files with two top-level keys (Figure~\ref{fig:trajectory-json}): \texttt{segments} (list of \{\texttt{level\_file}, \texttt{seed}, \texttt{actions}\} dictionaries, one per room) and \texttt{probes} (list of \{\texttt{segment}, \texttt{step}, \texttt{probe\_type}, \texttt{question}, \texttt{ground\_truth}, \texttt{metadata}\} dictionaries). This format is consumed by \texttt{run\_trajectory\_eval.py}, which replays the action sequence deterministically—same level files, same random seeds, same actions—and evaluates model responses against the stored ground truth. The deterministic replay guarantees that different models are probed about identical world states, enabling controlled comparison without confounding by stochastic environment variation.

\subsection{Integrated Workflow: From Prototyping to Evaluation}
\label{appendix:workflow}
The tools form an end-to-end pipeline that lowers the barrier to creating reproducible hallucination benchmarks:
\begin{enumerate}
    \item \textbf{Design}: Researcher sketches a hallucination scenario (e.g., ``test if models trust stale notice boards over direct observation'').

    \item \textbf{Prototype}: Level Editor is used to construct the environment. For C6 Flood-Fire Escape, this involved placing flood tiles with timed activation (rows 1--5, activating at steps 0--4), a fire barrier at row~6 that blocks the agent's path, a pushable boulder at (col\,=\,7, row\,=\,6), a pressure plate at (col\,=\,9, row\,=\,6) linked to an exit door, and a notice board with deliberately wrong flood timing claims (``flood reaches fire at step~2'', actual: step~4)—all configured visually in the web editor.

    \item \textbf{Experience}: Trajectory Recorder loads \texttt{c6\_flood\_fire\_escape.txt} with seed~42. The researcher manually solves the puzzle: waits until step~14 (flood extinguishes fire), navigates to the now-passable fire barrier location, pushes the boulder east twice onto the pressure plate (triggering the exit door to open), and walks to the goal. This manual walkthrough ensures the researcher observes the exact sequence of state changes (flood advancing, fire extinguishing, door opening).

    \item \textbf{Annotate}: At critical steps, the researcher plants probes by pressing \texttt{P}. For example, in the world C6 if one just moves upwards, then at step~8: ``Is the fire barrier at row~6 currently passable?'' (ground truth: \texttt{false}, fire still active). At step~16: ``Is the fire barrier passable now?'' (ground truth: \texttt{true}, fire extinguished at step~14). At step~5: ``According to the notice board, when does the flood reach the fire barrier?'' (ground truth: ``step 2'', testing whether models defer to written information despite observing the actual timing).

    \item \textbf{Evaluate}: \texttt{run\_trajectory\_eval.py} replays the trajectory for each model. At each step, the harness feeds the model the observation (rendered by the specified serializer: grid/memory/symbolic) and trajectory history, then collects the model's probe response. Hallucination rates are computed by comparing model answers to the stored ground-truth labels.

    \item \textbf{Extend}: The level \texttt{.txt} file and trajectory \texttt{.json} are committed to the repository (e.g., \texttt{levels/c6\_flood\_fire\_escape.txt}, \texttt{trajectories/c6\_s42.json}). Other researchers can load the level in the editor to create variants (e.g., change flood timing to step~6, add decoy paths with different fire patterns) or record alternative trajectories with different probe placements (e.g., focus on boulder pushing mechanics instead of flood timing).
\end{enumerate}
This workflow ensures reproducibility: anyone with the repository can rerun evaluations on identical world states (same levels, same seeds, same action sequences). It also enables extensibility: the tools are open-source, and the plain-text formats (level \texttt{.txt}, trajectory \texttt{.json}) are human-readable and version-control-friendly, facilitating pull requests for new levels and trajectories.

\subsection{Integration and Extensibility}
\label{appendix:extensibility}
A key design goal is enabling external researchers to contribute new test cases without modifying core benchmark code. The tools support this through modular architecture and shared trajectory collections for egocentric probing.

\paragraph{Modular object system.}
New object types can be added to \texttt{halluworld/envs/objects/} by subclassing \texttt{WorldObj} and implementing \texttt{\_\_init\_\_}, \texttt{render}, and \texttt{toggle} methods. The Level Editor automatically detects registered object classes via Python's class registry and adds them to the object palette. For example, adding \texttt{TorchObject} (a toggleable light source affecting \texttt{DarkZone} visibility) requires 40 lines of code and zero editor UI changes

\paragraph{Serializer plugins.}
Alternative observation formats can be tested by implementing the \texttt{Serializer} interface (\texttt{serialize(env, step) -> str}). The benchmark harness accepts \texttt{--serializer grid|memory|symbolic} flags, enabling A/B testing of whether representation format affects hallucination rates. Our three serializers differ by $\sim$2$\times$ in verbosity (grid: 150 tokens/step, memory: 80 tokens/step, symbolic: 40 tokens/step), yet show <5pp hallucination rate difference for most models—suggesting format granularity matters less than world complexity for hallucination detection.

\paragraph{Probe type extensibility.}
New probe categories can be added by subclassing \texttt{BaseProbe} and implementing \texttt{generate(env, trajectory, step)} (returns question string and ground truth) and \texttt{evaluate(ground\_truth, model\_response)} (returns correctness boolean). The Trajectory Recorder's probe type dropdown is populated from registered \texttt{BaseProbe} subclasses, making new probe types immediately available in the annotation UI. For instance, a \texttt{SpatialRelationProbe} asking ``Is object X to the left/right of object Y from the agent's perspective?'' (allocentric reasoning) was added by implementing a 60-line subclass and became usable in the recorder without UI code changes.

\paragraph{Shared trajectory repository.}
We release all 250+ navigation trajectories (5 episodes $\times$ 31 worlds $\times$ 1--3 serializers, depending on canonical serializer per world) used in our experiments, enabling exact replication of our evaluation setup even in the egocentric case. Researchers can fork trajectories (e.g., extend M2 Witness Stand from 5 chambers to 10), add new probes to existing trajectories (e.g., test counterfactual reasoning: ``If the agent had turned left at step~5, would the fire be visible at step~10?'') inter alia.

\section{Related Works Expanded}
\label{app:related_works}

\paragraph{Definitions and scope.} Hallucination work in the early days was mainly grounded in a source. For example, in neural machine translation, hallucination was defined as translations detached from the input, and in abstractive summarization, it referred to content unsupported by the source document \cite{Lee2019HallucinationsTranslation,Maynez2020OnSummarization,Ji2024SurveyGeneration}. Later with LLMs, the term broadened to encompass fluent but nonfactual or unfaithful content. Recent audits and benchmark papers note that the field still lacks a single stable definition of hallucination, and often mixes source inconsistency, world-knowledge error, and other failure types under one label \cite{Venkit2024AnNLP,Huang2024AQuestions,Bang2025HalluLens:Benchmark}. The recent ``world model'' view is closest to \benchmarkname. It makes the reference world explicit and treats hallucination as an observable mismatch between the model's internal beliefs and that reference world \cite{liu2026unifieddefinitionhallucinationits}. 

\paragraph{Static hallucination benchmarks in text and RAG.} Summarization benchmarks such as FactCC, QAGS, FRANK, and FaithBench are strong tests of document faithfulness, but they define truth with respect to a fixed source article rather than an evolving world state \cite{Kryscinski2019EvaluatingSummarization,Wang2020AskingSummaries,Pagnoni2021UnderstandingMetrics,Bao2024FaithBench:LLMs}. Open-domain benchmarks such as TruthfulQA, HaluEval, FActScore, HALoGEN, and HalluLens broaden coverage to short-form QA, long-form claim decomposition, and cross-domain verification, but they still evaluate static outputs against fixed facts, corpora, or retrieved evidence \cite{Lin2022TruthfulQA:Falsehoods,Li2023HaluEval:Models,Min2023FActScore:Generation,ravichander-etal-2025-halogen}. Expert-level QA benchmarks such as HLE further stress-test frontier models on difficult academic questions, but primarily measure answer correctness on static problems rather than hallucination as mismatch against an evolving reference world \cite{CenterForAISafety2026HLE}. RAG benchmarks such as RGB, RAGTruth, RAGBench, FACTS Grounding, \emph{Resolving Knowledge Conflicts in Large Language Models}, and ConflictBank study conflicts between parametric and contextual knowledge, but the reference world is still the retrieved or provided evidence rather than a controllable simulator with explicit hidden state and observability \cite{Chen2023BenchmarkingGeneration,Niu2024RAGTruth:Models,Friel2025RAGBench:Systems,Jacovi2025TheInput,Wang2024ResolvingModels,Su2024ConflictBank:LLM}. Recent work on RAG-considerate pretraining studies how models should allocate knowledge between parametric memory and retrieval during pretraining, highlighting that hallucination tendencies may also depend on what information is made externally observable vs. learned in the parameters \cite{singh2026memorizeretrievescalinglaws}.

\paragraph{Multimodal and agent benchmarks.} In VLMs, CHAIR and POPE focus on object hallucination, MMHal-Bench and HallusionBench stress open-ended visual grounding and control-paired reasoning, and AMBER and HaloQuest expand evaluation to attributes, relations, and synthetic imagery \cite{Rohrbach2019ObjectCaptioning,Li2023EvaluatingModels,Sun2023AligningRLHF,Guan2024HallusionBench:Models,Wang2024AMBER:Evaluation,Wang2024HaloQuest:Reasoning}. These benchmarks are central for multimodal grounding, but most are single-scene evaluations and hence do not test hallucinations under a temporally evolving latent state \cite{Bai2025HallucinationSurvey}. For agents, MIRAGE-Bench and AgentHallu are the closest prior work because they explicitly target action-level hallucinations and step attribution, while WebArena, WorkArena, OSWorld, and GAIA expose related failures through realistic task execution \cite{Zhang2025MIRAGE-Bench:Them,Liu2026AgentHallu:Agents,Zhou2024WebArena:Agents,Drouin2024WorkArena:Tasks,Xie2024OSWorld:Environments,Mialon2023GAIA:Assistants}. However, MIRAGE-Bench audits decision snapshots from existing environments and AgentHallu relies on human-annotated trajectories, whereas \benchmarkname is designed around synthetic environments with labels available by construction. 

\paragraph{Controllable environments, world models, and structured reasoning.} Synthetic environments such as bAbI, Dyna-bAbI, TextWorld, BabyAI, and ALFWorld, together with formal planning benchmarks such as PlanBench and ACPBench, demonstrate the value of controllable generation, reproducible dynamics, and exact evaluation \cite{Weston2015TowardsTasks,Tamari2021Dyna-bAbI:Benchmarking,Cote2019TextWorld:Games,Chevalier-Boisvert2019BabyAI:Learning,Shridhar2021ALFWorld:Learning,Valmeekam2023PlanBench:Change,Kokel2026ACPBench:Planning}. But these resources primarily score QA, planning, or task success, not hallucination as an observable false claim. A parallel literature on world models and state tracking - from World Models, PlaNet, MuZero, Dreamer, Dynalang, and language-guided world models to EntNet, CLUTRR, RuleTaker, ProofWriter, BoardgameQA, and recent explicit state-tracking probes - studies whether models can represent and update latent state over time \cite{Ha2018WorldModels,Hafner2019LearningPixels,Schrittwieser2020MasteringModel,Hafner2020DreamImagination,Zhang2024Language-GuidedControl,Lin2024LearningLanguage,Henaff2017TrackingNetworks,Sinha2019CLUTRR:Text,Clark2020TransformersLanguage,Tafjord2021ProofWriter:Language,Kazemi2023BoardgameQA:Information,Li2025HowState,Harang2025TrackingChess}. This literature is relevant to our motivation, but its standard targets are control, prediction, proof generation, or answer correctness rather than automatic hallucination labels against a fully specified reference world. Related work on commonsense-aware generation and reasoning has also explored grounding generation in retrieved visual/contextual evidence, concept-level constraints, and domain-specific reasoning dimensions \cite{lin-etal-2020-commongen,Feng_Lu_Tao_Alikhani_Mitamura_Hovy_Gangal_2022,feng-etal-2021-sapphire,feng-etal-2023-chard}. These approaches target more faithful or semantically constrained generation, whereas \benchmarkname evaluates whether model claims are true against a specified reference world.

\paragraph{Positioning \benchmarkname.} For the specific goal of measuring hallucination as observable false claims in partially observed, evolving worlds, \benchmarkname is not just another source-grounded or static factuality benchmark: unlike prior text and QA benchmarks, it is not limited to fixed documents, facts, or corpora. Unlike RAG-based benchmarks, it does not assume retrieved context is the entire reference world, and unlike multimodal grounding benchmarks, it is not confined to single-scene evaluation. Relative to MIRAGE-Bench and AgentHallu, its main advantage for our purpose is that the world is synthetic and fully specified, so hallucination labels can come from simulator state rather than snapshot audits or post-hoc human annotation. In that sense, \benchmarkname combines four ingredients that prior work usually treats separately: an explicit definition of hallucination, a benchmark, a controllable environment, and automatic labels.

\FloatBarrier 
\end{document}